\documentclass[12pt,a4paper]{article}

\usepackage[a4paper,textwidth=6.5in,textheight=660pt,centering]{geometry}
\usepackage{setspace}
\usepackage{amsmath,amssymb,amsthm,mathtools,bm,aliascnt}
\usepackage{booktabs,array,multirow,tabularx,graphicx}
\usepackage{algorithm}
\usepackage{algpseudocode}
\usepackage{enumitem}
\usepackage{microtype}
\usepackage[authoryear,round]{natbib}
\usepackage[dvipsnames,svgnames,x11names]{xcolor}
\usepackage{hyperref}

\hypersetup{
  colorlinks=true,
  linkcolor=blue,
  anchorcolor=blue,
  citecolor=blue,
  urlcolor=blue
}

\SetSinglespace{1.655}
\AddToHook{env/abstract/begin}{\normalsize}
\newtheorem{theorem}{Theorem}[section]
\newaliascnt{lemma}{theorem}
\newtheorem{lemma}[lemma]{Lemma}
\aliascntresetthe{lemma}
\newaliascnt{corollary}{theorem}
\newtheorem{corollary}[corollary]{Corollary}
\aliascntresetthe{corollary}
\newaliascnt{assumption}{theorem}
\newtheorem{assumption}[assumption]{Assumption}
\aliascntresetthe{assumption}
\newaliascnt{definition}{theorem}
\newtheorem{definition}[definition]{Definition}
\aliascntresetthe{definition}
\newaliascnt{remark}{theorem}
\newtheorem{remark}[remark]{Remark}
\aliascntresetthe{remark}
\newaliascnt{example}{theorem}

\aliascntresetthe{example}

\newcommand{\cS}{\mathcal S}
\newcommand{\cT}{\mathcal T}
\newcommand{\cA}{\mathcal A}
\newcommand{\cD}{\mathcal D}
\newcommand{\cF}{\mathcal F}
\newcommand{\cG}{\mathcal G}
\newcommand{\bE}{\mathbb E}
\newcommand{\bP}{\mathbb P}
\newcommand{\bR}{\mathbb R}
\newcommand{\ind}{\mathbf 1}
\newcommand{\norm}[1]{\left\lVert #1\right\rVert}
\newcommand{\inner}[2]{\left\langle #1,#2\right\rangle}
\newcommand{\KL}{\mathrm{KL}}
\newcommand{\TV}{\mathrm{TV}}
\newcommand{\osc}{\mathrm{osc}}
\newcommand{\argmin}{\mathop{\mathrm{arg\,min}}}
\newcommand{\argmax}{\mathop{\mathrm{arg\,max}}}
\newcommand{\dd}{\,\mathrm d}

\title{\textbf{Neural Actor--Critic Learning with Minimax Critics and Reflected Langevin Sampling}}

\author{
Kunqi Huang
\thanks{School of Mathematics and Statistics,
Wuhan University, Wuhan, 430072, China. Email: huangkunqi@whu.edu.cn }
\and
Lican Kang
\thanks{Institute for Math and AI,   Hubei Key Laboratory of Computational Science, and
School of Artificial Intelligence,
Wuhan University, Wuhan, 430072, China. Email: kanglican@whu.edu.cn }
\and 
Hongrui Shu
\thanks
{School of Artificial Intelligence,
Wuhan University, Wuhan, 430072, China. Email: shuhongrui18@gmail.com }
\and
Jincheng Ying 
\thanks{
School of Mathematics and Statistics,
 Wuhan University, Wuhan, 430072, China.
Email: jinchengying@whu.edu.cn
}
}
\date{}

\begin{document}
\maketitle

\begin{abstract}
Actor--critic methods rely on critic gradients to improve continuous-action policies,
yet finite-sample guarantees must control both value and gradient errors and their
propagation through the actor. We develop a neural actor–critic procedure that
combines a regularized minimax Bellman estimator over sigmoid networks with a finite
reflected Langevin sampler driven by the fitted critic gradient. Under uniform Bellman
regularity and covered reset sampling, a positive high-order action-derivative penalty
yields value and action-gradient risk bounds with classical nonparametric minimax
exponents up to logarithmic factors. We propagate estimation, mixing, discretization,
temperature, and policy-iteration errors to obtain return guarantees, with refinements
for bounded multimodal potentials and strongly concave regimes. Empirical studies
across controlled simulations, logged continuous-control data, and a simulated decision
task show favorable performance relative to the considered baselines.

\vspace{0.5cm}\noindent\textbf{Keywords:}
action-gradient estimation, actor--critic learning, minimax Bellman estimation,
reflected Langevin sampling 
\end{abstract}

\section{Introduction}

Actor--critic methods are a central approach to reinforcement learning
(RL) in continuous-state and continuous-action Markov decision processes
(MDPs).  They alternate between policy evaluation and improvement: the critic
estimates expected discounted returns, and the actor uses these estimates to
select actions.
Classical actor--critic theory studies the coupled updates through
two-time-scale stochastic approximation and compatible function
approximation \citep{konda2000actor,bhatnagar2009natural}.
Deterministic policy gradients make the action derivative of the critic
an explicit component of policy improvement \citep{silver2014dpg}.
Policy performance therefore depends on both critic accuracy and how
the actor uses the fitted value function.
A restrictive actor family can limit this improvement when several
separated actions have high value.  Gibbs policies offer an alternative:
they assign probability across the action space according to an
exponential transform of the value function.  This construction is
central to entropy-regularized control and soft actor--critic
\citep{haarnoja2018sac,geist2019regularized}.  For continuous state
and action spaces, let
$\cG_\tau(Q)$ denote the conditional law with density proportional to
$\exp\{Q(s,a)/\tau\}$, where $Q$ is an action-value function and $\tau>0$
is a temperature.  The action gradient of its log density is
$\tau^{-1}\nabla_aQ$.  Thus a Langevin sampler can use the critic
gradient to generate actions without computing the normalizing integral.
Q-score matching and direct Langevin Q-learning exploit related
connections between value gradients and expressive policies
\citep{psenka2024qscore,ki2026direct}.

The central question is how this sampling construction performs
when its critic is estimated from finitely many observations.
Small prediction error alone does not guarantee that the fitted
gradient is an accurate sampling drift.  This derivative-estimation
problem differs from controlling perturbations of the invariant Gibbs
law: these perturbations can also be bounded using value error alone.  Moreover, the next policy is
the output law of a finite numerical procedure, not its invariant Gibbs
law.  Errors in this output affect both the next critic fit and
subsequent policy improvement.  For fixed-policy evaluation,
\citet{chen2022wellposed} establish nonparametric rates for
both the action-value function and its derivatives.  Deep Sobolev
regression provides guarantees for neural function and derivative estimation without
derivative responses \citep{ding2025sobolev}, while constrained Langevin
theory controls mixing and discretization, including for nonconvex
potentials \citep{lamperski2021projected,zheng2022constrained}.
Connecting these results requires compatible error measures and
regularity conditions throughout the changing critic--actor sequence.
Temperature introduces a further interaction.  Our objective is the
ordinary discounted return $J$, and $\pi^\star$ denotes a stationary
Markov policy maximizing this return.  The Gibbs map approximates
greedy improvement; it does not introduce trajectory-level entropy
regularization into policy evaluation.  Reducing $\tau$ decreases
the approximation bias of the greedy step, but also increases
sensitivity to critic errors and can impede movement between action
modes.  Consequently, a fixed-temperature sampling guarantee does not
by itself establish convergence to an ordinary-return optimum.
The temperature, critic accuracy, sampling effort, and number of
policy updates must be calibrated jointly.

We address this problem by combining a regularized minimax Bellman
critic with a finite reflected Langevin actor.  At update $k$, a block of
transition observations and next-action draws from the current policy
$\widehat\pi_k$ is used to estimate $Q^{\widehat\pi_k}$ by
$\widehat Q_{k+1}$.  Its action gradient drives a finite reflected
sampling scheme whose output law is $\widehat\pi_{k+1}$.  The fitted
Gibbs law $\cG_\tau(\widehat Q_{k+1})$ is an analytical target for this
scheme, not a density that must be estimated or evaluated.  Both
function classes in the sample minimax criterion are sigmoid deep
neural networks (DNNs), with bounded parameters and outputs but no hard
H\"older constraint.  A strictly positive penalty on high-order action
derivatives is evaluated at independent uniform points.  These points
require neither new transition observations nor derivative labels.
The analysis links regularized Bellman estimation to finite sampling
and policy improvement.  A localized empirical-process bound retains
the high-order penalty, so Bellman stability and interpolation yield
function and gradient risk bounds.  Gibbs density and state-occupancy bounds
transfer these bounds to actor and return errors, while coupling and
entropy smoothing control the projection scheme uniformly in sampling
time.  Explicit dependence on temperature and the learned drift permits joint
error calibration for multimodal policies and sharper bounds under
strong concavity.
We evaluate the first-order implementation in two controlled Markov
environments and compare prediction and fixed-critic policy extraction
on six logged continuous-control datasets.  A supplementary study in
Appendix~\ref{app:energy-utility} assesses finite-action decisions on
simulated building-energy data.  A detailed discussion of related work
is provided in Appendix~\ref{app:related-work}.

\subsection{Contributions}

The contributions are as follows.
\begin{enumerate}[leftmargin=2.1em,label=\textbf{(\roman*)}]
\item \textbf{A sampled Bellman--Langevin actor--critic method.}
We combine a subtractive squared-loss Bellman criterion, a positive
action-derivative penalty, and a reflected Langevin actor.
Next-action observations come from the current finite sampler, so critic
estimation and return analysis concern the same policy.  The covered
reset design specifies how each transition block is generated; neither derivative
labels nor policy-density evaluation are required.

\item \textbf{A statistical-risk-to-return guarantee.}
Under the stated regularity and coverage conditions, with network
depth of order $\log n$ and size of order
$n^{d/(2\beta+d)}\log n$, we balance a positive high-order penalty
against approximation and empirical-process errors to prove
\begin{align*}
\bE_k\|\widehat Q_{k+1}-Q^{\widehat\pi_k}\|_{L^2(\mu_k)}^2
&\lesssim n^{-2\beta/(2\beta+d)}(\log n)^3,\\
\bE_k\|\nabla_a\widehat Q_{k+1}-\nabla_aQ^{\widehat\pi_k}\|_{L^2(\lambda_d)}^2
&\lesssim n^{-2(\beta-1)/(2\beta+d)}(\log n)^{3(\beta-1)/\beta},
\end{align*}
where $n$ is the common sample size per update, $\beta>2$ is the
H\"older smoothness, $d=d_s+d_a$ is the sum of the state and action
dimensions, $\mu_k$ is the conditional state--action law in the covered
reset experiment, $\lambda_d$ is the uniform probability measure on the joint cube,
and $\bE_k$ denotes expectation conditional on the history preceding
update $k$.  The value and action-gradient risk bounds have the classical
nonparametric minimax exponents up to logarithmic factors
\citep{stone1982optimal,chen2022wellposed}.
For bounded
multimodal potentials, logarithmic cooling gives
consistency with a quasipolynomial worst-case sampling budget in $n$.
This multimodal guarantee uses the value-risk branch of the regularized
critic analysis, without invoking the statistical gradient-risk bound.
Under the additional
strong-concavity conditions, direct control of the policy-improvement
error by the squared value and gradient risks gives
\begin{equation*}
\bE\{J(\pi^\star)-J(\widehat\pi_K)\}
\lesssim\frac{1}{(1-\gamma)^2}\min\Bigl\{
n^{-\frac{2\beta}{(2\beta+d)(d_a+2)}}
(\log n)^{\frac{3}{d_a+2}},\;
n^{-\frac{2(\beta-1)}{(2\beta+d)(d_a+1)}}
(\log n)^{\frac{3(\beta-1)}{\beta(d_a+1)}}\Bigr\}.
\end{equation*}
\item \textbf{Empirical assessment.}
Across 100 replications per controlled environment, the first-order
implementation has higher mean final returns than uniform and
polynomial-critic controls.  On six logged datasets, the sigmoid DNN
has lower displayed mean prediction errors than Linear; comparisons with
Quadratic and policy rankings under training-target standardized
discounted return depend on the dataset.  On simulated building-energy
data, neural-critic decisions have lower mean heating load and regret
than Linear and Uniform.
\end{enumerate}

\subsection{Paper Organization}
Section~\ref{sec:setup} introduces the MDP, neural function classes, and
sampling preliminaries.  Section~\ref{sec:algorithm} presents the critic
estimator and the executable actor update, with pseudocode in
Appendix~\ref{app:algorithm}.  Section~\ref{sec:theory}
establishes the critic, sampling, and return guarantees, including
temperature calibrations for multimodal and strongly concave potentials.
Section~\ref{sec:experiments} reports the interactive actor--critic
experiments and the six-dataset continuous-control study.
The appendices provide the related-work discussion
(Appendix~\ref{app:related-work}), the actor--critic pseudocode
(Appendix~\ref{app:algorithm}), complete proofs and supporting
analytical results, and supplementary experimental details
(Appendix~\ref{app:experiments}).

\section{Preliminaries}
\label{sec:setup}

Throughout, $n$ denotes the sample size per update, $\bE$ and $\bP$ denote
expectation and probability, and $\bE_k(\cdot)=\bE(\cdot\mid\cF_k)$,
where $\cF_k$ is the information available before update $k$, including
the transition blocks and penalty integration points from all earlier
updates.  The conditional expectation $\bE_k$ averages over the transition
block and integration points generated at the current update.
The Euclidean norm and inner product are $\|\cdot\|_2$ and
$\langle\cdot,\cdot\rangle$.  For a probability measure $\nu$ and $1\le p<\infty$, write
$\|f\|_{L^p(\nu)}=(\int\|f\|_2^p\,\dd\nu)^{1/p}$, with absolute value
for scalar $f$, and $\|f\|_\infty=\sup_z\|f(z)\|_2$.

Let $\lambda_{\cS}$, $\lambda_{\cA}$, and
$\lambda_d=\lambda_{\cS}\otimes\lambda_{\cA}$ be normalized Lebesgue
measures on the state, action, and joint cubes, respectively.  At update
$k$, the queried pair $Z_{k,i}=(S_{k,i},A_{k,i})$ has conditional design law
\begin{equation*}
\mu_k(B):=\bP\{Z_{k,i}\in B\mid\cF_k\},
\qquad B\in\mathcal B(\cS\times\cA),
\end{equation*}
where $\mathcal B(E)$ denotes the Borel sigma-field of $E$.
Assumption~\ref{ass:design} constructs this joint law by geometric
resetting.  The critic's value risk is measured in $L^2(\mu_k)$ and its
action-gradient risk in $L^2(\lambda_d)$.  Neither measure is the
conditional actor law.  Unless stated otherwise, action integrals use
$\lambda_{\cA}$.
The operators $\nabla_a$ and $\nabla_a^2$ are the action gradient and
Hessian.  For a matrix-valued function $H$, let
$\|H\|_{\mathrm{op},\infty}=\sup_z\|H(z)\|_{\mathrm{op}}$.
We write $\osc f=\operatorname*{ess\,sup}f-\operatorname*{ess\,inf}f$ and use
$\operatorname{Var}$ for variance.
The relation $x\lesssim y$ means $x\le Cy$
for a constant depending only on fixed structural parameters, including
the H\"older radii;
$x\asymp y$ means $x\lesssim y$ and $y\lesssim x$.
For symmetric matrices, 
$A\succeq B$ means that $A-B$ is positive semidefinite.

\subsection{Continuous-state/action MDP and actor--critic objective}

Consider the discounted MDP $(\cS,\cA,P,\mathcal R,\gamma,\rho_0)$ with bounded
rectangular state and action domains.  After affine rescaling, we take
$\cS=[0,1]^{d_s}$ and $\cA=[0,1]^{d_a}$ without loss of generality,
with joint dimension $d=d_s+d_a$.
Here $P(\cdot\mid s,a)$ and $\mathcal R(\cdot\mid s,a)$ are the
next-state and reward distributions of the environment's fixed Markov
transition mechanism.  The reward distribution is supported on
$[-R_{\max},R_{\max}]$ for a fixed $R_{\max}>0$,
$\gamma\in[0,1)$ is the discount factor, and $\rho_0$ is the initial-state
distribution.  Starting from $S_0\sim\rho_0$, the policy generates $A_t$,
and, conditional on the history through $(S_t,A_t)$, the environment
generates the reward and next state:
\[
R_t\sim\mathcal R(\cdot\mid S_t,A_t),\qquad
S_{t+1}\sim P(\cdot\mid S_t,A_t),\qquad
|R_t|\le R_{\max}\quad\text{almost surely}.
\]
Each critic observation also includes one next action sampled from the
current actor.

A policy $\pi$ is a Markov kernel from $\cS$ to $\cA$.  The method accesses policy samples through $A\sim\pi(\cdot\mid s)$, without an explicit
density or a policy-score oracle.  The policy's discounted return, state-value function, and action-value function are
\begin{align*}
J(\pi)
&:=
\bE_{\rho_0,\pi,P,\mathcal R}
\left[\sum_{t=0}^{\infty}\gamma^tR_t\right],
\\
V^\pi(s)
&:=
\bE_{\pi,P,\mathcal R}
\left[\left.\sum_{t=0}^{\infty}\gamma^tR_t\right|S_0=s\right],
\\
Q^\pi(s,a)
&:=
\bE_{\pi,P,\mathcal R}
\left[\left.\sum_{t=0}^{\infty}\gamma^tR_t\right|S_0=s,A_0=a\right].
\end{align*}
The value functions satisfy $V^\pi(s)=\int_{\cA}Q^\pi(s,a)\pi(\dd a\mid s)$ and
$\norm{V^\pi}_\infty\vee\norm{Q^\pi}_\infty\le
B_Q:=R_{\max}/(1-\gamma)$.

The target is a policy maximizing the ordinary discounted return:
$$
\pi^\star\in\argmax_{\pi}J(\pi),
$$
where the maximum is over stationary Markov policies.
Section~\ref{sec:end-to-end} establishes attainment under primitive
continuity conditions; $\pi^\star$ need not have a density.

For a policy $\pi$, its normalized discounted state-occupancy measure is
\begin{equation*}
d_\pi^{\rho_0}(B)
:=
(1-\gamma)\sum_{t=0}^{\infty}\gamma^t
\bP_{\rho_0,\pi,P}(S_t\in B),
\qquad B\subseteq\cS\ \text{Borel}.
\end{equation*}
This is a probability measure.  For any two policies $\pi$ and $\pi'$, the
performance-difference identity gives
\begin{equation}
\label{eq:performance-difference}
J(\pi')-J(\pi)
=
\frac{1}{1-\gamma}
\int_{\cS}d_{\pi'}^{\rho_0}(\dd s)
\left\{
\int_{\cA}Q^\pi(s,a)\,\pi'(\dd a\mid s)
-
\int_{\cA}Q^\pi(s,a)\,\pi(\dd a\mid s)
\right\}.
\end{equation}
Equation~\eqref{eq:performance-difference} expresses return improvement
as an occupancy-weighted action-value difference.

For a bounded function $Q:\cS\times\cA\to\bR$, define the policy-evaluation Markov operator and Bellman operator by
\begin{align*}
(P^\pi Q)(s,a)
&:=\bE_{S^{\prime}\sim P(\cdot\mid s,a),\,A^{\prime}\sim\pi(\cdot\mid S^{\prime})}Q(S^{\prime},A^{\prime}),
\\
(\cT^\pi Q)(s,a)
&:=\bE_{\pi,P,\mathcal R}
\{R_t+\gamma Q(S_{t+1},A_{t+1})\mid S_t=s,A_t=a\}.
\end{align*}
The operator $\cT^\pi$ is a $\gamma$-contraction in the supremum norm, and $Q^\pi$ is its unique bounded fixed point.  The temperature $\tau>0$ enters only the actor map below.

For a bounded measurable state function $V$, define the state-level policy and optimality
operators by
\begin{align*}
(\cT_V^\pi V)(s)
&:=
\int_{\cA}\bE_{P,\mathcal R}
\{R_t+\gamma V(S_{t+1})\mid S_t=s,A_t=a\}
\pi(\dd a\mid s),\\
(\cT_V^\star V)(s)
&:=
\sup_{a\in\cA}\bE_{P,\mathcal R}
\{R_t+\gamma V(S_{t+1})\mid S_t=s,A_t=a\}.
\end{align*}
The subscript $V$ distinguishes the state-value operators from the
state--action operator $\cT^\pi$.
In particular,
$V^\pi=\cT_V^\pi V^\pi$, and $V^\star$ is the fixed point of
$\cT_V^\star$.

\subsection{Gibbs policies and entropy divergences}

\begin{definition}[Gibbs improvement map]
For bounded measurable $Q$, define
\begin{equation*}
[\cG_\tau(Q)](\dd a\mid s)
:=
\frac{\exp\{Q(s,a)/\tau\}}
{Z_Q(s)}\dd a,
\qquad
Z_Q(s):=\int_{\cA}\exp\{Q(s,u)/\tau\}\dd u.
\end{equation*}
\end{definition}

\begin{definition}[Relative entropy and relative Fisher information]
For probability measures $\mu$ and $\nu$ on $\cA$ with $\mu\ll\nu$, meaning that $\mu$ is absolutely continuous with respect to $\nu$, the Kullback--Leibler (KL) divergence, also called relative entropy, is
\begin{equation*}
\KL(\mu\|\nu)
:=
\int_{\cA}\log\!\left(\frac{\dd\mu}{\dd\nu}\right)\dd\mu.
\end{equation*}
If $\mu$ and $\nu$ admit positive weakly differentiable Lebesgue densities $p_\mu$ and $p_\nu$ and $\nabla_a\log(p_\mu/p_\nu)\in L^2(\mu)$, their relative Fisher information is
\begin{equation*}
I(\mu\|\nu)
:=
\int_{\cA}
\norm{\nabla_a\log(p_\mu/p_\nu)}_2^2\dd\mu(a).
\end{equation*}
\end{definition}

\begin{definition}[Log-Sobolev inequality]
A probability measure $\nu$ on $\cA$ with a positive Lebesgue density
satisfies a log-Sobolev inequality (LSI) with constant $\alpha>0$ if
$
I(\mu\|\nu)\ge2\alpha\KL(\mu\|\nu)
$
for every probability measure $\mu$ with a sufficiently regular density.
\end{definition}
For Langevin dynamics with invariant law $\nu$, an LSI turns entropy
dissipation into exponential convergence in relative entropy.

For arbitrary probability measures $\mu$ and $\nu$, let
$p_\mu=\dd\mu/\dd\lambda$ and $p_\nu=\dd\nu/\dd\lambda$ for any
common dominating measure $\lambda$ (for example, $\mu+\nu$).
The total variation (TV) distance is
\[
\TV(\mu,\nu):=\frac12\int|p_\mu-p_\nu|\dd\lambda.
\]
This definition is independent of the dominating measure and also applies
to policy kernels with boundary mass.

For probability measures $\mu$ and $\nu$ on the compact action cube, define the
Wasserstein--1 distance
\begin{equation*}
W_1(\mu,\nu)
:=
\inf_{\Gamma\in\Pi(\mu,\nu)}
\int_{\cA\times\cA}\norm{a-a'}_2\Gamma(\dd a,\dd a'),
\end{equation*}
where $\Pi(\mu,\nu)$ is the set of couplings of $\mu$ and $\nu$.  Its
state-integrated counterpart is
\begin{equation*}
\mathsf w_1(\pi,\pi')
:=
\int_{\cS}W_1\{\pi(\cdot\mid s),\pi'(\cdot\mid s)\}
\lambda_{\cS}(\dd s).
\end{equation*}
The distance $\mathsf w_1$ controls both Lipschitz policy-improvement
residuals and projection-mesh error, without a convexity requirement on
the Gibbs potential.

\subsection{Regularity classes, sigmoid DNNs, and entropy quantities}

\begin{definition}[H\"older class]
Write $z=(s,a)\in[0,1]^d$.  For a multi-index $\alpha$, let $|\alpha|$ be its total order and $D^\alpha$ the corresponding mixed derivative.  For a smoothness index $\beta>0$, write $\beta=r_\beta+\nu$, where $r_\beta=\lceil\beta\rceil-1$ is the integer derivative order and $\nu\in(0,1]$ is the H\"older exponent.  Define
\begin{equation*}
\norm{f}_{\mathcal H^\beta}
:=
\max_{|\alpha|\le r_\beta}\norm{D^\alpha f}_\infty
+
\max_{|\alpha|=r_\beta}
\sup_{z\ne z'}
\frac{|D^\alpha f(z)-D^\alpha f(z')|}
{\norm{z-z'}_2^\nu}.
\end{equation*}
For a fixed radius $L>0$, the H\"older class is
\begin{equation*}
\mathcal H^\beta(L)
:=
\{f:[0,1]^d\to\bR:\norm{f}_{\mathcal H^\beta}\le L\}.
\end{equation*}
\end{definition}

\begin{definition}[Sigmoid deep neural network]
The logistic sigmoid activation is
$
\sigma_{\rm sig}(u)=\frac{1}{1+e^{-u}}.
$
A feed-forward sigmoid DNN with input dimension $p_0=d$ and hidden-layer dimensions $p_1,\ldots,p_{\mathcal D}$ is defined recursively by
\begin{equation*}
h_0(z)=z,\qquad
h_\ell(z)=\sigma_{\rm sig}\{W_\ell h_{\ell-1}(z)+b_\ell\},
\quad \ell=1,\ldots,\mathcal D,\qquad
f_\theta(z)=w^\top h_{\mathcal D}(z)+b,
\end{equation*}
where $W_\ell$ and $b_\ell$ are the hidden-layer weights and biases, $w$ and $b$ are the output weight vector and bias, and $\theta$ collects all network parameters.  Define the width, depth, and nonzero-parameter size by
\begin{equation*}
\mathcal W(f_\theta):=\max_{1\le\ell\le\mathcal D}p_\ell,
\qquad
\mathcal D(f_\theta):=\mathcal D,
\qquad
\mathcal M(f_\theta):=\norm{\theta}_0,
\end{equation*}
where $\norm{\theta}_0$ counts the nonzero parameters of this
representation.  A parameter bound $\mathcal B>0$ means
$\norm{\theta}_\infty\le\mathcal B$.  The activation acts coordinatewise.
A uniform bound $B$ on function values means
$\|f_\theta\|_\infty\le B$.
\end{definition}

\begin{definition}[Trainable sigmoid critic and auxiliary classes]
\label{def:localized-sigmoid-classes}
Fix a smoothness index $\beta>2$ and let $m=\lceil\beta\rceil$ be the
integer order of the action-derivative penalty.  For architecture bounds
$(\mathcal W_n,\mathcal D_n,\mathcal M_n)$ and parameter bound
$\mathcal B_n$, let $\Theta_n$ be the finite union, over all
feed-forward sparsity patterns satisfying the width and depth bounds,
of parameter cubes with at most
$\mathcal M_n$ nonzero entries and $\norm\theta_\infty\le\mathcal B_n$.
The associated networks have at most $\mathcal D_n$ hidden layers and at
most $\mathcal W_n$ neurons per hidden layer.  Define
\begin{equation*}
\cF_n^{\rm sig}
:=\{f_\theta:\theta\in\Theta_n,\
  \norm{f_\theta}_\infty\le2B_Q\},\qquad
\cG_n^{\rm sig}
:=\{g_\vartheta:\vartheta\in\Theta_n,\
  \norm{g_\vartheta}_\infty\le2B_Q\}.
\end{equation*}
\end{definition}

For Gibbs potentials on the cube, the following profile collects the
constants used in the sampling bounds.

\begin{definition}[Conditional sampling profile]

Fix a temperature $\tau\in(0,1]$ and a family $\mathcal P_F$ of action
potentials on $\cA$.  A triple
$(\alpha_\tau,\chi_\tau,D_\tau)$ is a sampling profile for $\mathcal P_F$ if,
for every $F\in\mathcal P_F$, the Gibbs law
$\pi_F(\dd a)\propto e^{F(a)/\tau}\dd a$ satisfies
\begin{equation*}
I(\nu\|\pi_F)\ge2\alpha_\tau\KL(\nu\|\pi_F),\qquad
W_1^2(\nu,\pi_F)\le2\chi_\tau\KL(\nu\|\pi_F),\qquad
\left\|\frac{\dd\pi_F}{\dd\lambda_{\cA}}\right\|_\infty\le D_\tau
\end{equation*}
for every admissible probability law $\nu$.
\end{definition}

Two regimes yield explicit profiles.  For a continuously
differentiable potential satisfying $\norm F_\infty\le B_F$, with no
restriction on its number of modes, bounded perturbation of the uniform
law, Pinsker's inequality, and the Gibbs normalizer give
\begin{equation}
\label{eq:multimodal-profile}
\alpha_\tau=\widetilde\alpha e^{-2B_F/\tau},
\qquad
\chi_\tau=d_a/4,
\qquad
D_\tau=e^{2B_F/\tau},
\end{equation}
where $\widetilde\alpha>0$ is the LSI constant of normalized Lebesgue
measure on the cube. 
 If every potential is also $\rho$-strongly concave, where
$\rho>0$ is a fixed curvature parameter, the convex-domain
Bakry--Emery criterion and its Herbst transport consequence give
the faster profile
\begin{equation}
\label{eq:strong-concave-profile}
\alpha_\tau=\rho/\tau,
\qquad
\chi_\tau=\tau/\rho,
\qquad
D_\tau\lesssim\tau^{-d_a}.
\end{equation}
Here, the constant in the density bound depends only on $B_F$ and $d_a$:
concavity provides a homothetic neighborhood of a maximizer on which
the potential is nearly maximal.  
Lemma~\ref{lem:profile-verification}
establishes both profiles, including the case of boundary maximizers.
The profiles describe the target distributions; separate,
architecture-dependent drift bounds control numerical discretization.

\subsection{Reflected Langevin dynamics on the action cube}
\label{sec:langevin-prelim}

Let $F:\cS\times\cA\to\bR$ be jointly measurable and continuously differentiable in the action argument, with $a\mapsto\nabla_aF(s,a)$ Lipschitz uniformly in $s$.  For a fixed state $s$, the normally reflected overdamped Langevin diffusion associated with $F$ is the solution of the Skorokhod stochastic differential equation (SDE)
\begin{equation}
\label{eq:reflected-langevin-prelim}
\dd A_t
=
\nabla_aF(s,A_t)\,\dd t
+
\sqrt{2\tau}\,\dd B_t
+
\dd K_t,
\qquad A_t\in\cA.
\end{equation}
Here, $B_t$ is standard $d_a$-dimensional Brownian motion and $K_t$ is the boundary regulator
\begin{equation}
\label{eq:cube-local-times}
\dd K_t
=
\sum_{j=1}^{d_a}e_j\,\dd L_t^{j,0}
-
\sum_{j=1}^{d_a}e_j\,\dd L_t^{j,1}.
\end{equation}
The vector $e_j$ is the $j$th coordinate vector, and $L_t^{j,0}$ and
$L_t^{j,1}$ are nondecreasing boundary local times supported on
$\{A_{t,j}=0\}$ and $\{A_{t,j}=1\}$, respectively.  The regulator keeps the
path in the cube and acts only in inward normal directions.
Lemma~\ref{lem:aux-reflection} gives existence and uniqueness.

For a twice continuously differentiable test function $\varphi$, the interior generator is
\begin{equation*}
\mathcal L_{F,s}\varphi(a)
=
\inner{\nabla_aF(s,a)}{\nabla_a\varphi(a)}
+
\tau\Delta_a\varphi(a),
\end{equation*}
where $\Delta_a$ is the action-coordinate Laplacian.  The generator is equipped with the reflecting (Neumann) boundary condition.  If $\rho_t$ denotes the density of $A_t$, its probability flux is
\[
j_t(a)=\rho_t(a)\nabla_aF(s,a)-\tau\nabla_a\rho_t(a),
\qquad
j_t\cdot n_{\rm out}=0\quad\text{on each face of }\partial\cA,
\]
where $n_{\rm out}$ is the outward unit normal to that face; no single
normal vector is assigned to edges or corners.
The Gibbs density $\pi_F(\cdot\mid s):=\cG_\tau(F)(\cdot\mid s)$
is invariant because $\tau\nabla_a\pi_F=\pi_F\nabla_aF$ makes the flux
identically zero.

\section{Methodology}
\label{sec:algorithm}

The proposed method alternates between regularized critic estimation and conditional
Langevin sampling.  Its population counterpart is temperature-smoothed
policy iteration initialized at $\pi_0$, with exact policy evaluation
followed by Gibbs improvement:
\begin{equation*}
\pi_{k+1}(\dd a\mid s)
=
\frac{\exp\{Q^{\pi_k}(s,a)/\tau\}}
{\int_{\cA}\exp\{Q^{\pi_k}(s,u)/\tau\}\dd u}\dd a
=\cG_\tau(Q^{\pi_k})(\dd a\mid s),
\quad k\ge0.
\end{equation*}
The first two population updates are
\begin{equation*}
\pi_0\longrightarrow Q^{\pi_0}
\longrightarrow\pi_1=\cG_\tau(Q^{\pi_0})
\longrightarrow Q^{\pi_1}
\longrightarrow\pi_2=\cG_\tau(Q^{\pi_1}).
\end{equation*}
For each $k$, the reflected Langevin SDE with drift
$\nabla_aQ^{\pi_k}$ has invariant conditional law $\pi_{k+1}$.
The sample-based method approximates this recursion with $K$ outer
actor--critic iterations, indexed by
$k=0,\ldots,K-1$.

At iteration $k$, a fresh covered transition
block $\cD_k$ with state--action design law $\mu_k$ is used to estimate
$Q^{\widehat\pi_k}$ as $\widehat Q_{k+1}$; the current finite-chain policy
$\widehat\pi_k$ supplies the next-action draw in each Bellman target.  Initialize
$\widehat\pi_0=\pi_0$.  The fitted Gibbs law
\begin{equation*}
\widetilde\pi_{k+1}:=\cG_\tau(\widehat Q_{k+1})
\end{equation*}
is the invariant law of the fitted reflected diffusion, while
$\widehat\pi_{k+1}$ is the terminal law after $M$ frozen-drift macrosteps,
each evaluated on a polygonal Brownian microgrid of mesh $\delta>0$.
The implemented policy $\widehat\pi_{k+1}$ supplies next-action draws
at iteration $k+1$; $\widetilde\pi_{k+1}$ serves as its analytical
sampling target.  The resulting recursion is
\begin{equation}
\label{eq:ac-arrow-iteration}
\begin{aligned}
\widehat\pi_k
&\ \xrightarrow{\text{generate one-step block}}
\ \cD_k
\ \xrightarrow[\text{target: }Q^{\widehat\pi_k}=\cT^{\widehat\pi_k}Q^{\widehat\pi_k}]
{\text{regularized minimax fit}}
\ (\widehat Q_{k+1},\nabla_a\widehat Q_{k+1})
\\[-1mm]
\widehat Q_{k+1}
&\ \xrightarrow[\text{invariant target: }\widetilde\pi_{k+1}=\cG_\tau(\widehat Q_{k+1})]
{\text{$\nabla_a\widehat Q_{k+1}$ drives $M$ reflected macrosteps}}
\ \widehat\pi_{k+1}.
\end{aligned}
\end{equation}
Each update in \eqref{eq:ac-arrow-iteration} uses the same block size $n$, network
architecture $(\mathcal W_n,\mathcal D_n,\mathcal M_n)$, penalty coefficient
$\lambda_n>0$, derivative order $m=\lceil\beta\rceil$, number of uniform
integration points $\ell_n$, temperature $\tau$, and sampler settings $(M,\eta,\delta)$.
Algorithm~\ref{alg:sac-sampling} in Appendix~\ref{app:algorithm}
summarizes this recursion.  The critic estimator and actor sampler
are specified below.

\subsection{Critic: regularized minimax Bellman estimation}
\label{sec:critic-method}

Conditional on the pre-update history $\cF_k$, the policy $\widehat\pi_k$ and covered design law
$\mu_k$ are fixed.  Write $Z=(S,A)\sim\mu_k$,
$R\sim\mathcal R(\cdot\mid S,A)$, and
$Z^{\prime}=(S^{\prime},A^{\prime})$, where
$S^{\prime}\sim P(\cdot\mid S,A)$ and an independent actor call gives
$A^{\prime}\sim\widehat\pi_k(\cdot\mid S^{\prime})$.  At $k=0$, the next-action call uses the supplied sampler $\pi_0$;
for $k\ge1$, it reruns the reflected Euler--Maruyama (REM) projection
microgrid with fresh randomness.  The action-value function satisfies the Bellman equation
\begin{equation*}
Q^{\widehat\pi_k}(s,a)
=(\cT^{\widehat\pi_k}Q^{\widehat\pi_k})(s,a)
=\bE_k\left[R+\gamma Q^{\widehat\pi_k}(Z^{\prime})\mid Z=(s,a)\right].
\end{equation*}
The population criterion uses the critic and auxiliary H\"older classes
$\mathcal H^\beta(L)$ and $\mathcal H^\beta(L_{\cT})$,
respectively, with the fixed auxiliary radius enlarged as in
Lemma~\ref{lem:population-bellman-extension}.
This population comparison uses the architecture sequence calibrated
in Theorem~\ref{thm:critic-l2}; the fitted class itself has no fixed
H\"older constraint.  For bounded measurable $f$ and
$g\in\mathcal H^\beta(L_{\cT})$, let
$Y_k(f):=R+\gamma f(Z^{\prime})$ and define
\begin{equation}
\mathcal L_k(f,g)
:=\bE_k\!\left[
\{f(Z)-Y_k(f)\}^2-\{g(Z)-Y_k(f)\}^2
\right].
\label{eq:population-minimax-loss}
\end{equation}
Because
$\bE_k\left[Y_k(f)\mid Z\right]=(\cT^{\widehat\pi_k}f)(Z)$, the conditional-mean
Pythagorean identity gives
\begin{equation}
\mathcal L_k(f,g)
=
\norm{f-\cT^{\widehat\pi_k}f}_{L^2(\mu_k)}^2
-
\norm{g-\cT^{\widehat\pi_k}f}_{L^2(\mu_k)}^2.
\label{eq:minimax-identification}
\end{equation}
The design law $\mu_k$ need not be the state--action law induced by
$\widehat\pi_k$.  Assumption~\ref{ass:holder} places
$\cT^{\widehat\pi_k}f$ in the auxiliary class for
$f\in\cF_n^{\rm sig}$; Lemma~\ref{lem:population-bellman-extension}
extends this property to $f\in\mathcal H^\beta(L)$ by approximation,
closedness, and the affine Bellman identity.  Therefore, on
the union of these two critic classes, \eqref{eq:minimax-identification} gives
\begin{equation*}
\sup_{g\in\mathcal H^\beta(L_{\cT})}\mathcal L_k(f,g)
=
\norm{f-\cT^{\widehat\pi_k}f}_{L^2(\mu_k)}^2.
\end{equation*}
Under the coverage condition in Assumption~\ref{ass:design}, it follows that
\begin{equation*}
Q^{\widehat\pi_k}
=
\argmin_{f\in\mathcal H^\beta(L)}
\sup_{g\in\mathcal H^\beta(L_{\cT})}\mathcal L_k(f,g).
\end{equation*}
Indeed, $Q^{\widehat\pi_k}$ belongs to the outer class, and a zero
$L^2(\mu_k)$ residual vanishes everywhere by continuity and
$\mu_k\ge(1-\gamma)\lambda_d$.  Uniqueness then follows from Bellman
contraction.  The subtraction in
\eqref{eq:population-minimax-loss} cancels the common conditional target
variance and avoids the double-sampling bias of a directly squared empirical
Bellman residual \citep{antos2008learning,chen2019information}.  Moment-based
minimax Q-learning is related but uses a different adversarial criterion
\citep{uehara2020minimax}.

At the sample level, the complete observation block is
\begin{equation}
\begin{aligned}
\cD_k
&:=
\{(Z_{k,i},R_{k,i},S_{k,i}^{\prime},A_{k,i}^{\prime}):i=1,\ldots,n\},
\\
Z_{k,i}&\sim\mu_k,
\qquad
S_{k,i}^{\prime}\sim P(\cdot\mid Z_{k,i}),
\qquad
A_{k,i}^{\prime}\sim\widehat\pi_k(\cdot\mid S_{k,i}^{\prime}).
\end{aligned}
\label{eq:one-step-data}
\end{equation}
Each next action is drawn once and held fixed during fitting.  For
$f\in\cF_n^{\rm sig}$, set
\[
Y_{k,i}(f):=R_{k,i}+\gamma f(S_{k,i}^{\prime},A_{k,i}^{\prime})
\]
and, for $g\in\cG_n^{\rm sig}$, define
\begin{equation*}
\widehat{\mathcal L}_k(f,g)
:=\frac1n\sum_{i=1}^n
\left[
\{f(Z_{k,i})-Y_{k,i}(f)\}^2
-\{g(Z_{k,i})-Y_{k,i}(f)\}^2
\right].
\end{equation*}
To control the action-gradient error, we penalize high-order action
derivatives.
Independently of $\cD_k$, draw $\ell_n$ independent uniform
integration points $U_{k,r}\sim\lambda_d$, $r=1,\ldots,\ell_n$.
Write $\mathcal U_k=\{U_{k,r}:r=1,\ldots,\ell_n\}$ for this collection.
These points require neither rewards nor simulator transitions.
For $m=\lceil\beta\rceil$, define the action roughness functional
\[
\mathcal P_m(f)
:=
\sum_{j=1}^{d_a}\int_{[0,1]^d}
|\partial_{a_j}^{m}f(z)|^2\,\lambda_d(\dd z).
\]
Only pure derivatives in the action coordinates are penalized.
Replacing the population function classes by sigmoid DNN classes and
the penalty integral by its empirical average gives
\begin{equation}
\label{eq:sigmoid-minimax}
\widehat Q_{k+1}\in
\argmin_{f\in\cF_n^{\rm sig}}
\left\{
\sup_{g\in\cG_n^{\rm sig}}\widehat{\mathcal L}_k(f,g)
+\frac{\lambda_n}{\ell_n}\sum_{r=1}^{\ell_n}
                  \sum_{j=1}^{d_a}|\partial_{a_j}^{m}f(U_{k,r})|^2
\right\}.
\end{equation}
The corresponding action-gradient estimate is the analytical derivative
$\nabla_a\widehat Q_{k+1}$.  Derivative regularization also appears in
deep Sobolev regression \citep{ding2025sobolev}.  Here the order
$m=\lceil\beta\rceil$ allows a finite-difference argument to control
first-derivative error through Bellman value error and the penalized
roughness.  Neither derivative labels nor boundary conditions on the
unknown value function are required.

The number of integration points is calibrated uniformly over the
trainable class.  For a sufficiently large fixed constant $C_m$
depending only on $m$, set
\[
A_{m,n}
:=
d_a\{C_m(\mathcal W_n+d+1)(\mathcal B_n+1)\}^{2(m\mathcal D_n+1)}.
\]
The chain rule bounds
$\sum_j|\partial_{a_j}^mf(z)|^2$ by $A_{m,n}$ for every admissible
network and every $z$.
Lemma~\ref{lem:penalty-collocation} establishes this uniform derivative bound and controls the
uniform integration error.  Theorem~\ref{thm:critic-l2} specifies
$\lambda_n>0$ and $\ell_n$ so that this error is no larger than the
critic's approximation and statistical errors.

\subsection{Actor: reflected conditional Langevin sampling}

Fix a queried state $s$.  Substituting $F=\widehat Q_{k+1}$ into
\eqref{eq:reflected-langevin-prelim} gives the fitted reflected
diffusion on $\cA=[0,1]^{d_a}$:
\begin{equation*}
\dd A_{k,t}^{s}
=
\nabla_a\widehat Q_{k+1}(s,A_{k,t}^{s})\,\dd t
+
\sqrt{2\tau}\,\dd B_t
+
\dd K_{k,t}^{s}.
\end{equation*}
The regulator $K_{k,t}^{s}$ is defined in \eqref{eq:cube-local-times}.
By the no-flux calculation in Section~\ref{sec:langevin-prelim}, the diffusion's
invariant law is $\widetilde\pi_{k+1}=\cG_\tau(\widehat Q_{k+1})$.
The density of this fitted Gibbs target satisfies
\[
\nabla_a\log\widetilde\pi_{k+1}(a\mid s)
=\tau^{-1}\nabla_a\widehat Q_{k+1}(s,a).
\]
Thus its score differs from that of $\cG_\tau(Q^{\widehat\pi_k})$ by
$\tau^{-1}(\nabla_a\widehat Q_{k+1}-\nabla_aQ^{\widehat\pi_k})$.
This identity links critic-gradient risk to sampling error through
relative Fisher information.  Before defining the finite sampler, we
bound the temperature bias of its Gibbs target.

\begin{lemma}[Soft-greedy gap]
\label{lem:soft-gap}
For every state $s$, bounded $Q$, and policy law $\pi(\cdot\mid s)$ with
Lebesgue density $p_\pi(\cdot\mid s)$, the following identity holds in the
extended-real sense:
\begin{equation}
\label{eq:soft-gap}
\tau\log Z_Q(s)
-
\int_{\cA}
\left\{Q(s,a)-\tau\log p_\pi(a\mid s)\right\}
\pi(\dd a\mid s)
=
\tau\KL\left(
\pi(\cdot\mid s)
\middle\|
\cG_\tau(Q)(\cdot\mid s)
\right).
\end{equation}
If, in addition, $Q(s,\cdot)$ is continuously differentiable and
$\max_j\|\partial_{a_j}Q(s,\cdot)\|_\infty\le L$, then
\begin{equation}
\label{eq:continuous-soft-greedy}
0\le
\max_{a\in\cA}Q(s,a)
-
\int_{\cA}Q(s,a)\,[\cG_\tau(Q)](\dd a\mid s)
\le
\tau\left[1+d_a\log\left(1+\frac{d_aL}{\tau}\right)\right].
\end{equation}
If, in addition, $Q(s,\cdot)$ is concave, this temperature error admits the sharper bound
\begin{equation}
\label{eq:concave-soft-greedy}
0\le
\max_{a\in\cA}Q(s,a)
-
\int_{\cA}Q(s,a)\,[\cG_\tau(Q)](\dd a\mid s)
\le d_a\tau.
\end{equation}
\end{lemma}

For the finite-step actor, let $\nu_{k,0}(\cdot\mid s)$ be the initialization law.  Fix the macrostep size $\eta>0$ and the number of macrosteps $M\in\mathbb N$ at every outer iteration; $M\eta$ is the total simulated Langevin time.  Let $\Gamma_{\cA}$ be the normal Skorokhod map on the cube: for a continuous driving path $x$ with $x(0)\in\cA$, $\Gamma_{\cA}(x)=x+K$ is the unique constrained path whose bounded-variation regulator $K$ acts only in the inward normal direction at the boundary.  Denote by $L_\Gamma$ its fixed uniform-path Lipschitz constant.  Starting from $A_{k,0}^{s,{\rm path}}\sim\nu_{k,0}(\cdot\mid s)$ and using independent Brownian increments $W_{k,j}(u):=B_{j\eta+u}-B_{j\eta}$, $0\le u\le\eta$, the exact pathwise REM comparison process freezes the fitted drift at the beginning of each macrostep and sets
\begin{equation}
\label{eq:ula-actor}
\begin{aligned}
X_{k,j}^{s}(u)
&:=A_{k,j}^{s,{\rm path}}
+u\nabla_a\widehat Q_{k+1}(s,A_{k,j}^{s,{\rm path}})
+\sqrt{2\tau}\,W_{k,j}(u),
\quad 0\le u\le\eta,
\\
A_{k,j+1}^{s,{\rm path}}
&:=\bigl[\Gamma_{\cA}(X_{k,j}^{s})\bigr](\eta),
\quad j=0,\ldots,M-1.
\end{aligned}
\end{equation}
In this pathwise REM step, also called an Euler--Peano step, the drift
is frozen while reflection acts on the entire Brownian path.  The
resulting comparison process is used to establish the entropy estimate
in the sampling analysis.

The executable sampler replaces each Brownian path by its polygonal interpolation.  Choose $0<\delta\le\eta$ such that $q_\delta:=\eta/\delta\in\mathbb N$.  Starting from $A_{k,0}^{s,\delta}\sim\nu_{k,0}(\cdot\mid s)$, set $A_{k,j,0}^{s,\delta}=A_{k,j}^{s,\delta}$ and, for $\ell=0,\ldots,q_\delta-1$, compute
\begin{equation}
\label{eq:microgrid-actor}
\begin{aligned}
A_{k,j,\ell+1}^{s,\delta}
&:=\operatorname{proj}_{\cA}\!\left[
A_{k,j,\ell}^{s,\delta}
+\delta\nabla_a\widehat Q_{k+1}(s,A_{k,j}^{s,\delta})
+\sqrt{2\tau}\{B_{j\eta+(\ell+1)\delta}-B_{j\eta+\ell\delta}\}
\right],\\
A_{k,j+1}^{s,\delta}
&:=A_{k,j,q_\delta}^{s,\delta},
\qquad j=0,\ldots,M-1,
\end{aligned}
\end{equation}
where $\operatorname{proj}_{\cA}$ is coordinatewise Euclidean projection onto the action cube.  The fitted drift remains frozen across the $q_\delta$ microincrements in a macrostep.  Coordinatewise projection evaluates the Skorokhod image of the polygonal driver exactly at its knots.  Thus \eqref{eq:microgrid-actor} is a finite algorithm, whereas \eqref{eq:ula-actor} is its Brownian-path limit.  We use $\delta=0$ to denote the pathwise comparison and $\delta>0$ for
the executable microgrid.

The next policy is defined directly as a probability kernel by
\begin{equation*}
\widehat\pi_{k+1}(B\mid s)
:=
\bP\{A_{k,M}^{s,\delta}\in B\mid\cF_k,\cD_k,\mathcal U_k\},
\qquad B\in\mathcal B(\cA).
\end{equation*}
Independent runs of this microgrid supply next-action draws at update
$k+1$.  Because the finite output law may have boundary mass, we compare
it with the Gibbs target in Wasserstein distance.

\section{Theory}
\label{sec:theory}
The analysis follows the method's three components: critic estimation,
conditional actor sampling, and outer policy iteration.  Critic error
arises from function approximation, finite transition data, and numerical
integration of the high-order penalty.  Actor error combines finite-time
mixing and numerical approximation of the reflected diffusion.  We
first control these errors and then propagate them through policy
iteration.  Under the regularity and
coverage conditions stated below, Bellman policy improvement gives the
decomposition
\begin{align}
\bE\{J(\pi^\star)-J(\widehat\pi_K)\}
\lesssim
&
\underbrace{\frac{\gamma^K}{1-\gamma}}_{\text{outer iteration}}
+\underbrace{\frac{b_\tau}{(1-\gamma)^2}}_{\text{temperature bias}}
\notag\\
&+\frac{1}{1-\gamma}\sum_{k=0}^{K-1}\gamma^{K-1-k}
\Bigl[\underbrace{\bE\,\mathsf w_1(\widehat\pi_{k+1},\widetilde\pi_{k+1})}
_{\text{sampling error for fitted target}}
+\underbrace{\bE\,\mathsf w_1\left(\widetilde\pi_{k+1},
\cG_\tau(Q^{\widehat\pi_k})\right)}
_{\text{critic-induced target error}}
\Bigr].
\label{eq:outer-error-overview}
\end{align}
Here
$
b_\tau:=\tau\left[1+d_a\log\left(1+\frac{d_aL}{\tau}\right)\right]
$
is the soft-greedy bound from Lemma~\ref{lem:soft-gap}, where $L$ is a
uniform bound on each action partial derivative of the population
value functions.  The proof of
Theorem~\ref{thm:annealed-end-to-end}
in Appendix~\ref{app:return-propagation} establishes
\eqref{eq:outer-error-overview}.  Section~\ref{sec:critic} derives the value and
gradient risks underlying the critic-induced term.
Section~\ref{sec:sampling} combines the two critic risks with mixing and
discretization bounds.  Section~\ref{sec:end-to-end} combines this
recursion with a sharper direct value-risk bound on approximate
greediness to obtain the return guarantee.  

\subsection{Joint value and action-gradient estimation}
\label{sec:critic}
We first derive joint value and action-gradient risk bounds for the
regularized neural critic.  To this end,
Assumption~\ref{ass:holder} imposes uniform H\"older regularity on
population values and Bellman images of the neural class, while
Assumption~\ref{ass:design} specifies the covered reset-stationary
sampling experiment.  Under these conditions,
Theorem~\ref{thm:critic-l2} gives nonasymptotic bounds by calibrating
the network architecture, high-order penalty weight, and integration budget
to the sample size.

\begin{assumption}
\label{ass:holder}
There exist $\beta>2$ and fixed constants $L_{\cT}\ge L>0$ such that,
for every stationary Markov policy $\pi$,
$Q^\pi\in\mathcal H^\beta(L)$ and
$
\cT^\pi f\in\mathcal H^\beta(L_{\cT})$ 
for each 
$f\in\cF_n^{\rm sig}.
$
\end{assumption}

\begin{assumption}
\label{ass:design}
Conditional on $\cF_k$, the $n$ transition tuples in
\eqref{eq:one-step-data} are independent and identically distributed.
Let $P^{\widehat\pi_k}$ denote their state--action transition kernel and
let $\lambda_d$ be normalized Lebesgue measure on $[0,1]^d$.  The query law
is
\begin{equation*}
\mu_k=(1-\gamma)\sum_{j=0}^{\infty}
              \gamma^j\lambda_d(P^{\widehat\pi_k})^j.
\end{equation*}
\end{assumption}

\begin{theorem}
\label{thm:critic-l2}
Under Assumptions~\ref{ass:holder} and~\ref{ass:design}, use the trainable
classes in Definition~\ref{def:localized-sigmoid-classes}.
There are architecture bounds satisfying
\begin{equation*}
\mathcal D_n\asymp\log n,\quad
\mathcal W_n\asymp n^{d/(2\beta+d)},\quad
\mathcal M_n\asymp n^{d/(2\beta+d)}\log n,\quad
\mathcal B_n\asymp n^{c_B},
\end{equation*}
where $c_B$ is a fixed constant depending on
$(\beta,d)$.
For $m=\lceil\beta\rceil$, there are positive tuning parameters satisfying
\[
\lambda_n\asymp
n^{-2m/(2\beta+d)}(\log n)^{3m/\beta},
\qquad
\ell_n\asymp
\left\lceil
A_{m,n}\mathcal M_n(\log n)^2
\{n^{-2\beta/(2\beta+d)}(\log n)^3\}^{m/\beta-1}
\right\rceil,
\]
for which the global regularized minimax estimator satisfies
\begin{align}
\bE_k\norm{\widehat Q_{k+1}-Q^{\widehat\pi_k}}_{L^2(\mu_k)}^2
&\lesssim n^{-2\beta/(2\beta+d)}(\log n)^3,
\label{eq:value-l2-rate}\\
\bE_k\norm{\nabla_a\widehat Q_{k+1}
                -\nabla_aQ^{\widehat\pi_k}}_{L^2(\lambda_d)}^2
&\lesssim n^{-2(\beta-1)/(2\beta+d)}
                  (\log n)^{3(\beta-1)/\beta}.
\label{eq:gradient-l2-rate}
\end{align}
\end{theorem}

\begin{remark}
Assumption~\ref{ass:holder} requires smooth true values and Bellman
images of the trainable DNN class, not a fixed H\"older constraint on
fitted networks.  Related Bellman-image smoothness conditions appear
in Theorem~21 and Remark~22 of \citet{feng2023overparameterized} for
optimality updates, and in Assumption~5.1 of \citet{jiao2025deep} for
policy evaluation.  Here the H\"older radius is uniform over the
growing sigmoid classes and the policies used by the actor.
Neural approximation yields approximate Bellman completeness, not
exact closure within the finite network class.
Lemma~\ref{lem:primitive-bellman-regularity}
verifies this regularity when $(s,a)\mapsto\int u\,\mathcal R(\dd u\mid s,a)$ is
H\"older and the transition density is
H\"older in the queried state--action variables as an $L^1$-valued map;
no smoothness of the actor kernel is required.

The design in Assumption~\ref{ass:design} satisfies
$\mu_k\ge(1-\gamma)\lambda_d$ and
$\gamma\mu_kP^{\widehat\pi_k}\le\mu_k$.  These inequalities give
coverage and Bellman coercivity, respectively; bounded rewards
control empirical fluctuations.  The value risk uses $\mu_k$, whereas
the gradient risk and roughness integral use $\lambda_d$, including
when the actor has boundary atoms.
To sample from this design, draw $T$ with
$\Pr(T=j)=(1-\gamma)\gamma^j$, start from $\lambda_d$, and run the
current actor for $T$ transitions.  Independent repetitions yield the
block; only the queried transitions are retained for estimation.
The expected cost per query, including its successor, is
$1/(1-\gamma)$.
\end{remark}

\begin{remark}
The value and action-gradient risk bounds have the classical nonparametric
minimax exponents up to logarithmic factors
\citep{stone1982optimal,chen2022wellposed}.
In Theorem~\ref{thm:critic-l2}, the difference between the exponents in
\eqref{eq:value-l2-rate} and \eqref{eq:gradient-l2-rate} reflects the loss
of one order of smoothness.
With $m=\lceil\beta\rceil$ and $\lambda_n>0$, controlling the integration
error leaves $\lambda_n\mathcal P_m(\widehat Q_{k+1})$ in the basic
inequality.
A comparison resolution of order $\lambda_n^{-1/(2m)}$ makes the
combined approximation and penalty cost of order $\lambda_n^{\beta/m}$.
Using a finite-difference scale of order $\lambda_n^{1/(2m)}$, capped
at a fixed cube scale, balances the terms
\[
\lambda_n^{-1/m}n^{-2\beta/(2\beta+d)}(\log n)^3
\quad\text{and}\quad
\lambda_n^{(\beta-1)/m}.
\]
Balancing these terms gives the stated $\lambda_n$ and
\eqref{eq:gradient-l2-rate}.  The comparison network's $m$th
derivatives may grow with resolution; the target need not have
square-integrable derivatives of that order when $\beta$ is noninteger.
Lemmas~\ref{lem:profiled-risk-curvature}
and~\ref{lem:critic-error-decomposition} give the details.

For comparison, Theorem 2 of \citet{ding2025sobolev} gives a squared
gradient risk of $O\{n^{-s/(d+4s)}(\log n)^2\}$ under a first-order
penalty for integer smoothness $s\ge2$.
For a common integer smoothness $\beta=s>2$, the exponent in
\eqref{eq:gradient-l2-rate} is $2(s-1)/(2s+d)$, the classical
gradient benchmark.  The improved exponent combines high-order bias
calibration with the localized statistical bound; it does not rule out
faster first-order estimators under other conditions.
The penalty controls integrated derivative risk,
whereas uniform drift bounds and strong concavity require separate
architecture and geometric conditions.
The uniform integration guarantee is worst-case over the trainable
class; its derivative-evaluation budget can be quasipolynomial because
$A_{m,n}=\exp\{O((\log n)^2)\}$.
\end{remark}

\subsection{Langevin sampling error with an estimated critic}
\label{sec:sampling}
The actor analysis compares the numerical sampler with the diffusion
associated with the fitted critic, its invariant Gibbs law, and the population Gibbs target.
These comparisons separate numerical error, finite-time mixing, and
critic-estimation error, all integrated over the state cube.  Value and
gradient perturbations provide complementary target-error bounds.
Numerical discretization is controlled separately by bounds on the drift
and its Lipschitz constant supplied by the network architecture.
For the calibrated architecture in Theorem~\ref{thm:critic-l2},
set $B_F=2B_Q$ and define
\begin{align*}
A_n:=\mathcal W_n\mathcal B_n,~~
B_{1,n}:=\sqrt{d_a}A_n^{\mathcal D_n+1},~~L_{b,n}:=d_a(\mathcal D_n+1)A_n^{2(\mathcal D_n+1)}.
\end{align*}
Lemma~\ref{lem:architecture-drift} gives deterministic bounds on the
action-gradient norm and the drift Lipschitz constant for both fitted and
population potentials.
For a common sampling profile, put
\begin{equation*}
\mathcal C_{\tau,n}:=
1+\sqrt{\frac{1+L_{b,n}}{\tau}}
\left[1+
\frac{1+\log\{1+d_a(1+L_{b,n})/\tau\}}
     {\alpha_\tau\tau}\right].
\end{equation*}
This factor bounds the time-integrated Lipschitz seminorm of the exact
diffusion semigroup and depends polynomially on $L_{b,n}$.  If all
fitted potentials are $\rho$-strongly concave,
the results below instead hold with
$\mathcal C_{\tau,n}=1+\rho^{-1}$, by synchronous contraction.

\begin{theorem}
\label{thm:annealed-langevin}
Under Assumptions~\ref{ass:holder} and~\ref{ass:design}, use the critic
calibration in Theorem~\ref{thm:critic-l2}.  Suppose the fitted and population
potentials have the common sampling profile $(\alpha_\tau,\chi_\tau,D_\tau)$,
uniformly in the queried state and iteration.  Initialize from normalized Lebesgue measure and use
the frozen-drift macrostep size $\eta$ and the executable polygonal
projection mesh size $\delta$, where
$0<\eta\le1$,
$0<\delta\le\eta/2$, and $\eta/\delta\in\mathbb N$.  Then
\begin{align*}
\bE_k\mathsf w_1\left(\widehat\pi_{k+1},
                  \cG_\tau(Q^{\widehat\pi_k})\right)
\lesssim&
\mathcal C_{\tau,n}L_{b,n}\left[
 B_{1,n}\eta+\sqrt{\tau\{\eta+\delta\log(e/\delta)\}}\right]
 +\left(\frac{\chi_\tau}{\tau}\right)^{1/2}
 e^{-\alpha_\tau\tau M\eta}\\
&+\min\Biggl\{
 \left(\frac{\chi_\tau\min\{D_\tau,\tau^{-d_a}\}}
                 {\alpha_\tau\tau^2}\right)^{1/2}
 n^{-(\beta-1)/(2\beta+d)}(\log n)^{3(\beta-1)/(2\beta)},\\
&\hspace{21mm}
 \left(\frac{\chi_\tau}{\tau}\right)^{1/2}e^{3B_F/(4\tau)}
 n^{-\beta/(2\beta+d)}(\log n)^{3/2}
 \Biggr\}.
\end{align*}
\end{theorem}

\begin{remark}%
Smoothing and mixing replace $\exp(L_{b,n}M\eta)$ by
$\mathcal C_{\tau,n}$, making the numerical remainder uniform in
$M\eta$.  Short-block propagation removes the reciprocal-macrostep
factor from the mesh error.  With fixed temperature and drift bounds,
this error vanishes as $\eta\downarrow0$ even at a fixed ratio
$\delta/\eta\le1/2$.  The inverse log-Sobolev constant can nevertheless
grow exponentially as $\tau$ decreases, and the derivative bounds for the calibrated networks
can grow as $\exp\{O((\log n)^2)\}$.
Strong concavity gives $\mathcal C_{\tau,n}=1+\rho^{-1}$ by synchronous
contraction, but does not remove the local drift dependence.
The minimum compares gradient and value perturbations of the Gibbs
target.  Reverse relative Fisher information weights the gradient error
by the population Gibbs law, whose $O(\tau^{-d_a})$ density bound follows
from Lipschitz regularity, including for multimodal potentials and
boundary maximizers.  The value route uses relative-entropy interpolation
and yields the sharper sample-size exponent under multimodal logarithmic
cooling.  Thus computing the fitted gradient need not entail using its
statistical risk in the return bound.  Both routes use integrated critic
risks, not uniform-in-state estimation bounds.

\end{remark}

\subsection{End-to-end actor--critic error propagation}
\label{sec:end-to-end}
We next propagate the integrated actor error through policy improvement.
For this purpose, Assumption~\ref{ass:primitive-mdp} imposes continuity
and state-density bounds that allow the error to be transferred to the
state distributions in the recursion.  Combined with Bellman contraction
and the critic and actor bounds, these conditions yield the end-to-end
return bound in Theorem~\ref{thm:annealed-end-to-end}.

\begin{assumption}
\label{ass:primitive-mdp}
The map $(s,a)\mapsto\int u\,\mathcal R(\dd u\mid s,a)$ is continuous.
The transition kernel has a jointly
continuous density $p(s'\mid s,a)$ with respect to $\lambda_{\cS}$ satisfying
$p\le\overline p<\infty$.  The initial-state law has density at most
$\overline\rho<\infty$ with respect to $\lambda_{\cS}$.
\end{assumption}

\begin{theorem}[End-to-end bound]
\label{thm:annealed-end-to-end}
Suppose Assumptions~\ref{ass:holder}, \ref{ass:design}, and
\ref{ass:primitive-mdp} hold.  Use the critic calibration
of Theorem~\ref{thm:critic-l2}.  Suppose the fitted and population
potentials share the sampling profile $(\alpha_\tau,\chi_\tau,D_\tau)$
uniformly in state and iteration.  Use the actor settings of
Theorem~\ref{thm:annealed-langevin}, with the same $n,\tau,\eta,M,\delta$
in all $K$ iterations.  Then, we have 
\begin{align}
&\bE\left[J(\pi^\star)-J(\widehat\pi_K)\right]\notag\\
&\lesssim \frac{\gamma^K}{1-\gamma}
+\frac{1}{(1-\gamma)^2}\Biggl[b_\tau+\mathcal C_{\tau,n}L_{b,n}\left[
B_{1,n}\eta+\sqrt{\tau\{\eta+\delta\log(e/\delta)\}}\right]
+\left(\frac{\chi_\tau}{\tau}\right)^{1/2}
 e^{-\alpha_\tau\tau M\eta}\notag\\
&+\min\Biggl\{
 \left(\frac{\chi_\tau\min\{D_\tau,\tau^{-d_a}\}}
                 {\alpha_\tau\tau^2}\right)^{1/2}
 n^{-(\beta-1)/(2\beta+d)}(\log n)^{3(\beta-1)/(2\beta)},
 e^{3B_F/(2\tau)}n^{-2\beta/(2\beta+d)}(\log n)^3
 \Biggr\}\Biggr].
\label{eq:annealed-error-decomposition}
\end{align}
\end{theorem}

\begin{remark}
Theorem~\ref{thm:annealed-end-to-end} separates outer iteration,
temperature bias, numerical error, mixing, and critic estimation.
For outer propagation, the additional MDP conditions are those in
Assumption~\ref{ass:primitive-mdp}: its state-density bounds transfer
integrated actor error, while the discounted Bellman operator supplies
contraction.  No separate contraction assumption on the Gibbs policy
map is needed.  The numerical
term retains sampling-geometry and architecture dependence.
The displayed discount factors describe outer propagation; the remaining
constants treat $\gamma$ as fixed.
The statistical minimum compares gradient-based transport with squared
value-risk control of approximate greediness, which avoids a square-root
loss.
To make the temperature bias vanish, the temperature and sampling
budgets must be calibrated jointly with the critic risk.
Corollary~\ref{cor:multimodal-end-to-end} uses the bounded-potential
profile to obtain logarithmic cooling for multimodal critics.
Under strong concavity, improved sampling and approximate-greediness
bounds permit faster cooling and the algebraic return rates in
Corollary~\ref{cor:strong-concave-end-to-end}.
\end{remark}

\begin{corollary}[Multimodal logarithmic cooling]
\label{cor:multimodal-end-to-end}
Under the conditions of Theorem~\ref{thm:annealed-end-to-end}, use the bounded-potential
profile and choose a fixed
$c_\tau>3B_F(2\beta+d)/(4\beta)$.  Set
$\tau_n=\min\{1,c_\tau/\log n\}$ and $K\asymp\log n$, and choose
\begin{align*}
M\eta\asymp\frac{\log n}{\alpha_{\tau_n}\tau_n},\qquad
\eta\asymp
\frac{1}{\mathcal C_{\tau_n,n}^{2}(1+L_{b,n})^2
(1+B_{1,n})^2(\log n)^4},\qquad
\delta=\eta/2.
\end{align*}
Then, we have
\begin{equation*}
\bE\left[J(\pi^\star)-J(\widehat\pi_K)\right]
\lesssim\frac{1}{(1-\gamma)^2}\left[
\frac{1+\log(\log n)}{\log n}
+n^{-2\beta/(2\beta+d)+3B_F/(2c_\tau)}(\log n)^3\right].
\end{equation*}
\end{corollary}

\begin{corollary}[Strongly concave return rates]
\label{cor:strong-concave-end-to-end}
Suppose that, almost surely, the fitted critic
$a\mapsto\widehat Q_{k+1}(s,a)$ is $\rho$-strongly concave
and the corresponding population action-value function
$a\mapsto Q^{\widehat\pi_k}(s,a)$ is $2\rho$-strongly concave,
uniformly in $s$ and $k$, where $\rho>0$ is fixed.
Under the other conditions of
Theorem~\ref{thm:annealed-end-to-end}, choose $\tau_n$ of the order of
the right-hand side of \eqref{eq:strong-concave-return-rate} and take
\begin{equation*}
\eta\asymp\frac{\tau_n}{(1+L_{b,n})^2(1+B_{1,n})^2},\qquad
M\eta\asymp K\asymp\log n,\qquad
\delta\asymp\frac{\eta}{(\log n)^2}.
\end{equation*}
Choose the calibration to satisfy
$\gamma^K\le\tau_n$, $e^{-\rho M\eta}\le\tau_n$,
$\tau_n\le1$, $\eta\le1$, and $\delta\le\eta/2$.
Then,
\begin{equation}
\bE\left[J(\pi^\star)-J(\widehat\pi_K)\right]
\lesssim\frac{1}{(1-\gamma)^2}\min\Bigl\{
n^{-\frac{2\beta}{(2\beta+d)(d_a+2)}}
(\log n)^{\frac{3}{d_a+2}},\;
n^{-\frac{2(\beta-1)}{(2\beta+d)(d_a+1)}}
(\log n)^{\frac{3(\beta-1)}{\beta(d_a+1)}}\Bigr\}.
\label{eq:strong-concave-return-rate}
\end{equation}
\end{corollary}

\begin{remark}
Corollary~\ref{cor:multimodal-end-to-end} allows multimodal critics.
The bounded-potential profile deteriorates exponentially as the
temperature decreases; logarithmic cooling accommodates this dependence
while making the temperature bias vanish.  Its calibration uses the
value-risk branch of Theorem~\ref{thm:annealed-end-to-end}, leading to
a logarithmic return rate up to an iterated-logarithmic factor.

Corollary~\ref{cor:strong-concave-end-to-end} obtains algebraic rates
by imposing strong concavity on both the population values and the
fitted critics. Strong concavity sharpens both sampling and approximate-greediness
control, permitting faster cooling.
The two terms in \eqref{eq:strong-concave-return-rate} follow from the
value and gradient risks, respectively.  The value-based rate is faster when
$\beta<d_a+2$, whereas the gradient-based rate is faster when $\beta>d_a+2$;
the rates, including their logarithmic factors, coincide at equality.
The two rates use the same critic and actor with different temperature
calibrations; their derivation is in Appendix~\ref{app:return-propagation}.

\end{remark}

\section{Numerical Experiments}
\label{sec:experiments}

We examine final returns and learning progress in two controlled Markov
environments and compare prediction with fixed-critic policy performance
on six logged continuous-control datasets.  Appendix~\ref{app:energy-utility}
provides a complementary study of decision quality on observed
building designs.
The implementation with first-order action-derivative regularization
is called \emph{sigmoid DNN}.  The controlled simulations use trajectory
blocks and Gaussian reward noise.  
%These experiments assess finite-budget
% performance; they do not establish the high-order regularity,
% covered-reset, or strong-concavity conditions or the asymptotic rates
% of the theory.

\subsection{Interactive actor--critic experiments}
\label{sec:interactive-experiments}

Both interactive studies use 100 independent replications in
environments with fixed reward and transition laws.
For every state $s\in\cS$, we set
\[
\pi_0(\cdot\mid s)=\nu_{k,0}(\cdot\mid s)=\mathrm{Unif}([0,1]^{d_a}),\qquad k=0,\ldots,K-1.
\]
The initial interaction data are collected under $\pi_0$.
Each Langevin actor call starts from a fresh draw from
$\nu_{k,0}(\cdot\mid s)$, independent of its Brownian increments.
Performance is measured by the finite-horizon discounted return
$$
J_{18}(\pi):=\bE_{\rho_0,\pi,P,\mathcal R}\left\{\sum_{t=0}^{17}\gamma^tR_t\right\},
$$
estimated on independent evaluation trajectories using conditional
reward means.  Methods share environment seeds and initial states,
with training and evaluation trajectories kept separate.
Tables report means and Monte Carlo (MC) standard errors (SEs); 95\% MC
intervals use the $t_{99}$ critical value.  Return means and gains are
displayed to two decimals: sigmoid DNN values are truncated, linear
and quadratic means are rounded upward, and uniform means are rounded
to the nearest hundredth.  MC SEs are multiplied by 100 and displayed with a percent
sign, truncated to two decimals; $p\%$ denotes an absolute SE of
$p/100$ return units, not a relative SE.  Interval endpoints are rounded
outward to two decimals, and calculations use unrounded values.
Boldface marks the best displayed mean among implementable methods,
including ties, and does not indicate statistical significance.

Both interactive studies use $\gamma=0.9$ and $\ell_n=2n$
independent uniform integration points.  The sigmoid DNN uses
the first-order variant of \eqref{eq:sigmoid-minimax}, with $m=1$.
Four independent development replications per environment select
$\lambda_n=10^{-5}$ from $\{10^{-5},10^{-4},10^{-3}\}$ by average proportional
return gain over the uniform policy across all updates and both
environments.  This value is fixed for all subsequent updates.
The linear and quadratic critics use the minimax criterion in their
polynomial spans, with a first-order penalty coefficient of $10^{-5}$
and a ridge regularization coefficient of $10^{-8}$.

\begin{figure}[H]
\centering
\includegraphics[width=0.80\textwidth]{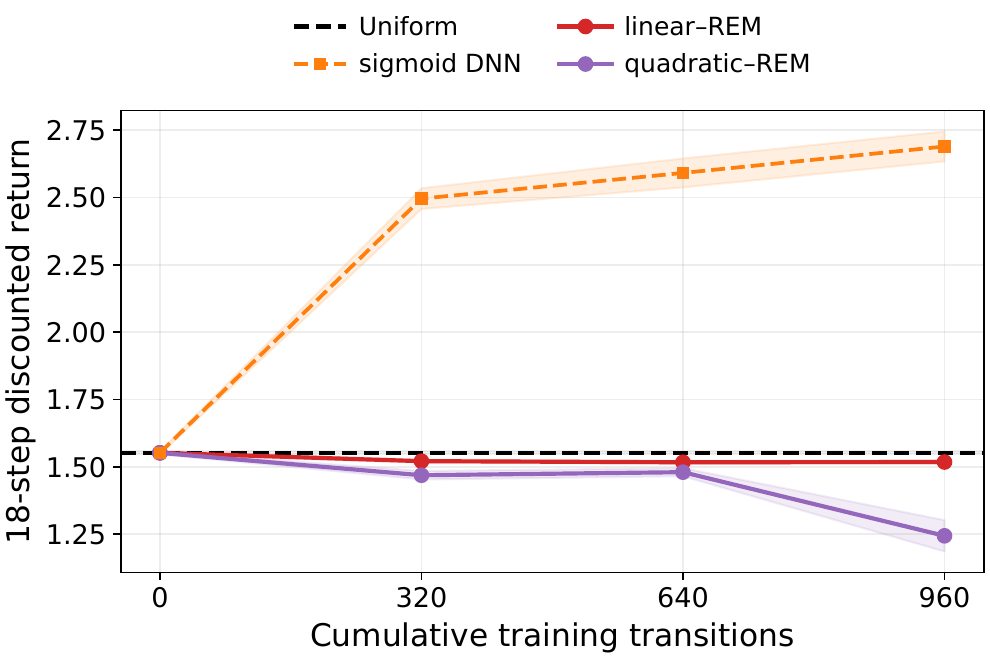}
\caption{Sequential actor--critic return against cumulative training transitions, with pointwise 95\% Monte Carlo bands.}
\label{fig:sequential-ac}
\end{figure}

\paragraph{A sequential controlled Markov environment.}
We consider stochastic state transitions and a reward surface with two
state-dependent action modes.  The state is $S_t=(S_{t,1},S_{t,2})\in[0,1]^2$
and the action is $A_t\in[0,1]$.  With
$x=s_1-0.5$ and $y=s_2-0.5$, the reward and transition laws are
\begin{align*}
\ell(s)
&=
0.06+0.30\sigma_{\rm sig}(5x-3y+2xy),\\
u(s)
&=
0.64+0.30\sigma_{\rm sig}(-3x+5y-2xy),\\
w(s)
&=
0.25+0.50\sigma_{\rm sig}(4x+3y+4xy),\\
R_t
&=
w(S_t)e^{-85\{A_t-\ell(S_t)\}^2}
+\{1-w(S_t)\}e^{-85\{A_t-u(S_t)\}^2}\\
&\hspace{10mm}
-0.06(S_{t,1}-0.72)^2-0.04(S_{t,2}-0.28)^2+0.06\xi_t,\\
S_{t+1,1}
&=
\operatorname{refl}_{[0,1]}\!\left(
0.55S_{t,1}+0.24A_t+0.16S_{t,2}^2+0.025+0.025\epsilon_{t,1}
\right),\\
S_{t+1,2}
&=
\operatorname{refl}_{[0,1]}\!\left(
0.60S_{t,2}+0.18S_{t,1}+0.14(1-A_t)^2+0.035
+0.025\epsilon_{t,2}
\right),
\end{align*}
where $\xi_t,\epsilon_{t,1},\epsilon_{t,2}$ are independent standard normal
variables and $\operatorname{refl}_{[0,1]}$ denotes repeated normal reflection
into the unit interval.
Starting from $S_0\sim\mathrm{Unif}([0,1]^2)$, trajectories follow
the specified reward and transition laws with actions from the current
policy.  We use $\tau=0.055$, $K=3$ updates, and $n=320$ transitions per
update, collected from 32 independent trajectories of length 10.
The reflected actor uses $M=48$ REM macrosteps with $\eta=0.016$
and projection-microgrid step size $\delta=0.004$.
We compare the sigmoid DNN critic with linear and quadratic
critics, all with the same REM actor.
The learned methods use the same number of transitions under their
respective policies.  Each policy is evaluated by $J_{18}$ on 240
independent trajectories.
Table~\ref{tab:sequential-ac} and Figure~\ref{fig:sequential-ac}
summarize final policy performance and learning progress, respectively.
The sigmoid DNN has a higher mean final return than the uniform,
linear-critic, and quadratic-critic policies.  Its paired gain over
Uniform is $1.13$ ($95\%$ MC interval $[1.08,1.20]$), and the gain
from the first fitted update to the final update is $0.19$
($95\%$ MC interval $[0.16,0.23]$).

\begin{table}[H]
\centering
\caption{Sequential Markov-environment performance}
\label{tab:sequential-ac}
\vspace{0.4em}
\small
\setlength{\tabcolsep}{8pt}
\setlength{\extrarowheight}{2pt}
\renewcommand{\arraystretch}{1.10}
\begin{tabular*}{0.96\textwidth}{@{\extracolsep{\fill}}lrrrr@{}}
\toprule
Method & Uniform & sigmoid DNN & linear--REM & quadratic--REM \\
\midrule
\shortstack{Return\\(MC SE)} & \shortstack{1.55\\(0.27\%)} & \shortstack{\textbf{2.68}\\(2.80\%)} & \shortstack{1.52\\(0.29\%)} & \shortstack{1.25\\(2.93\%)} \\
\bottomrule
\end{tabular*}
\end{table}

\paragraph{A high-dimensional controlled Markov environment.}

\begin{figure}[H]
\centering
\includegraphics[width=0.80\textwidth]{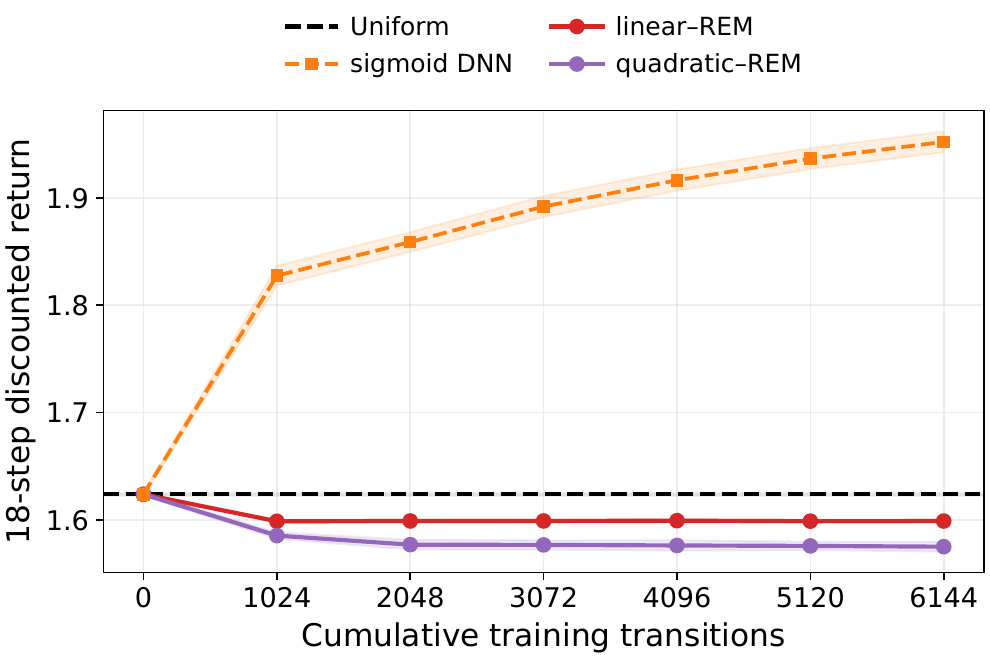}
\caption{Actor--critic return against cumulative training transitions in the structured high-dimensional environment, with pointwise 95\% Monte Carlo bands.}
\label{fig:actor-critic-interaction}
\end{figure}

We next consider a controlled MDP of joint dimension $d=50$:
$
d_s=35,~ d_a=15,~
K=6,~
 n=1024.
$
Initial states are uniform on $[0,1]^{d_s}$.  For
$j=1,\ldots,d_a$, define two action modes and their state-dependent mixing
weight by
\begin{align*}
\ell_j(s)
&=
0.08+0.30\sigma_{\rm sig}\{5x_j-3y_j\},
\\
u_j(s)
&=
0.62+0.30\sigma_{\rm sig}\{-3.5x_j+4y_j\},
\\
w_j(s)
&=
0.35+0.30\sigma_{\rm sig}\{6x_j+4y_j+5x_jy_j\},
\qquad
x_j:=s_j-0.5,\quad y_j:=s_{j+d_a}-0.5.
\end{align*}
The observed reward is
\begin{equation*}
R_t
=
\frac1{d_a}\sum_{j=1}^{d_a}
\left[
w_j(S_t)e^{-85\{A_{t,j}-\ell_j(S_t)\}^2}
+\{1-w_j(S_t)\}e^{-85\{A_{t,j}-u_j(S_t)\}^2}
\right]
+0.08\xi_t.
\end{equation*}
Here $\xi_t$ is standard normal.
For $j=1,\ldots,d_a$, the transition law is
\begin{align*}
S_{t+1,j}
&=\operatorname{refl}_{[0,1]}\!\left(
0.55S_{t,j}+0.24A_{t,j}+0.16S_{t,j+d_a}^2
+0.025+0.025\epsilon_{t,j}\right),\\
S_{t+1,j+d_a}
&=\operatorname{refl}_{[0,1]}\!\left(
0.60S_{t,j+d_a}+0.18S_{t,j}+0.14(1-A_{t,j})^2
+0.035+0.025\epsilon_{t,j+d_a}\right).
\end{align*}
The remaining coordinates satisfy
\[
S_{t+1,j}=\operatorname{refl}_{[0,1]}\!\left(
0.70S_{t,j}+0.15+0.025\epsilon_{t,j}\right),
\qquad j=2d_a+1,\ldots,d_s.
\]
The transition noises are independent standard normal variables and
are independent of the reward noise.  Each training block comprises 128
independent trajectories of length 8 generated by the current actor.
We compare Langevin actors driven by sigmoid, linear, and structured
quadratic critics, with the uniform actor as a no-learning reference.
The structured quadratic includes squared action coordinates and
the interactions $s_ja_j$ and $s_{j+d_a}a_j$.
All learned actors use $\tau=0.008$, $M=50$, $\eta=0.040$, and
$\delta=0.010$.  We evaluate $J_{18}$ on 400 independent trajectories.
Table~\ref{tab:actor-critic-interaction} and Figure~\ref{fig:actor-critic-interaction}
summarize final policy performance and learning progress, respectively.
The sigmoid DNN again has the highest mean final return among the four
methods.  Its paired gain over Uniform is $0.32$ ($95\%$ MC interval
$[0.31,0.34]$), and the gain from the first fitted update to the final
update is $0.12$ ($95\%$ MC interval $[0.12,0.13]$).

\begin{table}[H]
\centering
\caption{Final actor--critic performance in the structured 50-dimensional environment over 100 replications.}
\label{tab:actor-critic-interaction}
\vspace{0.4em}
\small
\setlength{\tabcolsep}{8pt}
\setlength{\extrarowheight}{2pt}
\renewcommand{\arraystretch}{1.10}
\begin{tabular*}{0.96\textwidth}{@{\extracolsep{\fill}}lrrrr@{}}
\toprule
Method & Uniform & sigmoid DNN & linear--REM & quadratic--REM \\
\midrule
\shortstack{Return\\(MC SE)} & \shortstack{1.62\\(0.05\%)} & \shortstack{\textbf{1.95}\\(0.48\%)} & \shortstack{1.60\\(0.05\%)} & \shortstack{1.58\\(0.22\%)} \\
\bottomrule
\end{tabular*}
\end{table}

\subsection{Continuous-control critic prediction and policy extraction}
\label{sec:continuous-control}
We use the medium-replay-v2 and medium-v2 variants of HalfCheetah,
Hopper, and Walker2d from D4RL \citep{fu2020d4rl,kumar2020cql}, giving
six datasets from three environments.  Each dataset has 100 paired
outer splits, with 80\% of complete trajectories assigned to training.
Model selection uses a further group-disjoint partition of the outer
training data.  Replay is a reanalysis of previously examined splits; Medium uses new
prespecified splits.  For a logged trajectory of length $T$, the
prediction target is the observed return-to-go
\[
Y_t=\sum_{u=t}^{T-1}0.99^{u-t}R_u.
\]
For $n_{\rm te}$ held-out observations, let $Y_i$ and $\widehat Y_i$
denote the target and fitted prediction, both standardized using
training-target statistics.  We report root mean squared error (RMSE)
and mean absolute error (MAE):
\begin{align*}
\mathrm{RMSE}&=\left\{n_{\rm te}^{-1}
             \sum_i(\widehat Y_i-Y_i)^2\right\}^{1/2},&
\mathrm{MAE}&=n_{\rm te}^{-1}\sum_i|\widehat Y_i-Y_i|.
\end{align*}
The sigmoid DNN uses supervised squared loss with a first-order
action-derivative penalty coefficient of $10^{-4}$ and 512 uniform integration
points.  We compare it with linear and structured quadratic ridge
regressions.  Inner trajectory-held-out RMSE selects the neural
architecture and learning rate, or the baseline ridge coefficient;
the selected model is then refitted on the complete outer training sample.

Each fitted critic defines a fixed REM policy, evaluated for five
episodes in the corresponding MuJoCo-v5 environment without further
training.  Episodes stop at termination or truncation, with at most
1,000 steps, and evaluation random streams are paired across methods.
For split $j$ and method $h$, define
\[
\overline G_{jh}=\frac15\sum_{e=1}^{5}
\sum_{t=0}^{T_{jhe}-1}0.99^t R_{jhet},\qquad
Z_{jh}=\frac{\overline G_{jh}-\mu_j}{\sigma_j},
\]
where $\mu_j$ and $\sigma_j$ are the outer-training target mean and
standard deviation, shared by the three methods.  Thus $Z_{jh}$
standardizes the complete discounted return, not individual rewards.
We use this training-target score to compare policies within each
dataset; Appendix~\ref{app:d4rl-details} gives the evaluation,
aggregation, and interpretation details.

Tables~\ref{tab:d4rl-data} and \ref{tab:d4rl-policy-z} report mean
$\pm$ sample standard deviation over 100 paired splits.  The dispersions
describe partition variability and, for policy returns, simulation
randomness; they are not confidence intervals.  Smaller prediction
errors and larger standardized discounted returns are better.
Means and standard deviations are displayed to two decimal places, with sigmoid DNN
entries truncated toward zero and Linear and Quadratic entries rounded toward
$+\infty$.  At this displayed precision, the sigmoid DNN has lower
mean RMSE and MAE than Linear on all six datasets.  Against Quadratic,
each prediction metric has two lower means, two ties, and two higher
means.  For standardized discounted return, the sigmoid DNN leads on
HalfCheetah Medium and Walker2d Medium and ties with Linear on their
Replay variants; Quadratic leads on both Hopper datasets.  Hopper
Medium therefore illustrates the different prediction and policy
rankings: the sigmoid DNN has the lowest prediction errors, whereas
Quadratic has the highest standardized return.  Boldface marks the
best displayed mean, including ties; these descriptive comparisons
do not establish statistical significance or equality of underlying
means.

\begin{table}[H]
\centering
\small
\setlength{\tabcolsep}{5pt}
\renewcommand{\arraystretch}{0.90}
\caption{Trajectory-held-out prediction on six D4RL datasets.}
\label{tab:d4rl-data}
\begin{tabular}{lccc}
\toprule
Dataset & sigmoid DNN & Linear & Quadratic \\
\midrule
\multicolumn{4}{l}{\textit{RMSE}} \\
HalfCheetah Replay & $\mathbf{0.55}\pm0.01$ & $0.58\pm0.02$ & $0.56\pm0.02$ \\
Hopper Replay & $0.71\pm0.02$ & $0.80\pm0.02$ & $\mathbf{0.69}\pm0.02$ \\
Walker2d Replay & $0.67\pm0.01$ & $0.72\pm0.02$ & $\mathbf{0.66}\pm0.02$ \\
HalfCheetah Medium & $\mathbf{0.94}\pm0.02$ & $0.96\pm0.04$ & $\mathbf{0.94}\pm0.03$ \\
Hopper Medium & $\mathbf{0.34}\pm0.02$ & $0.54\pm0.01$ & $0.41\pm0.01$ \\
Walker2d Medium & $\mathbf{0.62}\pm0.01$ & $0.68\pm0.02$ & $\mathbf{0.62}\pm0.02$ \\
\midrule
\multicolumn{4}{l}{\textit{MAE}} \\
HalfCheetah Replay & $\mathbf{0.41}\pm0.01$ & $0.44\pm0.02$ & $0.42\pm0.02$ \\
Hopper Replay & $0.55\pm0.01$ & $0.65\pm0.02$ & $\mathbf{0.54}\pm0.02$ \\
Walker2d Replay & $\mathbf{0.53}\pm0.01$ & $0.59\pm0.02$ & $\mathbf{0.53}\pm0.02$ \\
HalfCheetah Medium & $\mathbf{0.64}\pm0.01$ & $0.66\pm0.02$ & $\mathbf{0.64}\pm0.02$ \\
Hopper Medium & $\mathbf{0.26}\pm0.02$ & $0.43\pm0.01$ & $0.31\pm0.01$ \\
Walker2d Medium & $0.41\pm0.00$ & $0.47\pm0.02$ & $\mathbf{0.40}\pm0.01$ \\
\bottomrule
\end{tabular}
\end{table}

\begin{table}[H]
\centering
\small
\setlength{\tabcolsep}{5pt}
\renewcommand{\arraystretch}{0.90}
\caption{Training-target standardized discounted return in MuJoCo-v5.}
\label{tab:d4rl-policy-z}
\begin{tabular}{lccc}
\toprule
Dataset & sigmoid DNN & Linear & Quadratic \\
\midrule
HalfCheetah Replay & $\mathbf{-1.72}\pm0.06$ & $\mathbf{-1.72}\pm0.08$ & $-1.73\pm0.08$ \\
Hopper Replay & $-1.79\pm0.07$ & $-1.83\pm0.04$ & $\mathbf{-1.77}\pm0.07$ \\
Walker2d Replay & $\mathbf{-1.69}\pm0.03$ & $\mathbf{-1.69}\pm0.04$ & $-1.70\pm0.04$ \\
HalfCheetah Medium & $\mathbf{-4.26}\pm0.09$ & $-4.27\pm0.09$ & $-4.27\pm0.10$ \\
Hopper Medium & $-3.53\pm0.06$ & $-3.52\pm0.05$ & $\mathbf{-3.47}\pm0.09$ \\
Walker2d Medium & $\mathbf{-2.81}\pm0.05$ & $-2.84\pm0.06$ & $-2.82\pm0.06$ \\
\bottomrule
\end{tabular}
\end{table}

\section{Conclusion}
We develop  a regularized minimax Bellman critic and a finite reflected
Langevin actor for policy iteration with continuous states and actions.
Under the stated regularity and coverage conditions, a positive
high-order action-derivative penalty, together with Bellman stability
and interpolation, controls value and gradient risks.  The resulting bounds yield
complementary sampling and ordinary-return bounds that separate critic
estimation, finite-time mixing, discretization, temperature bias, and
outer-iteration errors.  Joint calibration yields consistency for bounded
multimodal potentials and faster algebraic rates when both population
values and fitted critics are strongly concave.  The first-order
implementation has higher mean final returns than the tested uniform
and polynomial-critic controls in two Markov environments.  
In the logged-data study, the sigmoid DNN has lower displayed mean prediction
errors than Linear, while comparisons with Quadratic and rankings by
training-target standardized discounted return depend on the dataset.
The building-energy study also shows lower mean heating
load and regret for neural-critic decisions than for Linear and Uniform.
Extending the analysis to dependent trajectories, linking logged-return
prediction to policy evaluation under distribution shift, and reducing
multimodal sampling costs remain directions for future work.

\section*{Appendix}
This appendix is organized as follows.
Appendix~\ref{app:related-work} provides the detailed discussion of
related work.  Appendix~\ref{app:algorithm} gives the pseudocode for
the actor--critic recursion.
Appendix~\ref{app:policy-iteration} proves the Gibbs and policy-improvement
identities.  Appendix~\ref{app:critic} establishes the joint value and
action-gradient estimation bounds in Section~\ref{sec:critic}.
Appendix~\ref{app:sampling} proves the Langevin sampling bound with an
estimated critic in Section~\ref{sec:sampling}.
Appendix~\ref{app:return-propagation} proves the end-to-end actor--critic
bounds and temperature calibrations in Section~\ref{sec:end-to-end}.
Appendix~\ref{app:auxiliary} collects the supporting results from the literature and
derives the transport consequence used in the sampling analysis.
Appendix~\ref{app:experiments} collects continuous-control evaluation
details and the building-energy study in
Sections~\ref{app:d4rl-details} and~\ref{app:energy-utility}, respectively.
\appendix
\section{Related work}
\label{app:related-work}

\paragraph{Actor--critic convergence and finite-sample theory.}
Two-time-scale actor--critic theory establishes convergence by allowing
the critic to track a more slowly updated actor \citep{konda2000actor};
natural actor--critic uses compatible approximation and the geometry
of the policy parameterization \citep{bhatnagar2009natural}.
Finite-sample analyses quantify the interaction between these updates
with Markovian observations.  \citet{wu2020finite} study
two-time-scale actor--critic, and \citet{xu2020improving} analyze
mini-batch actor--critic and natural actor--critic.
\citet{khodadadian2021offpolicy} develop off-policy natural
actor--critic with explicit control of estimation error.
Single-loop analyses further address critic errors that evolve with
the policy \citep{chen2023single,wang2024singleloop}, while
\citet{chen2025approximate} study approximate policy improvement with
off-policy temporal-difference evaluation.
Neural actor--critic analyses establish last-iterate convergence on
continuous state--action spaces \citep{gaur2024closing} and improve
accuracy dependence using data dropping \citep{ganesh2025order}.
With a linear critic, \citet{chen2025continuous} analyze single-timescale
updates in continuous state--action spaces; \citet{chen2026deep}
study average-reward actor--critic with deep neural approximation.
These works analyze specific policy-update mechanisms and observation
models.  We study how function and derivative estimation errors
provide alternative bounds for a conditional Langevin actor and
repeated approximate-greedy updates.

\paragraph{Neural Bellman estimation and action-gradient risk.}
Smooth nonparametric regression provides the basic relation between
sample size, dimension, and regularity \citep{stone1982optimal}.
For fixed-policy off-policy evaluation, \citet{chen2022wellposed}
establish Bellman well-posedness and rate-optimal estimation of
action-value functions and their derivatives using an
instrumental-variable formulation.  Their statistical exponents
provide a benchmark for our critic rates; our contribution concerns
the regularized neural Bellman estimator, its finite-point derivative
penalty, and propagation through successive finite-sampler policies.
The covered reset design and uniform Bellman regularity used here
differ from their fixed-policy offline observation model.
Neural regression theory links
such statistical rates to network approximation and complexity
\citep{schmidthieber2020regression,farrell2021deep}.
Smooth-activation approximation includes constructive rates for smooth
and analytic functions \citep{mhaskar1996optimal} and quantitative
deep sigmoid constructions \citep{langer2021sigmoid}.
Simultaneous derivative approximation is studied by
\citet{hornik1990derivatives} and, quantitatively for hyperbolic
tangent networks, by \citet{deryck2021tanh}.
For Bellman estimation, subtractive squared-loss criteria address the
conditional-variance obstruction in residual minimization
\citep{antos2008learning,chen2019information}, and minimax Q-function
learning controls adversarial Bellman moments \citep{uehara2020minimax}.
Deep Sobolev regression uses a fitted-gradient penalty and derives
function and plug-in derivative rates without derivative responses
\citep{ding2025sobolev}.  Its derivative-rate analysis assumes a
homogeneous Neumann condition on the target (Assumption 4); the
semi-supervised analysis with a sampled penalty additionally imposes
fixed uniform bounds on first derivatives in the fitted class (Assumption 6).
Our one-sided finite-difference argument retains a high-order
penalty and does not require either condition.
High-order derivative penalties are classical
in smoothing splines \citep{cox1984multivariate}.  We use a high-order
action penalty in a neural Bellman game and bound its finite-point
approximation error.  Bellman coercivity controls the function error,
while the retained penalty energy controls the derivative error;
neither step assumes a fixed H\"older radius for the fitted network.
In RL, model-based action-gradient estimation (MAGE) uses differentiable
dynamics \citep{doro2020useful}, and first-order Sobolev RL constructs
derivative supervision from differentiated Bellman backups
\citep{schramm2025sobolev}.  Critic gradients also guide direct
on-policy actor updates \citep{gurumurthy2023critic}.
Our critic analysis handles the candidate-dependent Bellman response and
both fitted sigmoid classes jointly, including noninteger target
smoothness.  The reset experiment then yields conditional
risks for the successive learned policies; it is a simulator-based
observation model, rather than a relaxation of general offline coverage
conditions.

\paragraph{Generative policies and conditional sampling.}
Generative actors expand the range of action distributions available
to RL algorithms.  Diffusion Q-learning combines offline behavior
cloning with value maximization \citep{wang2023diffusionql}.
Diffusion actor--critic with entropy regulator (DACER) uses estimated
entropy to regulate exploration \citep{wang2024dacer}, whereas
Q-weighted variational policy optimization (QVPO) constructs a
value-weighted diffusion objective \citep{ding2024qvpo}.
Q-score matching (QSM) connects the critic's action gradient to
off-policy policy-score estimation \citep{psenka2024qscore}.
Reweighted score matching supports diffusion policy mirror descent
and soft diffusion actor--critic \citep{ma2025efficient}.
Flow Q-learning (FQL) trains a value-maximizing one-step actor
regularized by distillation from a behavior-cloned flow policy
\citep{park2025flowq}; reverse flow matching (RFM) combines value and
gradient information through a posterior-mean representation
\citep{li2026rfm}.
\citet{ki2026direct} propose Langevin Q-learning (LQL), which is
particularly close to our actor construction and generates actions directly
from critic-gradient dynamics and fits ordinary Bellman
temporal-difference targets.  Their noise-conditioned extension
smooths the value landscape to improve multimodal exploration.
This work provides an algorithmic precedent for direct
critic-driven sampling.  Our analysis establishes return guarantees
that account for statistical error in a derivative-regularized neural
critic and finite reflected sampling.
Wasserstein policy gradient (WPG) analyzes entropy-regularized
distribution-space updates with exact soft critics \citep{zhu2026wpg},
and Wasserstein proximal policy gradient (WPPG) studies related
entropy-regularized proximal updates under prescribed soft-critic
value and action-gradient errors \citep{zhu2026wppg}.
Reflected stochastic differential equations and their Euler
approximations provide the boundary framework
\citep{lions1984reflected,slominski2001euler}; constrained Langevin
analysis highlights the importance of treating projection and mixing
separately \citep{bubeck2018projected}.
\citet{lamperski2021projected} establishes nonasymptotic
$W_1$ guarantees for nonconvex projected Langevin sampling on compact
convex domains.  \citet{zheng2022constrained} obtain a time-uniform
$W_1$ algorithm-to-diffusion error of order
$\{h\log(h^{-1})\}^{1/2}$ for a step size $h$ on polyhedral domains,
under their Lipschitz, moment, and stationary $L$-mixing assumptions
(Theorem 1 and Lemma 3).  The strong-convexity condition in that work for
sufficiently separated points is vacuous when the separation threshold
exceeds the diameter of a compact domain; its scope therefore already
includes multimodal potentials on a cube.
Our discretization analysis separates the sampling profile from polynomial
and logarithmic dependence on the learned drift bounds.  For bounded
potentials, the exponential temperature dependence is determined by
their amplitude; the displayed constants in
\citet{zheng2022constrained} also couple inverse temperature to the
drift-Lipschitz scale.  We connect this explicit profile to data-derived
critic risks and ordinary return for finite-sampler policies.
Their algorithm projects after each gradient step, whereas ours freezes
the drift over a macrostep and uses a separate projection microgrid.
Accordingly, time-uniform control is not new in itself, and the present
comparison does not establish a uniformly better projection complexity.

Table~\ref{tab:comparison} summarizes the closest statistical and
sampling comparisons; the works differ in their observation models and
policy objectives.  Here, \emph{assumed} means that a critic
error is prescribed as an oracle condition rather than derived from
data.

\begin{table}[H]
\centering
\caption{Representative theoretical coverage.}
\label{tab:comparison}
\footnotesize
\setlength{\tabcolsep}{4pt}
\renewcommand{\arraystretch}{1.14}
\begin{tabularx}{\textwidth}{@{}>{\raggedright\arraybackslash}p{0.23\textwidth}>{\raggedright\arraybackslash}p{0.23\textwidth}>{\raggedright\arraybackslash}p{0.27\textwidth}>{\raggedright\arraybackslash}X@{}}
\toprule
Work & Critic information & Sampling and numerical layer & RL guarantees \\
\midrule
Finite-time actor--critic \citep{wu2020finite,chen2025approximate,chen2026deep}
& Data-derived value error; no separate action-gradient risk bound
& No iterative action-space sampling layer
& Finite-sample actor--critic convergence \\
Nonparametric policy evaluation \citep{chen2022wellposed}
& Data-derived value and derivative rates
& No conditional action sampler
& Fixed-policy off-policy evaluation \\
LQL and noise-conditioned LQL \citep{ki2026direct}
& Neural temporal-difference critic; learned action gradients
& Direct Langevin actions; noise-conditioned smoothing
& Algorithmic construction and empirical policy performance \\
WPG \citep{zhu2026wpg}
& Exact soft critic and its action gradient
& Langevin evolution with discretization control
& Entropy-regularized convergence with an exact critic \\
WPPG \citep{zhu2026wppg}
& Soft-critic value and gradient errors assumed
& Wasserstein proximal update; heat-step implementation
& Proximal-update convergence under prescribed critic errors \\
Projected Langevin \citep{lamperski2021projected,zheng2022constrained}
& Prescribed potential; exact or stochastic gradients
& Nonconvex constrained sampling; time-uniform control in the latter work
& Sampling and optimization, not neural Bellman policy iteration \\
Present paper
& Integrated $Q$ and $\nabla_aQ$ risks under covered resets
& Finite reflected actor; learned drift and horizon-uniform numerical control
& Ordinary-return bounds with multimodal and strong-concavity calibrations \\
\bottomrule
\end{tabularx}
\end{table}

\paragraph{Empirical benchmarks and polynomial controls.}
D4RL provides logged datasets with differing behavior-policy quality
and coverage \citep{fu2020d4rl,kumar2020cql}.  We use its medium and
medium-replay variants for supervised return prediction and fixed-critic
policy extraction.  Linear and structured quadratic ridge regressions
are supervised baselines, separate from the offline RL algorithms
reviewed above.  The Energy Efficiency data of \citet{tsanas2012energy}
record simulated outcomes for 64 feasible designs per shell, allowing
decisions to be evaluated on that observed design set.

\section{Actor--critic algorithm}
\label{app:algorithm}

Algorithm~\ref{alg:sac-sampling} summarizes the recursion in
Section~\ref{sec:algorithm}, using the notation and update definitions
given there.

\begingroup
\small
\begin{algorithm}[H]
\caption{Regularized minimax critic with a reflected Langevin actor}
\label{alg:sac-sampling}
\begin{algorithmic}[1]
\Require Initial policy sampler $\widehat\pi_0=\pi_0$;
 number of policy updates $K$; transitions per update $n$;
 action-derivative order $m=\lceil\beta\rceil$,
 penalty coefficient $\lambda_n>0$, and number of uniform integration points $\ell_n$;
 sampling temperature $\tau$; number of Langevin macrosteps $M$,
 macrostep size $\eta$, and projection-microgrid step size $\delta$;
 initial action laws $\nu_{k,0}(\cdot\mid s)$.
\Ensure Final actor sampler $\widehat\pi_K$ and fitted critics $\{\widehat Q_{k+1}\}_{k=0}^{K-1}$, with action gradients obtained by differentiation.
\For{$k=0,1,\ldots,K-1$}
  \State Generate $Z_{k,i}=(S_{k,i},A_{k,i})\sim\mu_k$; query $(R_{k,i},S_{k,i}^{\prime})$, and independently draw $A_{k,i}^{\prime}\sim\widehat\pi_k(\cdot\mid S_{k,i}^{\prime})$ once for each transition, as in \eqref{eq:one-step-data}.
  \State Independently generate $U_{k,1},\ldots,U_{k,\ell_n}\sim\lambda_d$ for the action-derivative penalty.
  \State Compute $\widehat Q_{k+1}$ using the regularized minimax sigmoid-DNN criterion in \eqref{eq:sigmoid-minimax}.
  \State Define the fitted Gibbs target $\widetilde\pi_{k+1}=\cG_\tau(\widehat Q_{k+1})$ implicitly, differentiate the fitted critic, and use $\nabla_a\widehat Q_{k+1}$ as the actor drift.
  \For{each state $s$ at which an action is requested}
    \State Initialize $A_{k,0}^{s,\delta}\sim\nu_{k,0}(\cdot\mid s)$ and run the projection microgrid \eqref{eq:microgrid-actor} for $M$ reflected macrosteps.
    \State Return $A_{k,M}^{s,\delta}$ with conditional law $\widehat\pi_{k+1}(\cdot\mid s)$; this sampler is used at iteration $k+1$.
  \EndFor
\EndFor
\end{algorithmic}
\end{algorithm}
\endgroup

\section{Proofs for the Gibbs and Policy-Iteration Results}
\label{app:policy-iteration}

The Bellman equation and the definition of $V^\pi$ in
Section~\ref{sec:setup} give
\begin{equation*}
Q^\pi(s,a)-V^\pi(s)
=\bE_{P,\mathcal R}\{R_t+\gamma V^\pi(S_{t+1})\mid S_t=s,A_t=a\}
-V^\pi(s).
\end{equation*}
Taking expectations along a trajectory generated by $\pi'$ and
telescoping the discounted value terms yields
\begin{align*}
J(\pi')-J(\pi)
&=
\sum_{t=0}^{\infty}\gamma^t
\bE_{\rho_0,\pi',P}
\left[
Q^\pi(S_t,A_t)-
\int_{\cA}Q^\pi(S_t,a)\pi(\dd a\mid S_t)
\right].
\end{align*}
The definition of $d_{\pi'}^{\rho_0}$ then gives
\eqref{eq:performance-difference}.

\begin{proof}[Proof of Lemma~\ref{lem:soft-gap}]
Fix $s$, write $p_\pi$ for the Lebesgue density of $\pi(\cdot\mid s)$, and abbreviate $\pi_Q=\cG_\tau(Q)(\cdot\mid s)$.  Since the density of $\pi_Q$ satisfies
\[
\log p_{\pi_Q}(a)
=
Q(s,a)/\tau-\log Z_Q(s),
\]
we have
\begin{align*}
\tau\KL(\pi\|\pi_Q)
&=
\tau\int p_\pi(a)\log p_\pi(a)\dd a
-
\int Q(s,a)\pi(\dd a\mid s)
+
\tau\log Z_Q(s),
\end{align*}
which proves \eqref{eq:soft-gap}.

For the general temperature bound, let
$m=\max_aQ(s,a)$ and choose a maximizer $a_\star$.  Since the action cube has
unit volume, the differential entropy of $\pi_Q$ is at most zero; the Gibbs
variational identity therefore gives
$\int Q\,\dd\pi_Q\ge\tau\log Z_Q(s)$.  Put
$r=\min\{1,\tau/(d_aL)\}$, with $r=1$ when $L=0$.  The intersection of
the $\ell_\infty$-ball of radius $r$ about $a_\star$ with $[0,1]^{d_a}$ has
volume at least $r^{d_a}$, and every point in it satisfies
$Q(s,a)\ge m-d_aLr$.  Consequently,
\[
Z_Q(s)\ge r^{d_a}\exp\{(m-d_aLr)/\tau\},
\]
and hence
\[
m-\int Q\,\dd\pi_Q
\le d_aLr+\tau d_a\log(1/r)
 \le\tau\left[1+d_a\log\left(1+
 \frac{d_aL}{\tau}\right)\right].
 \]
This proves \eqref{eq:continuous-soft-greedy}.

For concave $Q$, put $U(a)=m-Q(s,a)$, so $U$ is convex and nonnegative
and has a minimizer $a_\star$ with $U(a_\star)=0$.  Convexity gives
\[
U(a)\le\inner{\nabla U(a)}{a-a_\star}.
\]
Let $q_\tau(a)=e^{-U(a)/\tau}/\int_{\cA}e^{-U(u)/\tau}\dd u$.
Integration by parts on the cube gives
\begin{align*}
\int_{\cA}\inner{\nabla U(a)}{a-a_\star}q_\tau(a)\dd a
&=
d_a\tau
-
\tau\int_{\partial\cA}
\inner{a-a_\star}{n_{\rm out}(a)}q_\tau(a)\dd\sigma(a)\\
&\le d_a\tau.
\end{align*}
Here $\dd\sigma$ denotes surface measure.  The boundary integral is
nonnegative because the cube is convex and
$a_\star\in\cA$.  Since $q_\tau$ is also the density of $\cG_\tau(Q)$,
the last display proves \eqref{eq:concave-soft-greedy}.
\end{proof}

\section{Proofs for Joint Value and Action-Gradient Estimation}
\label{app:critic}
\begin{lemma}%[Primitive conditions for uniform Bellman regularity]
\label{lem:primitive-bellman-regularity}
Write $\beta=r_\beta+\nu$ with $r_\beta=\lceil\beta\rceil-1$ and $0<\nu\le1$.
Suppose $z\mapsto\int u\,\mathcal R(\dd u\mid z)$ belongs to
$\mathcal H^\beta(L_R)$ and
$P(\dd s'\mid z)=p(s'\mid z)\zeta(\dd s')$ for a fixed measure $\zeta$.
Assume the map $z\mapsto p(\cdot\mid z)$ is $r_\beta$ times continuously
differentiable as a map into $L^1(\zeta)$ and satisfies
\begin{align*}
&\max_{|\alpha|\le r_\beta}\sup_z
 \norm{D_z^\alpha p(\cdot\mid z)}_{L^1(\zeta)}+
\max_{|\alpha|=r_\beta}\sup_{z\ne y}
\frac{\norm{D_z^\alpha p(\cdot\mid z)
           -D_z^\alpha p(\cdot\mid y)}_{L^1(\zeta)}}{\norm{z-y}_2^\nu}
\le L_P.
\end{align*}
Then, for every measurable stationary actor $\pi$,
\begin{equation*}
\norm{Q^\pi}_{\mathcal H^\beta}\le L_R+\gamma B_QL_P,
\qquad
\norm{\cT^\pi f}_{\mathcal H^\beta}
\le L_R+2\gamma B_QL_P
\quad(\norm f_\infty\le2B_Q).
\end{equation*}
In particular, Assumption~\ref{ass:holder} holds with radii independent of
the actor, the architecture, and its fitted parameters.
\end{lemma}
\begin{proof}
For bounded measurable $f$, define
$m_f(s')=\int f(s',a')\pi(\dd a'\mid s')$.
It is measurable and satisfies $\norm{m_f}_\infty\le\norm f_\infty$,
including when $\pi$ has atoms.  Integration against $m_f$ is a bounded
linear functional on $L^1(\zeta)$ and commutes with the assumed derivatives:
\begin{equation*}
D_z^\alpha(P^\pi f)(z)
=\int D_z^\alpha p(s'\mid z)m_f(s')\,\zeta(\dd s'),
\qquad |\alpha|\le r_\beta.
\end{equation*}
Taking absolute values bounds each derivative by the corresponding
$L^1(\zeta)$ norm times $\norm f_\infty$.  Subtracting the expressions at
$z,y$ gives the corresponding bound for the top-order H\"older seminorm.  Hence
$\norm{P^\pi f}_{\mathcal H^\beta}\le L_P\norm f_\infty$.
Using
\[
(\cT^\pi f)(z)=\int u\,\mathcal R(\dd u\mid z)+\gamma(P^\pi f)(z)
\]
gives $\norm{\cT^\pi f}_{\mathcal H^\beta}\le L_R+\gamma L_P\norm f_\infty$.
Apply this inequality first to $f=Q^\pi=\cT^\pi Q^\pi$ and then to functions
satisfying $\norm f_\infty\le2B_Q$ to obtain both assertions.
\end{proof}

\subsection{Simultaneous approximation by trainable sigmoid DNN classes}

We construct simultaneous value and derivative approximants using
logistic partitions, Taylor polynomials, and sigmoid arithmetic, with
controlled roughness and polynomially bounded parameters.
Related polynomial constructions appear in
\citet{langer2021sigmoid,deryck2021tanh}.

For a function on a fixed compact interval or cube, use
$\|u\|_{C^m}:=\max_{|\alpha|\le m}\|D^\alpha u\|_\infty$,
where the maximum ranges over all multi-indices of order at most $m$.

\begin{lemma}%[Smooth logistic arithmetic]
\label{lem:logistic-gates}
Fix an integer $m\ge1$ and a bounded input interval.  For every
$0<\epsilon\le1$, sigmoid networks with a fixed number of units implement
approximations to the identity and multiplication whose $C^m$ errors are
at most $C\epsilon$ and whose weights are bounded by a fixed power of
$\epsilon^{-1}$.  The constants depend only on $m$ and the input interval.
\end{lemma}
\begin{proof}
Write $\sigma=\sigma_{\rm sig}$ and take the input interval to be $[-R,R]$,
where $R\ge1$ is fixed.  Since
$\sigma'=\sigma(1-\sigma)$ and
$\sigma''=\sigma(1-\sigma)(1-2\sigma)$, the point $t_0=1$ satisfies
$\sigma'(t_0)\sigma''(t_0)\ne0$.  For $0<u\le1$, define
\begin{equation*}
I_u(x)=\frac{\sigma(t_0+ux)-\sigma(t_0)}{u\sigma'(t_0)},\qquad
S_u(x)=\frac{\sigma(t_0+ux)+\sigma(t_0-ux)-2\sigma(t_0)}
                     {u^2\sigma''(t_0)}.
\end{equation*}
All derivatives of $\sigma$ needed below are bounded on
$[t_0-2R,t_0+2R]$.  Taylor's formula gives
$|I_u(x)-x|\lesssim u$ and
$|I_u'(x)-1|\lesssim u$, uniformly on $[-2R,2R]$.
For $j\ge2$, the exact identity
\[
I_u^{(j)}(x)=
\frac{u^{j-1}\sigma^{(j)}(t_0+ux)}{\sigma'(t_0)}
\]
gives $|I_u^{(j)}(x)|\lesssim u$.  Consequently,
$\|I_u-\mathrm{id}\|_{C^m([-2R,2R])}\lesssim u$.

For the square gate, cancellation of the odd Taylor terms gives
$|S_u(x)-x^2|\lesssim u^2$.  The first two derivatives are
\begin{equation*}
S_u'(x)
=\frac{\sigma'(t_0+ux)-\sigma'(t_0-ux)}
        {u\sigma''(t_0)},\qquad
S_u''(x)
=\frac{\sigma''(t_0+ux)+\sigma''(t_0-ux)}
        {\sigma''(t_0)}.
\end{equation*}
Expanding the numerators about $t_0$ therefore gives
$|S_u'(x)-2x|+|S_u''(x)-2|\lesssim u^2$.
For $j\ge3$,
\[
S_u^{(j)}(x)=
\frac{u^{j-2}}{\sigma''(t_0)}
\{\sigma^{(j)}(t_0+ux)+(-1)^j\sigma^{(j)}(t_0-ux)\}.
\]
If $j=3$, the difference in braces is $O(u)$ by the mean-value
theorem; if $j\ge4$, the factor $u^{j-2}$ is at most $u^2$.
These estimates prove
$\|S_u-(x\mapsto x^2)\|_{C^m([-2R,2R])}\lesssim u^2$.

Define the two-input gate
\[
P_u(x,y)=\tfrac14\{S_u(x+y)-S_u(x-y)\}.
\]
The polarization identity
$\{(x+y)^2-(x-y)^2\}/4=xy$ and differentiation of these affine
arguments give
$\|P_u-((x,y)\mapsto xy)\|_{C^m([-R,R]^2)}\lesssim u^2$.
The identity gate uses one sigmoid unit, and the multiplication gate uses four;
both have an affine output.  Choose $u=c\epsilon$, with a fixed
sufficiently small $c>0$, to make both $C^m$ errors at most $\epsilon$.
Their weights and biases are bounded by $C\epsilon^{-2}$.

For an affine readout
$v=a^\top h+b$ of a sigmoid hidden layer, a unit in the next gate has the form
$\sigma(t+\omega v)=\sigma(\omega a^\top h+t+\omega b)$.
Thus affine readouts are absorbed into the next preactivation, and
$I_u$ transports signals to later layers with its stated error.
\end{proof}

\begin{lemma}%[H\"older approximation with controlled derivatives]
\label{lem:sigmoid-dnn}
Let $\beta=r+\nu>2$, where $r=\lceil\beta\rceil-1$ and $0<\nu\le1$.
For $q\in\mathcal H^\beta(L)$ and every integer $J\ge2$, there is a sigmoid
DNN $q_J$ such that, for $0\le j\le r$,
\begin{equation}
\max_{|\alpha|=j}\norm{D^\alpha(q_J-q)}_\infty
\lesssim J^{-(\beta-j)},
\qquad
\norm{q_J}_{\mathcal H^\beta}\lesssim L.
\label{eq:sigmoid-simultaneous-approx}
\end{equation}
It has width at most $CJ^d$, any prescribed depth
$\mathcal D\ge D_0(\beta,d)$, size at most $CJ^d\mathcal D$, and parameter
magnitudes at most $C\{J(\mathcal D+1)\}^{c_0}$ for fixed $C,c_0$.
If $\norm q_\infty\le2B_Q$, the approximant may also be chosen with
$\norm{q_J}_\infty\le2B_Q$.  The constants depend only on the displayed
structural radii and dimensions.
In particular,
\begin{equation*}
\norm{\nabla_a^2q_J-\nabla_a^2q}_{\mathrm{op},\infty}
\lesssim J^{-(\beta-2)}.
\end{equation*}
For $m=\lceil\beta\rceil$ the same network satisfies
\[
\max_{|\alpha|=m}\|D^\alpha q_J\|_\infty\lesssim J^{m-\beta},
\qquad
\mathcal P_m(q_J)\lesssim J^{2(m-\beta)}.
\]
\end{lemma}
\begin{proof}
If $L=0$, the zero network proves the assertion.  Otherwise, construct an
approximant for $q/L$ and multiply its final affine output by $L$.
We may therefore work with $L=1$ until the last step.  Constants in
Steps 1--5 depend only on $(\beta,d)$, not on $q$, $J$, or the prescribed
depth.  Put $h=J^{-1}$ and $m=r+1$.

\paragraph{Step 1: an exact smooth partition and its derivative moments.}
Write $\sigma=\sigma_{\rm sig}$ and $c_j=(j+1/2)h$.  On $[0,1]$, set
\begin{align*}
w_0(x)&=1-\sigma((x-h)/h),\\
w_j(x)&=\sigma((x-jh)/h)-\sigma((x-(j+1)h)/h),
                    &&1\le j\le J-2,\\
w_{J-1}(x)&=\sigma((x-(J-1)h)/h).
\end{align*}
The middle family is empty when $J=2$.  Monotonicity of $\sigma$ gives
$w_j\ge0$, and telescoping gives $\sum_{j=0}^{J-1}w_j=1$ exactly.
For localization and derivative bounds, use
\[
\sigma(t)\le e^t\ (t\le0),\qquad
1-\sigma(t)\le e^{-t}\ (t\ge0),\qquad
\sigma^{(l)}(t)=\sigma(t)\{1-\sigma(t)\}p_l(\sigma(t)),\quad l\ge1,
\]
where $p_l$ is a polynomial depending only on $l$.  The last identity
follows by induction using $\sigma'=\sigma(1-\sigma)$ and gives
$|\sigma^{(l)}(t)|\lesssim e^{-|t|}$.
For an interior $w_j$ and $l=0$, use
$w_j(x)\le\sigma((x-jh)/h)$ to the left of its cell and
$w_j(x)\le1-\sigma((x-(j+1)h)/h)$ to the right.
Inside the cell, $w_j\le1$ and $|x-c_j|\le h/2$.
For $l\ge1$, differentiate the two sigmoid terms and use
\[
\left|\frac{x-jh}{h}\right|,
\left|\frac{x-(j+1)h}{h}\right|
\ge \frac{|x-c_j|}{h}-\frac12.
\]
The same argument applies to the endpoint weights on $[0,1]$:
their nondecaying tails lie outside this interval.  It follows that
\[
|w_j^{(l)}(x)|\lesssim h^{-l}e^{-|x-c_j|/h},
\qquad 0\le l\le m,\quad x\in[0,1].
\]

For $v\in\{0,\ldots,J-1\}^d$, define
$c_v=(c_{v_1},\ldots,c_{v_d})$ and
$w_v(x)=\prod_{i=1}^d w_{v_i}(x_i)$.  Then $\sum_vw_v=1$ and
\[
|D^\alpha w_v(x)|
\lesssim h^{-|\alpha|}e^{-\|x-c_v\|_1/h}.
\]
Here $\|\cdot\|_1$ is the sum of the absolute values of the coordinates.  For $t\ge0$,
divide the centers into shells
$l\le\|x-c_v\|_1/h<l+1$, $l=0,1,\ldots$.
Each shell contains at most $C_d(l+2)^d$ centers, as can be seen by counting
the grid points in the enclosing coordinate box.  Hence
\begin{align}
\sum_v |D^\alpha w_v(x)|\,\|x-c_v\|_2^t
&\lesssim h^{t-|\alpha|}
 \sum_{l=0}^\infty(l+2)^{d+t}e^{-l}
\lesssim h^{t-|\alpha|},
\qquad |\alpha|\le m.
\label{eq:logistic-partition-moment}
\end{align}
\paragraph{Step 2: simultaneous Taylor approximation.}
For a center $c$, let
\[
T_c(x)=\sum_{|\rho|\le r}\frac{D^\rho q(c)}{\rho!}(x-c)^\rho,
\qquad
q_h(x)=\sum_vw_v(x)T_{c_v}(x).
\]
Here $\rho!=\prod_i\rho_i!$ and
$(x-c)^\rho=\prod_i(x_i-c_i)^{\rho_i}$.
Fix $|\zeta|\le r$,
put $p=r-|\zeta|$, and let $g=D^\zeta q$.
When $p\ge1$, the integral Taylor formula along the segment from $c$
to $x$ gives
\begin{align*}
D^\zeta(q-T_c)(x)
={}&p\sum_{|\rho|=p}\frac{(x-c)^\rho}{\rho!}
 \int_0^1(1-t)^{p-1} \times
 \{D^\rho g(c+t(x-c))-D^\rho g(c)\}\,\dd t .
\end{align*}
The segment remains in the cube, where the order-$r$ derivatives are
$\nu$-H\"older.  Hence
\[
|D^\zeta(q-T_c)(x)|\lesssim\|x-c\|_2^{p+\nu}
=\|x-c\|_2^{\beta-|\zeta|}.
\]
When $p=0$, this is exactly the H\"older condition on $D^\zeta q$.

Using $\sum_vw_v=1$, the product rule now gives, for $|\alpha|\le r$,
\begin{align*}
|D^\alpha(q_h-q)(x)|
&\le\sum_{\zeta\le\alpha}\binom{\alpha}{\zeta}
 \sum_v |D^{\alpha-\zeta}w_v(x)|
             |D^\zeta(T_{c_v}-q)(x)|\\
&\lesssim
 \sum_{\zeta\le\alpha}
 h^{\beta-|\zeta|-|\alpha-\zeta|}
\lesssim h^{\beta-|\alpha|}.
\end{align*}
The second inequality follows from \eqref{eq:logistic-partition-moment} with
$t=\beta-|\zeta|$.

\paragraph{Step 3: highest-order derivative control.}
The penalty uses derivatives of order $m=r+1$, which the target need
not possess.  To bound these derivatives of the comparison function,
fix $x_0$ and regard $T_{x_0}(y)$ as a polynomial
in $y$ with its coefficients held fixed.  Since
\[
q_h(y)-T_{x_0}(y)
=\sum_v w_v(y)\{T_{c_v}(y)-T_{x_0}(y)\},
\]
differentiation to order $|\alpha|=r+1$ at $y=x_0$ gives
\begin{align*}
D^\alpha q_h(x_0)
={}&\sum_v\sum_{\substack{\zeta\le\alpha\\|\zeta|\le r}}
 \binom{\alpha}{\zeta}
 D^{\alpha-\zeta}w_v(x_0)
 \{D^\zeta T_{c_v}(x_0)-D^\zeta q(x_0)\}.
\end{align*}
Derivatives of order $r+1$ of both Taylor polynomials vanish.
Step 2 and \eqref{eq:logistic-partition-moment} imply
\[
\max_{|\alpha|=r+1}\|D^\alpha q_h\|_\infty
\lesssim h^{\beta-r-1}=h^{\nu-1}.
\]

For $|\alpha|=r$, write $\rho_0=\|x-y\|_2$.
If $\rho_0\le h$, the mean-value theorem on the segment in the cube gives
\[
|D^\alpha q_h(x)-D^\alpha q_h(y)|
\lesssim h^{\nu-1}\rho_0\le\rho_0^\nu.
\]
If $\rho_0>h$, insert $D^\alpha q$ at the two endpoints and use Step 2:
\[
|D^\alpha q_h(x)-D^\alpha q_h(y)|
\lesssim h^\nu+\rho_0^\nu\lesssim\rho_0^\nu.
\]
Lower-order derivatives are bounded by Step 2 and
$\|q\|_{\mathcal H^\beta}\le1$.
Consequently, $\|q_h\|_{\mathcal H^\beta}\lesssim1$.

\paragraph{Step 4: a pure sigmoid realization and its error budget.}
To realize $q_h$ with sigmoid gates, expand
\[
q_h(x)=
\sum_v\sum_{|\rho|\le r}
\frac{D^\rho q(c_v)}{\rho!}
\left\{\prod_{i=1}^d w_{v_i}(x_i)\right\}
\left\{\prod_{i=1}^d(x_i-c_{v_i})^{\rho_i}\right\}.
\]
There are $\binom{r+d}{d}$ monomials per center, a number independent
of $J$.  Each summand is a product of at most $d+r$ factors, all in
$[-1,1]$, times a coefficient bounded by one.  Append constant factors
equal to one to give each branch exactly $d+r$ factors.

The layered implementation begins with a hidden layer that computes the
sigmoid terms defining each $w_{v_i}$ and identity approximations to the
coordinates $x_i$; subtracting $c_{v_i}$ is an affine readout.
Subsequent layers successively multiply the accumulated product by the
next factor.  Every unused factor is transported to the next layer by
an approximate identity.  The four-unit multiplication and one-unit
identity gates are those of Lemma~\ref{lem:logistic-gates}, each with
$C^m$ error at most $\epsilon$ on a fixed enlarged input interval.
This uses
$
D_0=d+r
$
hidden layers: one initial layer and $d+r-1$ multiplication layers.
The final affine output sums the monomial branches; intermediate affine
readouts are absorbed as in Lemma~\ref{lem:logistic-gates}.

For the $C^m$ implementation error, every exact live signal is a factor,
a partial product of at most $d+r$ factors, or a bounded affine
combination of such signals.  The product rule and Step 1 show
\[
\max_{\text{exact live signals }u}\|u\|_{C^m}\lesssim h^{-m}.
\]
Indeed, each term differentiates the factors to a total order at most
$m$, so the powers of $h^{-1}$ multiply to at most $h^{-m}$.

The product and chain rules give
\begin{align*}
\|uv\|_{C^m}&\lesssim\|u\|_{C^m}\|v\|_{C^m},\\
\|E(u_1,\ldots,u_b)\|_{C^m}
&\lesssim\|E\|_{C^m}
        (1+\max_i\|u_i\|_{C^m})^m ,
\end{align*}
where $b$ is fixed and the values of the $u_i$ lie in the domain of $E$.
For the second inequality, a derivative of order $1\le|\alpha|\le m$
is a finite sum of terms
\[
(\partial_{i_1}\cdots\partial_{i_j}E)(u)
 \prod_{\ell=1}^j D^{\alpha_\ell}u_{i_\ell},
\quad
\sum_{\ell=1}^j\alpha_\ell=\alpha,\quad
|\alpha_\ell|\ge1,\quad j\le|\alpha|;
\]
the order-zero term is bounded by $\|E\|_\infty$.

Let $e_\ell$ be the largest $C^m$ error of a live signal after hidden
layer $\ell$, taking the maximum over all branches.
At the first layer, $e_1\lesssim\epsilon$.
As long as $e_\ell\le1$, approximate signals have $C^m$ norm at most
$C h^{-m}$ and their values remain in a fixed enlargement of $[-1,1]$.
For a multiplication, write $\widetilde u,\widetilde v$ for the
approximate inputs and $P_\epsilon$ for the chosen gate.  The identity
\[
P_\epsilon(\widetilde u,\widetilde v)-uv
=(P_\epsilon-\mathrm{mult})(\widetilde u,\widetilde v)
 +(\widetilde u-u)\widetilde v+u(\widetilde v-v),
\qquad \mathrm{mult}(x,y)=xy,
\]
gives
\[
e_{\ell+1}\le C h^{-m}e_\ell+C\epsilon h^{-m^2}.
\]
Identity transports satisfy the same bound by writing
$I_\epsilon(\widetilde u)-u
=(I_\epsilon-\mathrm{id})(\widetilde u)+\widetilde u-u$.
Iterating the displayed recurrence for at most $D_0$ layers yields
\[
e_\ell\lesssim\epsilon h^{-m^2-mD_0},\qquad 1\le\ell\le D_0.
\]
Set
\[
\epsilon=c\,h^{\beta+d+m^2+mD_0+2},
\]
with a sufficiently small fixed $c>0$.
This choice closes the induction.  Summing the $J^d\binom{r+d}{d}$ branches
produces a base network $q_J^{(0)}$ with
\[
\|q_J^{(0)}-q_h\|_{C^m}
\lesssim J^d\epsilon h^{-m^2-mD_0}
\lesssim h^{\beta+2}.
\]

Each branch uses a fixed number of live signals and sigmoid units per
layer, with zero cross-connections between distinct branches.
The resulting sparse feed-forward network has width and
nonzero-parameter size at most $CJ^d$ and depth $D_0$.
Partition weights and biases are bounded by $CJ$; arithmetic-gate
parameters are bounded by $C\epsilon^{-2}$.
Absorbing affine readouts only multiplies coefficients from two
adjacent gates, while summing branch biases contributes at most a factor
$CJ^d$.  For an explicit bound, set
$a_0=\beta+d+m^2+mD_0+2$, the exponent in the chosen gate accuracy.
Then all base-network parameters are bounded by
$C J^{d+1+4a_0}$; the factor $J$ covers the partition preactivations.

\paragraph{Step 5: prescribed depth with controlled transport error.}
If $\mathcal D=D_0$, use the base network.
Otherwise let $p=\mathcal D-D_0$ and append $p$ one-unit identity gates
to its scalar output.

Choose a fixed interval $[-R,R]$ containing the range of $q_J^{(0)}$
with a margin of two; such an $R$ exists by Steps 2 and 4.
Let $I$ be a gate satisfying
$\|I-\mathrm{id}\|_{C^m([-R,R])}\le\epsilon_I$.
On $[-R+1,R-1]$, set $F_0=\mathrm{id}$ and $F_{\ell+1}=I\circ F_\ell$.
If $\|F_\ell-\mathrm{id}\|_{C^m}\le1/2$, its range is inside $[-R,R]$
and its $C^m$ norm is bounded by a fixed constant.  The chain-rule estimate
in Step 4 then gives
\[
\|F_{\ell+1}-F_\ell\|_{C^m}
=\|(I-\mathrm{id})\circ F_\ell\|_{C^m}
\le C\epsilon_I.
\]
Induction and summation show
$\|F_p-\mathrm{id}\|_{C^m}\le Cp\epsilon_I$ whenever
$Cp\epsilon_I\le1/2$.

The base network satisfies $\|q_J^{(0)}\|_{C^m}\lesssim h^{-m}$ by
Steps 3 and 4.  Another application of the chain rule gives
\[
\|F_p\circ q_J^{(0)}-q_J^{(0)}\|_{C^m}
\lesssim p\epsilon_I h^{-m^2}.
\]
Take
$
\epsilon_I=\frac{c\,h^{\beta+m^2+2}}{(\mathcal D+1)^2}
$
with $c$ sufficiently small.  This choice satisfies the induction condition
and gives a transport error at most $C h^{\beta+2}$ for every prescribed
$\mathcal D\ge D_0$.
The resulting network has exactly $\mathcal D$ hidden layers, width
at most $CJ^d$, and size at most
$C(J^d+\mathcal D)\le CJ^d\mathcal D$.
Its gate parameters are bounded by a fixed power of
$J(\mathcal D+1)$.  In the layered representation, each additional
identity combines only adjacent affine coefficients, rather than
multiplying their bounds through all $\mathcal D$ layers.
In particular, the base-network bound and the gate bound give the
admissible upper bound
\[
C J^{d+1+4a_0}\epsilon_I^{-4}
\lesssim J^{d+1+4a_0+4(\beta+m^2+2)}(\mathcal D+1)^8
\le C\{J(\mathcal D+1)\}^{c_0},
\]
where one may take
$c_0=d+9+4a_0+4(\beta+m^2+2)$.

\paragraph{Step 6: derivative, H\"older, and output bounds.}
Denote the network after Step 5 by $q_J$.  Steps 2, 4, and 5 give
\[
\|D^\alpha(q_J-q)\|_\infty
\le\|q_J-q_h\|_{C^m}+\|D^\alpha(q_h-q)\|_\infty
\lesssim h^{\beta-|\alpha|},\qquad |\alpha|\le r.
\]
For a $C^{r+1}$ function $u$ on the cube, the mean-value theorem gives
\[
\sup_{x\ne y}
\frac{|D^\alpha u(x)-D^\alpha u(y)|}{\|x-y\|_2^\nu}
\le \sqrt d\,(\sqrt d)^{1-\nu}
       \max_{|\zeta|=r+1}\|D^\zeta u\|_\infty,
\qquad |\alpha|=r.
\]
It follows that $\|u\|_{\mathcal H^\beta}\lesssim\|u\|_{C^{r+1}}$.
Apply this to $u=q_J-q_h$, whose $C^{r+1}$ norm is
$O(h^{\beta+2})$, and combine it with Step 3.
Restoring the factor $L$ proves
$\|q_J\|_{\mathcal H^\beta}\lesssim L$ and
\eqref{eq:sigmoid-simultaneous-approx}.
The derivative bound needed for the penalty follows from Steps 3--5:
\[
\max_{|\alpha|=m}\|D^\alpha q_J\|_\infty
\le
\max_{|\alpha|=m}\|D^\alpha q_h\|_\infty
+\|q_J-q_h\|_{C^m}
\lesssim h^{\beta-m}.
\]

Finally, suppose $\|q\|_\infty\le B$, where $B=2B_Q>0$.
If necessary, rescale the affine output by
$
a=\min\{1,B/\|q_J\|_\infty\},
$
with $a=1$ when the network is zero.
If $a<1$, then
\[
1-a=\frac{\|q_J\|_\infty-B}{\|q_J\|_\infty}
\le\frac{\|q_J-q\|_\infty}{B}.
\]
For every $|\alpha|=j\le r$,
\begin{align*}
\|D^\alpha(aq_J-q)\|_\infty
&\le\|D^\alpha(q_J-q)\|_\infty
 +(1-a)\|D^\alpha q_J\|_\infty\\
&\lesssim h^{\beta-j}+h^\beta
\lesssim h^{\beta-j}.
\end{align*}
Replace $q_J$ by the rescaled network.  This enforces the output bound and
decreases its H\"older norm, output-parameter magnitudes, and $m$th
derivative norms.  Since $\lambda_d$ is a
probability measure, the preceding bound proves
$\mathcal P_m(q_J)\le d_a\max_j\|\partial_{a_j}^mq_J\|_\infty^2
\lesssim h^{2(\beta-m)}$.
For the action Hessian, each entry has error at most
$C h^{\beta-2}$, so
\[
\|\nabla_a^2q_J-\nabla_a^2q\|_{\mathrm{op},\infty}
\le d_a\max_{i,j}
       \|\partial_{a_i}\partial_{a_j}(q_J-q)\|_\infty
\lesssim h^{\beta-2}.
\]
\end{proof}

\subsection{Network complexity and population excess risk}

We combine parameter covering and penalty-integration bounds with
Bellman coercivity and a finite-difference inequality.

\begin{lemma}%[Extension to the population comparison class]
\label{lem:population-bellman-extension}
Under Assumption~\ref{ass:holder} and the architecture calibration of
Theorem~\ref{thm:critic-l2}, the fixed auxiliary radius $L_{\cT}$ may
be enlarged so that $\cT^\pi f\in\mathcal H^\beta(L_{\cT})$ for every
$f\in\mathcal H^\beta(L)$ and every stationary policy $\pi$.
\end{lemma}
\begin{proof}
Fix $f$ and $\pi$, and set $c=\min\{1,2B_Q/L\}$, so that
$\norm{cf}_\infty\le2B_Q$.  Apply Lemma~\ref{lem:sigmoid-dnn} to $cf$;
the lemma is independent of Assumption~\ref{ass:holder}.  Take
$J_n=\lfloor\log n\rfloor$ and prescribed depth $\mathcal D_n$.
For every sufficiently large $n$, the resulting width, size, and
parameter bounds are respectively
$O((\log n)^d)$, $O((\log n)^{d+1})$, and
$O((\log n)^{2c_0})$, and therefore satisfy the calibrated bounds.
The version of the lemma that preserves the output bound gives $f_n\in\cF_n^{\rm sig}$ with
$\norm{f_n-cf}_\infty\lesssim J_n^{-\beta}\to0$.
Consequently,
$\norm{\cT^\pi f_n-\cT^\pi(cf)}_\infty
\le\gamma\norm{f_n-cf}_\infty\to0$, while all
$\cT^\pi f_n$ lie in the same radius-$L_{\cT}$ H\"older ball.
To see that this ball is closed under uniform limits, write
$\beta=r_\beta+\nu$.  The derivatives of its elements through order $r_\beta$ are
uniformly bounded and equicontinuous, so a common Arzel\`a--Ascoli
subsequence converges uniformly at every derivative order.
The fundamental theorem of calculus along coordinate segments
identifies these limits as the derivatives of the limiting function.
The derivative sup norms pass to the limit, and the top-order
H\"older seminorm is lower semicontinuous, preserving the bound
$L_{\cT}$, including when $\nu=1$.  Thus the uniform limit
$\cT^\pi(cf)$ belongs to $\mathcal H^\beta(L_{\cT})$.
Since the zero network is admissible, the same bound holds for
$\cT^\pi0$.  The affine structure of the Bellman operator gives
\[
\cT^\pi f=c^{-1}\cT^\pi(cf)+(1-c^{-1})\cT^\pi0,
\qquad
\norm{\cT^\pi f}_{\mathcal H^\beta}\le(2/c-1)L_{\cT}.
\]
Enlarge $L_{\cT}$ by this fixed factor and retain the same notation.
\end{proof}

\begin{lemma}%[Covering all trainable parameters]
\label{lem:full-dnn-cover}
For either class in Definition~\ref{def:localized-sigmoid-classes}, let
$\mathcal N_\infty(\epsilon)$ be its uniform-norm covering number.
For $0<\epsilon<1$, with architecture bounds at least two,
\begin{align*}
\log\mathcal N_\infty(\epsilon)
\lesssim\mathcal M\Big[
 &(\mathcal D+1)\log\{(\mathcal W+1)(\mathcal B+1)\}\\
 &+\log\{(\mathcal D+1)/\epsilon\}
 +\log\{(\mathcal D+1)(\mathcal W+d+1)^2\}\Big].
\end{align*}
Under the calibration in Theorem~\ref{thm:critic-l2}, this is at most
$C\mathcal M_n(\log n)^2$ at $\epsilon=n^{-3}$.
\end{lemma}
\begin{proof}
Write $\sigma=\sigma_{\rm sig}$.  For counting purposes, embed the
parameter locations of all admissible architectures in a common array.
The number of possible locations is at most
$P=C(\mathcal D+1)(\mathcal W+d+1)^2$; summing over supports of size at
most $\mathcal M$ contributes at most
$\exp\{C\mathcal M\log(P+1)\}$.
Fix one support and two parameter vectors at distance $u$ in the maximum norm.
Because $0<\sigma<1$ and $\norm{\sigma'}_\infty\le1/4$, the maximum
hidden-coordinate discrepancy $e_l$ obeys
\begin{equation*}
e_l\le\tfrac14\mathcal W\mathcal B e_{l-1}
          +\tfrac14(\mathcal W+d+1)u,\qquad e_0=0.
\end{equation*}
The output discrepancy is at most
$\mathcal W\mathcal B e_{\mathcal D}+(\mathcal W+1)u$.
Consequently, the parameter-to-function map is uniformly Lipschitz with
constant at most
\begin{equation*}
L_\Theta=C(\mathcal D+1)(\mathcal W+d+1)
              \{(\mathcal W+d+1)(\mathcal B+1)\}^{\mathcal D+1}.
\end{equation*}
Cover the support's parameter cube by a grid of mesh
$\epsilon/(2L_\Theta)$.  For each grid box meeting the constrained parameter
set, choose one point from that intersection.  The selected networks belong
to the constrained class and form an $\epsilon$-net with at most
$(1+4\mathcal B L_\Theta/\epsilon)^{\mathcal M}$ elements.  Combining
this count with the support count proves the claim.  Shallower architectures
contribute an additional finite architecture factor covered by $P$.
\end{proof}

\begin{lemma}%[Architecture bounds and uniform integration of the penalty]
\label{lem:penalty-collocation}
For the classes in Definition~\ref{def:localized-sigmoid-classes}, the
constant $C_m$ in $A_{m,n}$ can be chosen depending only on $m$ so that
\[
\sum_{j=1}^{d_a}|\partial_{a_j}^mf(z)|^2\le A_{m,n}
\]
for every candidate network and every $z$.  For $\ell$ independent
uniform points, there is a nonnegative measurable random variable
$Z_{k,\ell}^{\rm pen}$ such that, simultaneously for all candidates,
\[
\mathcal P_m(f)
\le\frac2\ell\sum_{r=1}^{\ell}\sum_{j=1}^{d_a}
             |\partial_{a_j}^mf(U_{k,r})|^2+Z_{k,\ell}^{\rm pen},
\]
and
\begin{align*}
\bE_k Z_{k,\ell}^{\rm pen}
\lesssim\frac{A_{m,n}\mathcal M_n}{\ell}
\Big[
&(\mathcal D_n+1)
 \log\{(\mathcal W_n+d+1)(\mathcal B_n+1)\}+\log(\ell+1)
\Big].
\end{align*}
\end{lemma}
\begin{proof}
Abbreviate the architecture bounds by
$(\mathcal W,\mathcal D,\mathcal M,\mathcal B)$, put
$K=(\mathcal W+d+1)(\mathcal B+1)$, and set $T=C_mK$.
All sigmoid derivatives through order $m+1$ are bounded: this follows
inductively from $\sigma'=\sigma(1-\sigma)$ and $0<\sigma<1$.

For a hidden coordinate in layer $l$, induction gives
\[
\max_{|\alpha|=j}\|D_a^\alpha h_l\|_\infty\le T^{jl},
\qquad 1\le j\le m.
\]
For the induction step, note that a derivative of order $j$ of
$\sigma(v)$ is a sum over partitions of the $j$ differentiations.
A partition with $p$ blocks contributes a bounded sigmoid derivative
times $p$ derivatives of the preactivation, whose orders sum to $j$.
Each preactivation derivative is bounded by $K$ times the corresponding
preceding-layer derivative.  The contribution is therefore bounded by
$K^pT^{j(l-1)}$.  The finite number of terms depends only on $j$,
and $p\le j$; choosing $C_m$ larger than their coefficient sums for
$j\le m+1$ bounds their sum by $T^{jl}$.  The input coordinates start
the induction with first derivatives at most one and higher derivatives
zero.  The final affine layer gives
\[
\max_{|\alpha|=j}\|D_a^\alpha f\|_\infty\le T^{j\mathcal D+1}.
\]
In particular, with $S=T^{m\mathcal D+1}$,
$p_f(z):=\sum_j|\partial_{a_j}^mf(z)|^2$ is bounded by
$d_aS^2=A_{m,n}$.

To construct a parameter cover of $p_f$, consider the line segment joining
two vectors on a fixed sparsity pattern at maximum-norm distance $u$.
Differentiate once along this segment and at most $j\le m$ times in the action coordinates.
Induction using the same partition formula gives
\[
\max_{|\alpha|=j}
\|\partial_tD_a^\alpha h_l\|_\infty
\le u\,l\,T^{(j+1)l},\qquad 0\le j\le m.
\]
Indeed, differentiating a preactivation also produces the term from
the derivative of its weights.  Its bound is
$KuT^{j(l-1)}$, while the inherited term is bounded by
$Ku(l-1)T^{(j+1)(l-1)}$.
In a differentiated composition, one partition block contains the
segment derivative, and the others contain only action derivatives.
Their combined order is $j+1$ and their number is at most $j+1$,
so the preceding bounds and the choice of $C_m$ close the induction.
The affine output and integration over the segment imply
\[
\max_j\|\partial_{a_j}^mf_\theta-
             \partial_{a_j}^mf_{\theta'}\|_\infty
\le u\,(\mathcal D+1)T^{(m+1)(\mathcal D+1)}.
\]
Denote the coefficient of $u$ by $L_m$.
Consequently, $p_f$ is Lipschitz in the parameters with constant
$2d_aSL_m$.  Counting supports as in
Lemma~\ref{lem:full-dnn-cover} and selecting actual class members from
parameter grid boxes give
\[
\log\mathcal N_\infty(\epsilon,\{p_f\})
\lesssim\mathcal M
\log\!\left[
\{(\mathcal D+1)(\mathcal W+d+1)^2+1\}
\{1+8\mathcal B d_aSL_m/\epsilon\}
\right].
\]
At $\epsilon=A_{m,n}/\ell$, the right-hand side is bounded by the
bracketed architecture expression in the lemma times $C\mathcal M$.
Indeed, $d_aSL_m/A_{m,n}=L_m/S$ and
$L_m/S=(\mathcal D+1)T^{\mathcal D+m}$.

Write $P$ for uniform integration and $P_\ell$ for
the empirical average.  Each fixed nonnegative $p_f\le A_{m,n}$ satisfies
$\operatorname{Var}(p_f(U))\le A_{m,n}Pp_f$.
Its centered moments of order $q\ge2$ are bounded by
$2^qA_{m,n}^{q-1}Pp_f$.  Expanding the exponential series and applying
Chernoff's inequality therefore gives, for $t>0$,
\[
Pp_f-P_\ell p_f
\le C\sqrt{A_{m,n}Pp_f\,t/\ell}
       +C A_{m,n}t/\ell
\le\tfrac12Pp_f+C A_{m,n}t/\ell
\]
outside an event of probability at most $e^{-t}$.
Apply this inequality to an $\epsilon$-net with logarithmic cardinality
$H$ and use a union bound with $t+H$.
Passing from the net to the full class changes $Pp-2P_\ell p$ by at
most $3\epsilon$.  Thus
\[
\Pr_k\!\left\{
\sup_f[Pp_f-2P_\ell p_f]_+
>C A_{m,n}(H+t+1)/\ell\right\}\le e^{-t}.
\]
Define the displayed supremum as $Z_{k,\ell}^{\rm pen}$ and integrate the
tail bound to obtain the result.
\end{proof}

In the remaining risk arguments, condition on $\cF_k$.  For every
bounded measurable $f$, define the population Bellman risk by
\[
\mathcal E_k(f):=
\|f-\cT^{\widehat\pi_k}f\|_{L^2(\mu_k)}^2.
\]
For $f\in\cF_n^{\rm sig}\cup\mathcal H^\beta(L)$,
Assumption~\ref{ass:holder} and
Lemma~\ref{lem:population-bellman-extension} place
$\cT^{\widehat\pi_k}f$ in the population auxiliary class.  On this
domain, \eqref{eq:minimax-identification} therefore gives
\[
\mathcal E_k(f)=
\sup_{g\in\mathcal H^\beta(L_{\cT})}\mathcal L_k(f,g).
\]
In particular, $Q^{\widehat\pi_k}$ belongs to this domain and the
population minimum is zero there, so $\mathcal E_k$ is the corresponding
excess risk.  The residual definition, not a profiling identity for
arbitrary bounded functions, is used in the quadratic identities below.

\begin{lemma}%[Bellman curvature and penalty-driven derivative control]
\label{lem:profiled-risk-curvature}
Under Assumptions~\ref{ass:holder} and~\ref{ass:design}, for every
$f\in\cF_n^{\rm sig}$,
\begin{align}
\norm{f-Q^{\widehat\pi_k}}_{L^2(\mu_k)}^2
&\le(1-\sqrt\gamma)^{-2}\mathcal E_k(f),
\label{eq:excess-to-value}\\
\norm{\nabla_a f-\nabla_aQ^{\widehat\pi_k}}_{L^2(\lambda_d)}^2
&\lesssim t^{-2}\mathcal E_k(f)
       +t^{2m-2}\mathcal P_m(f)+t^{2\beta-2},
\label{eq:excess-to-gradient}
\end{align}
where the second inequality holds for every
$0<t\le1/\{2(m-1)\}$.
\end{lemma}
\begin{proof}
Condition on $\cF_k$ and write $\pi=\widehat\pi_k$, $\mu=\mu_k$,
and $u=f-Q^\pi$.  The design identity
$\mu=(1-\gamma)\lambda_d+\gamma\mu P^\pi$ and Jensen's inequality give
\begin{equation*}
\gamma\norm{P^\pi u}_{L^2(\mu)}^2
\le\gamma\int P^\pi(u^2)\,\dd\mu\le\norm u_{L^2(\mu)}^2.
\end{equation*}
For $\gamma=0$ the assertion below is immediate; otherwise the reverse
triangle inequality gives
\begin{equation*}
\norm{(I-\gamma P^\pi)u}_{L^2(\mu)}
\ge(1-\sqrt\gamma)\norm u_{L^2(\mu)}.
\end{equation*}
The square of the left-hand side is the Bellman risk, proving
\eqref{eq:excess-to-value}.
More generally, for two functions $f,g$ in the population critic class
and $0\le t\le1$, the quadratic identity gives
\begin{align*}
\mathcal E_k(tf+(1-t)g)
&=t\mathcal E_k(f)+(1-t)\mathcal E_k(g)
       -t(1-t)\norm{(I-\gamma P^\pi)(f-g)}_{L^2(\mu)}^2\\
&\le t\mathcal E_k(f)+(1-t)\mathcal E_k(g)
       -(1-\sqrt\gamma)^2t(1-t)\norm{f-g}_{L^2(\mu)}^2.
\end{align*}
Thus the population risk has quadratic curvature in function space,
not necessarily in network parameters.

For the derivative bound, choose
coefficients $c_0,\ldots,c_{m-1}$
satisfying
$\sum_{l=0}^{m-1}c_ll^q=\mathbf1\{q=1\}$ for $0\le q\le m-1$;
the Vandermonde matrix is invertible.  Fix an action coordinate $i$.
Use the forward stencil on $\{a_i\le1/2\}$ and the backward stencil on
$\{a_i>1/2\}$, with step $t\le1/\{2(m-1)\}$, and write
\[
D_{i,t}v(z)=\frac{\pm1}{t}
       \sum_{l=0}^{m-1}c_l v(z\pm lt e_{d_s+i}).
\]
The sign is chosen on each half-cube to keep all stencil points in the
domain.  Translation of a half-cube
into the cube and the triangle inequality give
$\|D_{i,t}v\|_{L^2(\lambda_d)}
\lesssim t^{-1}\|v\|_{L^2(\lambda_d)}$.

For the smooth network $f$, the integral Taylor remainder in the
coordinate $a_i$ is
\[
f(z\pm lt e_{d_s+i})
=\sum_{q=0}^{m-1}\frac{(\pm lt)^q}{q!}
       \partial_{a_i}^q f(z)
+\frac{(\pm lt)^m}{(m-1)!}
 \int_0^1(1-v)^{m-1}
       \partial_{a_i}^mf(z\pm vlt e_{d_s+i})\,\dd v.
\]
The stencil cancels every displayed polynomial term except the term corresponding
to the first derivative.  Minkowski's inequality, followed by translation on each
half-cube, therefore gives
\[
\|\partial_{a_i}f-D_{i,t}f\|_{L^2(\lambda_d)}
\lesssim t^{m-1}\|\partial_{a_i}^mf\|_{L^2(\lambda_d)}.
\]
For the target, the H\"older Taylor remainder of order $m-1$ instead
gives
$\|\partial_{a_i}Q^\pi-D_{i,t}Q^\pi\|_\infty
\lesssim t^{\beta-1}$.
Combining the preceding bounds, squaring, and summing over action
coordinates yields
\[
\|\nabla_a(f-Q^\pi)\|_{L^2(\lambda_d)}^2
\lesssim t^{-2}\|f-Q^\pi\|_{L^2(\lambda_d)}^2
       +t^{2m-2}\mathcal P_m(f)+t^{2\beta-2}.
\]
Use $\mu\ge(1-\gamma)\lambda_d$ and
\eqref{eq:excess-to-value} to finish the proof.
\end{proof}

The next lemma bounds Bellman risk and fitted roughness jointly.

\begin{lemma}%[Expected excess-risk decomposition for trainable DNNs]
\label{lem:critic-error-decomposition}
Under the assumptions and architecture calibration of
Theorem~\ref{thm:critic-l2}, let $q_0\in\cF_n^{\rm sig}$ be any comparator
fixed conditional on $\cF_k$.  For any $\lambda_n>0$ and positive integer
$\ell_n$, the estimator in \eqref{eq:sigmoid-minimax} satisfies
\begin{align}
\bE_k\mathcal E_k(\widehat Q_{k+1})
+\lambda_n\bE_k\mathcal P_m(\widehat Q_{k+1})
\lesssim
& \norm{q_0-Q^{\widehat\pi_k}}_\infty^2
+\sup_{f\in\cF_n^{\rm sig}}\inf_{g\in\cG_n^{\rm sig}}
\norm{g-\cT^{\widehat\pi_k}f}_{L^2(\mu_k)}^2\notag\\
&+\lambda_n\mathcal P_m(q_0)
+\frac{\mathcal M_n(\log n)^2}{n}\notag\\
&+\frac{\lambda_nA_{m,n}\mathcal M_n}{\ell_n}
       \{(\log n)^2+\log(\ell_n+1)\}.
\label{eq:excess-risk-decomposition}
\end{align}
\end{lemma}
\begin{proof}
Condition on $\cF_k$ and write $\pi=\widehat\pi_k$ and $\mu=\mu_k$.
Abbreviate the auxiliary approximation term by
$a_G^2=\sup_{f\in\cF_n^{\rm sig}}\inf_{g\in\cG_n^{\rm sig}}
\|g-\cT^{\widehat\pi_k}f\|_{L^2(\mu_k)}^2$.
For a single tuple, define
\begin{equation*}
\ell_{f,g}=(f(Z)-g(Z))
              \{f(Z)+g(Z)-2R-2\gamma f(Z')\}.
\end{equation*}
Let $T=\cT^\pi$ and $Q=Q^\pi$.  The population cancellation identity is
\begin{equation*}
P\ell_{f,g}=\mathcal E_k(f)-\norm{g-Tf}_{L^2(\mu)}^2.
\end{equation*}
Here $P$ denotes expectation and $P_n$ the corresponding empirical average.  The
identity follows from $\bE(R+\gamma f(Z')\mid Z)=Tf(Z)$.

For the localized fluctuation bound, the bound $|R|\le R_{\max}$ and the
uniform bounds on function values give
\[
|\ell_{f,g}|\lesssim |f(Z)-g(Z)|.
\]
Since $f-g$ also has a fixed uniform bound, integration gives constants
$v,b$ depending only on the structural bounds such that, for every
integer $q\ge2$,
\begin{equation*}
\bE|\ell_{f,g}-P\ell_{f,g}|^q
\le\frac{q!}{2}v b^{q-2}\norm{f-g}_{L^2(\mu)}^2.
\end{equation*}
Jensen's inequality with an independent copy gives the centered bound
after $2^q$ is absorbed into $v,b$.  The exponential series then gives
\begin{equation*}
\Pr\left\{|(P_n-P)\ell_{f,g}|>
 C\norm{f-g}_{L^2(\mu)}\sqrt{t/n}+Ct/n\right\}\le2e^{-t}.
\end{equation*}
Moreover,
$\norm{f-g}_{L^2(\mu)}^2\le
2\mathcal E_k(f)+2\norm{g-Tf}_{L^2(\mu)}^2$.
Young's inequality therefore implies, for each fixed $0<\theta<1$,
\begin{equation*}
|(P_n-P)\ell_{f,g}|
\le\theta\{\mathcal E_k(f)+\norm{g-Tf}_{L^2(\mu)}^2\}+C_\theta t/n
\end{equation*}
outside an event of probability at most $2e^{-t}$.

Take uniform $\epsilon$-nets of both network classes with
$\epsilon=n^{-3}$ and let $H$ be the logarithm of their combined cardinality.
Lemma~\ref{lem:full-dnn-cover} gives
$H\le C\mathcal M_n(\log n)^2$.
A union bound with $t+H+\log2$ gives the preceding inequality for every net
pair.  If $(f_0,g_0)$ is a net approximation to $(f,g)$, direct expansion of
$\ell$ gives
\begin{equation*}
|\ell_{f,g}-\ell_{f_0,g_0}|
\le C\epsilon.
\end{equation*}
Also, $\norm{Tf-Tf_0}_\infty\le\gamma\epsilon$, so the localization term
changes by at most $C\epsilon$.  Consequently, for the measurable random supremum
\begin{equation*}
Z_n=\sup_{f\in\cF_n^{\rm sig},g\in\cG_n^{\rm sig}}
\left[|(P_n-P)\ell_{f,g}|
 -\theta\{\mathcal E_k(f)+\norm{g-Tf}_{L^2(\mu)}^2\}\right]_+,
\end{equation*}
integrating the tail bound yields
\begin{equation}
\bE Z_n\le C_\theta(H+1)/n+C\epsilon
          \le C_\theta\mathcal M_n(\log n)^2/n.
\label{eq:localized-dnn-fluctuation}
\end{equation}

The cancellation identity and the definition of $Z_n$ give, for every $f$,
\begin{equation*}
\sup_gP_n\ell_{f,g}
\ge(1-\theta)\mathcal E_k(f)-(1+\theta)a_G^2-Z_n,\qquad
\sup_gP_n\ell_{q_0,g}
\le(1+\theta)\mathcal E_k(q_0)+Z_n.
\end{equation*}
The second inequality holds because the coefficient of
$\norm{g-Tq_0}_{L^2(\mu)}^2$ is $-(1-\theta)$.
Compare the penalized minimizer with $q_0$.  Write
$\widehat{\mathcal P}_{m,k}(f)=
\ell_n^{-1}\sum_{r=1}^{\ell_n}\sum_j
|\partial_{a_j}^mf(U_{k,r})|^2$ for the numerical integral in the
criterion.  With $\theta=1/4$,
\begin{equation*}
\tfrac34\mathcal E_k(\widehat Q_{k+1})
+\lambda_n\widehat{\mathcal P}_{m,k}(\widehat Q_{k+1})
\le\tfrac54\mathcal E_k(q_0)+\tfrac54a_G^2+2Z_n
          +\lambda_n\widehat{\mathcal P}_{m,k}(q_0).
\end{equation*}
Lemma~\ref{lem:penalty-collocation} bounds the penalty on the left-hand side from below by
$\{\mathcal P_m(\widehat Q_{k+1})-Z_{k,\ell_n}^{\rm pen}\}/2$,
uniformly over the fitted parameters.  Since $q_0$ is fixed conditional
on the history and the integration points are independent and uniform,
$
\bE_k\widehat{\mathcal P}_{m,k}(q_0)=\mathcal P_m(q_0).
$
Taking expectations yields
\begin{align*}
\tfrac34\bE_k\mathcal E_k(\widehat Q_{k+1})
+\tfrac{\lambda_n}{2}\bE_k\mathcal P_m(\widehat Q_{k+1})
\le{}&
\tfrac54\mathcal E_k(q_0)+\tfrac54a_G^2+2\bE_kZ_n+\lambda_n\mathcal P_m(q_0)
+\tfrac{\lambda_n}{2}\bE_kZ_{k,\ell_n}^{\rm pen}.
\end{align*}
Use $\mathcal E_k(q_0)\le(1+\gamma)^2\|q_0-Q\|_\infty^2$,
\eqref{eq:localized-dnn-fluctuation}, and
Lemma~\ref{lem:penalty-collocation}.  Under the network calibration,
$(\mathcal D_n+1)\log\{(\mathcal W_n+d+1)(\mathcal B_n+1)\}
\lesssim(\log n)^2$, proving \eqref{eq:excess-risk-decomposition}.

\end{proof}

\begin{proof}[Proof of Theorem~\ref{thm:critic-l2}]
Condition on $\cF_k$ and set $\pi=\widehat\pi_k$.
Write
$\delta_n=n^{-2\beta/(2\beta+d)}(\log n)^3$.
First suppose $\delta_n<1$.
Choose $J_n=\lceil c_0'\delta_n^{-1/(2\beta)}\rceil$, where $c_0'\ge2$
is fixed, and a depth
$\mathcal D_n\asymp\log n$ whose fixed multiplier ensures
$\mathcal D_n\ge D_0$.
Lemma~\ref{lem:sigmoid-dnn} requires the following architecture bounds:
\begin{equation*}
\mathcal W_n\ge C J_n^d,\qquad
\mathcal M_n\ge C J_n^d\mathcal D_n,\qquad
\mathcal B_n\ge C\{J_n(\mathcal D_n+1)\}^{c_0}.
\end{equation*}
The width and size orders in the theorem meet the first two inequalities
with sufficiently large fixed multipliers.
Since $J_n\lesssim n^{1/(2\beta+d)}$ and
$\mathcal D_n+1\lesssim\log n\lesssim n$, the third follows by taking,
for example, $c_B>c_0\{1+(2\beta+d)^{-1}\}$ and a sufficiently large
fixed multiplier in $\mathcal B_n$.

Apply the lemma first to $Q^\pi\in\mathcal H^\beta(L)$ and then to
$\cT^\pi f\in\mathcal H^\beta(L_{\cT})$, uniformly over
$f\in\cF_n^{\rm sig}$.  The latter satisfies
\[
\|\cT^\pi f\|_\infty
\le R_{\max}+2\gamma B_Q=(1+\gamma)B_Q\le2B_Q.
\]
Output rescaling in the lemma yields a feasible
$q_0\in\cF_n^{\rm sig}$ satisfying
\begin{align*}
\|q_0-Q^\pi\|_\infty^2&\lesssim J_n^{-2\beta},\\
\|\nabla_aq_0-\nabla_aQ^\pi\|_{L^2(\lambda_d)}^2
&\lesssim J_n^{-2(\beta-1)},\\
\mathcal P_m(q_0)&\lesssim J_n^{2(m-\beta)},\\
\sup_{f\in\cF_n^{\rm sig}}\inf_{g\in\cG_n^{\rm sig}}
\|g-\cT^\pi f\|_{L^2(\mu_k)}^2
&\lesssim J_n^{-2\beta}.
\end{align*}
The last inequality follows from uniform approximation of
$\cT^\pi f$ and the fact that $\mu_k$ is a probability measure.

The architecture gives $\log A_{m,n}\lesssim(\log n)^2$.
The specified integration budget consequently satisfies
$\log(\ell_n+1)\lesssim(\log n)^2$, since its remaining factors
have logarithms of order at most $\log n$.
Thus
\[
\frac{\lambda_nA_{m,n}\mathcal M_n}{\ell_n}
 \{(\log n)^2+\log(\ell_n+1)\}
\lesssim\delta_n.
\]
Also, $J_n^{-2\beta}\lesssim\delta_n$,
$\mathcal M_n(\log n)^2/n\lesssim\delta_n$, and
\[
\lambda_n\mathcal P_m(q_0)
\lesssim
\delta_n^{m/\beta}\delta_n^{1-m/\beta}
=\delta_n.
\]
Lemma~\ref{lem:critic-error-decomposition} now gives
\[
\bE_k\mathcal E_k(\widehat Q_{k+1})
+\lambda_n\bE_k\mathcal P_m(\widehat Q_{k+1})
\lesssim\delta_n.
\]
Bellman coercivity in Lemma~\ref{lem:profiled-risk-curvature} gives the
value bound.  For the gradient bound, apply
\eqref{eq:excess-to-gradient} with
$t=c\,\delta_n^{1/(2\beta)}$, where
$0<c\le1/\{2(m-1)\}$.  Since
$\lambda_n\asymp\delta_n^{m/\beta}$, the three terms in that
inequality are bounded in expectation by
\[
t^{-2}\delta_n
+t^{2m-2}\delta_n/\lambda_n+t^{2\beta-2}
\lesssim\delta_n^{(\beta-1)/\beta}.
\]
This proves \eqref{eq:gradient-l2-rate} with the stated logarithmic power.

It remains to consider $\delta_n\ge1$.  Use the zero network as
a comparator; its penalty is zero and its Bellman risk is bounded.
Taking the zero auxiliary network also bounds the auxiliary
approximation term by a structural constant.
The same integration-budget calculation and oracle inequality give
$\bE_k\mathcal E_k(\widehat Q_{k+1})
+\lambda_n\bE_k\mathcal P_m(\widehat Q_{k+1})
\lesssim\delta_n$.
Because $m\ge\beta$,
$\bE_k\mathcal P_m(\widehat Q_{k+1})
\lesssim\delta_n^{1-m/\beta}\le C$.
The uniform bound on function values controls $\mathcal E_k(\widehat Q_{k+1})$
deterministically by a structural constant.
Use a fixed admissible $t$ in \eqref{eq:excess-to-gradient} to bound
the expected derivative risk by a constant, which is at most
$C\delta_n^{(\beta-1)/\beta}$.
\end{proof}

\section{Proofs for Langevin Sampling Error with an Estimated Critic}
\label{app:sampling}

The drift bounds and sampling profiles yield frozen-step entropy decay,
exact-diffusion mixing, and a time-uniform numerical comparison.  Combining
these results with the value and gradient risks gives the actor bound in
Section~\ref{sec:sampling}.

\begin{lemma}
\label{lem:architecture-drift}
For the calibrated architecture bounds of Theorem~\ref{thm:critic-l2},
every critic in Definition~\ref{def:localized-sigmoid-classes} and every
population action-value function satisfying Assumption~\ref{ass:holder} obeys
\begin{equation*}
\sup_{s,a}\norm{\nabla_aF(s,a)}_2\le B_{1,n},\qquad
\sup_s\operatorname{Lip}\{\nabla_aF(s,\cdot)\}\le L_{b,n},
\end{equation*}
with the deterministic bounds in Section~\ref{sec:sampling}.
If $\mathcal W_n,\mathcal B_n$ are polynomial in $n$ and
$\mathcal D_n\asymp\log n$, then
$\log(1+B_{1,n}+L_{b,n})=O((\log n)^2)$.
\end{lemma}

\begin{proof}
The sigmoid satisfies $|\sigma_{\rm sig}'|\le1$ and
$|\sigma_{\rm sig}''|\le1$.  Let $J_j$ and $H_j$ be the largest absolute
first and second coordinate derivatives, respectively, among the units
in hidden layer $j$.  The input layer has $J_0=1,H_0=0$.
The architecture calibration includes
$\mathcal W_n\ge d+2$, $\mathcal B_n\ge1$, and
$A_n^{\mathcal D_n+1}\ge L$.  These conditions affect only the fixed architecture
factors.  The sum of the absolute coefficients in each linear row is at most $A_n$.
The chain rule therefore gives
\begin{equation*}
J_j\le A_nJ_{j-1},\qquad
H_j\le A_n^2J_{j-1}^2+A_nH_{j-1}.
\end{equation*}
Induction yields $J_j\le A_n^j$ and $H_j\le jA_n^{2j}$.
The affine output has first-order coordinate derivatives bounded
by $A_n^{\mathcal D_n+1}$ and second-order coordinate derivatives bounded by
$(\mathcal D_n+1)A_n^{2(\mathcal D_n+1)}$; these bounds also cover networks
with fewer hidden layers.  Passing to the gradient and Hessian norms
introduces factors $\sqrt{d_a}$ and
$d_a$, respectively.  The population derivatives are bounded by
$L$ under Assumption~\ref{ass:holder}; the mean-value theorem gives the
drift Lipschitz bound.  Finally, $\log A_n=O(\log n)$ and
$\mathcal D_n=O(\log n)$ prove the last assertion.
\end{proof}

\begin{lemma}
\label{lem:profile-verification}
The bounded-potential and strongly concave profiles in
\eqref{eq:multimodal-profile} and \eqref{eq:strong-concave-profile} are valid.
\end{lemma}

\begin{proof}
For $\norm F_\infty\le B_F$, Lemma~\ref{lem:derived-lsi} gives
$\alpha_\tau=\widetilde\alpha e^{-2B_F/\tau}$.  On the action cube,
$W_1(\mu,\nu)\le\sqrt{d_a}\TV(\mu,\nu)$; Pinsker's inequality in Lemma~\ref{lem:aux-information} therefore gives
$W_1^2(\mu,\nu)\le(d_a/2)\KL(\mu\|\nu)$, corresponding to the profile coordinate
$\chi_\tau=d_a/4$.

The unit-volume action cube and the uniform bound on function values give
$
\frac{e^{F(a)/\tau}}{\int_{\cA}e^{F(u)/\tau}\dd u}
\le e^{2B_F/\tau}.
$
This proves \eqref{eq:multimodal-profile} without a derivative constraint.

For a concave potential, continuity and compactness give a maximizer
$a_F$, possibly on the boundary.  Put $t=\tau/(1+2B_F)\le1$ and
consider the homothetic copy $a_F+t(\cA-a_F)$ of the cube.  It lies
in $\cA$, has volume $t^{d_a}$, and concavity implies, for $a\in\cA$,
\begin{equation*}
F\{a_F+t(a-a_F)\}
\ge (1-t)F(a_F)+tF(a)
\ge F(a_F)-2B_Ft\ge F(a_F)-\tau.
\end{equation*}
Consequently, the normalizer is at least
$t^{d_a}e^{F(a_F)/\tau-1}$ and
\begin{equation*}
\left\|\frac{\dd\pi_F}{\dd\lambda_{\cA}}\right\|_\infty
\le e(1+2B_F)^{d_a}\tau^{-d_a}.
\end{equation*}
This density bound uses only concavity and the uniform bound on function values.  If $F$ is
$\rho$-strongly concave, Lemma~\ref{lem:aux-strong-lsi-transport}
and the Herbst argument following it give
$\alpha_\tau=\rho/\tau$ and
$W_1^2\le2(\tau/\rho)\KL$, proving
\eqref{eq:strong-concave-profile}.
\end{proof}

\begin{lemma}%[Gibbs perturbation by value and gradient errors]
\label{lem:gibbs-risk-transfer}
Let $F,Q$ be continuously differentiable action potentials with
$\|F\|_\infty\le B_F=2B_Q$, $\|Q\|_\infty\le B_Q$, and
$\max_j\|\partial_{a_j}Q\|_\infty\le L$.  Suppose their Gibbs laws
$\pi_F,\pi_Q$ have the common profile $(\alpha_\tau,\chi_\tau,D_\tau)$.
For $0<\tau\le1$,
\begin{align*}
W_1^2(\pi_F,\pi_Q)
\lesssim\min\Bigl\{&
\frac{\chi_\tau\min\{D_\tau,\tau^{-d_a}\}}{\alpha_\tau\tau^2}
 \|\nabla F-\nabla Q\|_{L^2(\lambda_{\cA})}^2,\frac{\chi_\tau}{\tau}e^{3B_F/(2\tau)}
 \|F-Q\|_{L^2(\lambda_{\cA})}^2\Bigr\}.
\end{align*}
Moreover,
\begin{equation*}
\int_{\cA}\{\max Q-Q(a)\}\,\pi_F(\dd a)
\lesssim b_\tau+e^{3B_F/(2\tau)}\|F-Q\|_{L^2(\lambda_{\cA})}^2.
\end{equation*}
\end{lemma}

\begin{proof}
Choose $a^\star\in\argmax_{\cA}Q$ and put
$h=\tau/(\tau+d_aL)\le1$.  The set
$a^\star+h(\cA-a^\star)$ lies in the cube, has volume $h^{d_a}$, and
on this set $Q\ge Q(a^\star)-d_aLh\ge Q(a^\star)-\tau$.
Thus, without a concavity or interior-maximizer assumption,
\begin{equation*}
\left\|\frac{\dd\pi_Q}{\dd\lambda_{\cA}}\right\|_\infty
\le e(1+d_aL/\tau)^{d_a}\lesssim\tau^{-d_a}.
\end{equation*}
Use the transport inequality and LSI for $\pi_F$, in that order:
\begin{align*}
W_1^2(\pi_Q,\pi_F)
&\le2\chi_\tau\KL(\pi_Q\|\pi_F)\\
&\le\frac{\chi_\tau}{\alpha_\tau\tau^2}
 \int_{\cA}\|\nabla Q-\nabla F\|_2^2\,\dd\pi_Q\\
&\lesssim\frac{\chi_\tau\min\{D_\tau,\tau^{-d_a}\}}
                  {\alpha_\tau\tau^2}
 \|\nabla Q-\nabla F\|_{L^2(\lambda_{\cA})}^2.
\end{align*}
This proves the gradient route.

For the value route, set $H=F-Q$, $F_t=Q+tH$, and
$\psi(t)=\log\int_{\cA}e^{F_t/\tau}\dd a$ for $0\le t\le1$.
Differentiating under the integral gives
$\psi'(t)=\tau^{-1}\int H\,\dd\pi_{F_t}$ and
$\psi''(t)=\tau^{-2}\operatorname{Var}_{\pi_{F_t}}(H)$.  Hence
\begin{equation}
\KL(\pi_F\|\pi_Q)
=\psi'(1)-\psi(1)+\psi(0)
=\frac1{\tau^2}\int_0^1t\operatorname{Var}_{\pi_{F_t}}(H)\,\dd t.
\label{eq:gibbs-path-kl}
\end{equation}
Put $\epsilon=\|H\|_{L^2(\lambda_{\cA})}$.
If $\epsilon\le\tau$, Jensen's inequality and the cube's unit volume give
\[
\int_{\cA}e^{F_t/\tau}\dd a
\ge\exp\!\left\{\tau^{-1}\int_{\cA}(Q+tH)\dd a\right\}
\ge e^{(-B_Q-t\epsilon)/\tau}.
\]
Since $\sup F_t\le B_Q(1+t)$, the interpolated Gibbs density is at
most $e\exp\{(2+t)B_Q/\tau\}$.  Therefore
\begin{align*}
\KL(\pi_F\|\pi_Q)
&\le\frac{e\epsilon^2}{\tau^2}
       \int_0^1t e^{(2+t)B_Q/\tau}\dd t
\le\frac{e\,e^{3B_Q/\tau}}{B_Q\tau}\epsilon^2.
\end{align*}
If $\epsilon>\tau$, the bound $\|H\|_\infty\le3B_Q$ instead gives
\[
\KL(\pi_F\|\pi_Q)\le6B_Q/\tau
\lesssim\tau^{-1}e^{3B_Q/\tau}\epsilon^2,
\]
because $\sup_{0<v\le1}v^{-2}e^{-3B_Q/v}<\infty$.
Both cases give the same bound, with $3B_Q=3B_F/2$.
The case of a zero bound on function values is immediate.  The transport inequality for
$\pi_Q$ proves the value route.
The interpolation potentials need not belong to the neural class.

Finally, write $U=Q(a^\star)-Q\ge0$ and
$Z_U(v)=\int_{\cA}e^{-U/v}\dd a$.  The same homothetic cube gives
$Z_U(\tau)\ge e^{-1}h^{d_a}$, while $Z_U(2\tau)\le1$.
Jensen's inequality yields
\begin{align}
\frac1{2\tau}\int_{\cA}U\,\dd\pi_F-\KL(\pi_F\|\pi_Q)
&\le\log\int_{\cA}e^{U/(2\tau)}\,\dd\pi_Q
=\log\frac{Z_U(2\tau)}{Z_U(\tau)}.
\label{eq:gibbs-greedy-entropy}
\end{align}
The last logarithm is at most
$1+d_a\log(1+d_aL/\tau)=b_\tau/\tau$.
Combining this with the value-based bound on relative entropy proves the
greediness assertion.
\end{proof}

\begin{lemma}%[Uniform initialization bound]
\label{lem:init-entropy}
If $\norm{F}_\infty\le B_F$, then normalized Lebesgue initialization on the
action cube satisfies
$
\KL(\lambda_{\cA}\|\pi_F)\le 2B_F/\tau.
$
\end{lemma}

\begin{proof}
Since the action cube has unit volume,
\[
\KL(\lambda_{\cA}\|\pi_F)
=
-\tau^{-1}\int_{\cA}F(a)\dd a
+
\log\int_{\cA}e^{F(a)/\tau}\dd a
\le
2B_F/\tau.
\]
\end{proof}

\begin{lemma}%[Derived log-Sobolev constant on the cube]
\label{lem:derived-lsi}
If $\norm{F}_\infty\le B_F$, then $\pi_F$ satisfies an LSI with constant at
least
\begin{equation}
\label{eq:derived-lsi-constant}
\alpha_{\rm LSI}:=\widetilde\alpha\exp(-2B_F/\tau)>0,
\end{equation}
where $\widetilde\alpha$ is the LSI constant of normalized Lebesgue measure
on $[0,1]^{d_a}$.
\end{lemma}

\begin{proof}
For a smooth nonnegative $u$ on the unit interval with
$\int_0^1u^2=1$, put $E=\int_0^1|u'|^2$ and
$\operatorname{Ent}(u^2)=\int_0^1u^2\log u^2$.
The fundamental theorem of calculus and the Cauchy--Schwarz inequality give
$\|u\|_\infty\le1+\sqrt E$ and $\operatorname{Var}(u)\le E$.
If $E\le1$, the inequality $\log x\le x-1$ and the variance identity for two independent copies give
\[
\operatorname{Ent}(u^2)\le\operatorname{Var}(u^2)
\le4\|u\|_\infty^2\operatorname{Var}(u)\le16E.
\]
If $E>1$, then
$\operatorname{Ent}(u^2)\le2\log(1+\sqrt E)\le2E$.
Approximation extends the inequality to nonnegative absolutely continuous $u$ with a square-integrable derivative.
For a density $p=u^2$, $I(p\|1)=4E$, so the interval LSI holds with
$\alpha=1/8$.  Tensorization in Lemma~\ref{lem:aux-lsi-information}
therefore gives a fixed constant $\widetilde\alpha>0$ for $\lambda_{\cA}$.  Writing $\pi_F(a)$ for the Gibbs density with respect to $\lambda_{\cA}$,
\[
\pi_F(a)
=
\frac{e^{F(a)/\tau}}{\int e^{F(u)/\tau}\dd u},
\qquad
\frac{\sup_a\pi_F(a)}
{\inf_a\pi_F(a)}
\le e^{\osc(F)/\tau}
\le e^{2B_F/\tau}.
\]
Lemma~\ref{lem:aux-lsi-information} multiplies the Lebesgue LSI constant
by at least $e^{-\osc(F)/\tau}$, proving
\eqref{eq:derived-lsi-constant} uniformly in $s$, $k$, and the fitted
potential $F=\widehat Q_{k+1}$.
\end{proof}

\subsection{Frozen-step entropy on the cube}
Positive-time regularity and facewise flux identities justify entropy
dissipation for the frozen drift.

\begin{lemma}%[Regularity of a frozen reflected step]
\label{lem:frozen-kernel-rigorous}
Let $b:\cA\to\mathbb R^{d_a}$ satisfy
$\|b\|_\infty\le B_{1,n}$ and $\operatorname{Lip}(b)\le L_{b,n}$.
For an initial law $X_0\sim\nu$, conditional on $X_0=x$, let
$X_t=\Gamma_{\cA}\{x+tb(x)+\sqrt{2\tau}B_t\}$.
For every initial probability law, the law of $X_t$ has a density
$\rho_t$ that is strictly positive and smooth in the spatial variable on
the closed cube for $t>0$.  For every $\epsilon>0$, the density and its
spatial and temporal derivatives are bounded on compact time intervals in
$[\epsilon,\infty)$.  Put
\begin{equation*}
J_t(y):=\int b(x)p_t^{b(x)}(x,y)\nu(\dd x),
\qquad
\bar b_t(y):=J_t(y)/\rho_t(y).
\end{equation*}
Then, we have 
\begin{equation*}
\partial_t\rho_t=\tau\Delta\rho_t-\nabla\!\cdot J_t,
\qquad
(\tau\nabla\rho_t-J_t)\cdot n_{\rm out}=0
\end{equation*}
on each face of the cube.  Moreover,
\begin{equation*}
\int\norm{\bar b_t(y)-b(y)}_2^2\rho_t(y)\dd y
\lesssim L_{b,n}^2(d_a\tau t+B_{1,n}^2t^2).
\end{equation*}
\end{lemma}

\begin{proof}
Let $\lambda_1$ denote the uniform probability measure on $[0,1]$.
Conditional on the starting point, the constant-drift kernel factors
coordinatewise.  For a scalar drift $c$, write $a=c/(2\tau)$ and let
\begin{align*}
\phi_0^a(u)&:=\frac{e^{au}}{\{\int_0^1e^{2av}\dd v\}^{1/2}},
&\lambda_0^a&:=-a^2,\\
\phi_j^a(u)&:=
 \left\{\frac{2}{1+(a/(j\pi))^2}\right\}^{1/2}
 \left\{\cos(j\pi u)+\frac{a}{j\pi}\sin(j\pi u)\right\},
&\lambda_j^a&:=j^2\pi^2,\qquad j\ge1.
\end{align*}
These functions solve $-\phi''=\lambda\phi$ with
$\phi'(0)=a\phi(0)$ and $\phi'(1)=a\phi(1)$.  Integration by parts gives
orthogonality; the normalization factors give unit $L^2(\lambda_1)$ norms.
If $h\in L^2(\lambda_1)$ is orthogonal to this family, then
\begin{equation*}
\int_0^1h(u)\{v'(u)+av(u)\}\dd u=0
\end{equation*}
for every finite sine polynomial $v$, and hence for every
$v\in H_0^1([0,1])$ by the density of sine polynomials.  Here this Sobolev
space consists of $L^2(\lambda_1)$ functions with a weak derivative in
$L^2(\lambda_1)$ and zero endpoint traces.
Thus $h'=ah$ in distributions, so $h=Ce^{au}$ almost everywhere.
Orthogonality to $\phi_0^a$ forces $C=0$, proving completeness.
The forward kernel is
\begin{equation*}
p_t^c(x,y)
=e^{a(y-x)-\tau a^2t}
 \sum_{j=0}^{\infty}e^{-\tau\lambda_j^a t}
                  \phi_j^a(x)\phi_j^a(y).
\end{equation*}
Direct substitution verifies
$\partial_tp=\tau\partial_{yy}p-c\partial_yp$ and
$\partial_yp=(c/\tau)p$ at both endpoints.  Completeness gives the initial
Dirac mass, and parabolic uniqueness identifies the reflected kernel.
The maximum principle gives positivity in the interior; a zero endpoint value
and the Robin condition would contradict the boundary point lemma.

For $t\ge\epsilon>0$ and $|c|\le B_{1,n}$, each fixed-order derivative of the
series is dominated by a constant times
$\sum_{j\ge1}(1+j)^r e^{-\tau\epsilon j^2\pi^2}$ for some finite $r$.
This summable bound is uniform in $x,y,c$ on their compact ranges.
Taking products gives the cube kernel, and the uniform bounds justify
differentiation under the integral against $\nu(\dd x)$.
Integrating the conditional equations yields
$\rho_t(y)=\int p_t^{b(x)}(x,y)\nu(\dd x)$ and the asserted equation and
facewise flux identities for $J_t$ in the classical sense at positive times.  Positivity and
compactness give a positive lower bound for $\rho_t$ on each compact
positive-time interval.

Finally, conditional Jensen's inequality and Lipschitz continuity give
\begin{equation*}
\int\norm{\bar b_t-b}_2^2\rho_t
\le L_{b,n}^2\bE\norm{X_t-X_0}_2^2.
\end{equation*}
Apply the Lipschitz property of $\Gamma_{\cA}$ to the driven path and the
constant path $X_0$.  The Brownian maximal second-moment bound yields
$\bE\norm{X_t-X_0}_2^2\lesssim B_{1,n}^2t^2+d_a\tau t$, proving the claim.
\end{proof}

\begin{lemma}%[Derived reflected-Euler entropy error]
\label{lem:ula-entropy}
For a fitted potential $F$ in the DNN critic class of
Theorem~\ref{thm:critic-l2}, suppose its Gibbs law satisfies an LSI with
constant $\alpha>0$.  Write $P_{\eta,F}^{\rm refl}$ for the Brownian-path frozen-drift transition in \eqref{eq:ula-actor}.  For
$0<\eta\le\min\{1,(\alpha\tau)^{-1}\}$, $M\ge1$, and every initial law
$\nu_0$ with finite
relative entropy,
\begin{equation*}
\KL\left(\nu_0(P_{\eta,F}^{\rm refl})^M\middle\|\pi_F\right)
\lesssim
e^{-\alpha\tau M\eta}\KL(\nu_0\|\pi_F)
+\frac{d_aL_{b,n}^2\eta}{\alpha\tau}
+\frac{L_{b,n}^2B_{1,n}^2\eta^2}{\alpha\tau^2}.
\end{equation*}
\end{lemma}

\begin{proof}
Fix the state and the fitted potential, and write $b=\nabla_aF$ and
$\pi=\pi_F$.  The preceding positive-time regularity and flux identities
justify differentiation of $H_t=\KL(\rho_t\|\pi)$.
With $I_t=\int\norm{\nabla\log(\rho_t/\pi)}_2^2\rho_t$, integration by parts gives
\begin{align*}
H_t'
&=-\tau I_t+
  \int\langle\bar b_t-b,\nabla\log(\rho_t/\pi)\rangle\rho_t\\
&\le-\frac\tau2I_t+
  \frac1{2\tau}\int\norm{\bar b_t-b}_2^2\rho_t\\
&\le-\alpha\tau H_t+
 C L_{b,n}^2\{d_at+B_{1,n}^2t^2/\tau\}.
\end{align*}
The fundamental theorem of calculus and Fubini's theorem justify
coordinatewise integration by parts; the facewise flux identities cancel
the endpoint terms.  No smooth normal field at edges or corners is needed.
The constant $C$ depends only on the cube Skorokhod constant;
all dependence on the bounds on the drift and its Lipschitz constant is displayed.

To establish entropy continuity at zero for an arbitrary initial law with
finite entropy, compare the frozen process with the exact
reflected diffusion $\dd Y_t=b(Y_t)\dd t+\sqrt{2\tau}\dd B_t+\dd K_t$
started from $\pi$.  Change its initial distribution to $\nu$ and its
adapted drift to $b(X_0)$.  Boundedness of $b$ ensures Novikov's condition.
Girsanov's theorem on the driving probability space, followed by data
processing for the time-$t$ marginal, gives
\begin{equation*}
\KL(\rho_t\|\pi)
\le\KL(\nu\|\pi)
 +\frac1{4\tau}\int_0^t
        \bE\norm{b(X_0)-b(X_u)}_2^2\dd u.
\end{equation*}
The reference marginal remains $\pi$.  The preceding displacement bound
makes the additional term
$O\{L_{b,n}^2(d_at^2+B_{1,n}^2t^3/\tau)\}$ for each fixed
$n$ and $\tau>0$.  Moreover, $X_t\to X_0$ in probability, so lower semicontinuity of
relative entropy gives the matching lower limit.  Thus
$H_t\to\KL(\nu\|\pi)$ as $t\downarrow0$.

Integrate the differential inequality from $\epsilon$ to $\eta$ and let
$\epsilon\downarrow0$.  This yields
\begin{equation*}
H_\eta\le e^{-\alpha\tau\eta}H_0
             +C L_{b,n}^2\{d_a\eta^2+B_{1,n}^2\eta^3/\tau\}.
\end{equation*}
Iterate this recursion, retaining its exact factor $e^{-\alpha\tau\eta}$.
For $\alpha\tau\eta\le1$,
$1-e^{-\alpha\tau\eta}\ge\alpha\tau\eta/2$, and therefore
\begin{equation*}
\KL\{\nu_0(P_{\eta,F}^{\rm refl})^M\|\pi_F\}
\le e^{-\alpha\tau M\eta}\KL(\nu_0\|\pi_F)
       +C\left\{
        \frac{d_aL_{b,n}^2\eta}{\alpha\tau}
       +\frac{L_{b,n}^2B_{1,n}^2\eta^2}{\alpha\tau^2}\right\}.
\end{equation*}
\end{proof}

\begin{lemma}%[Entropy decay for the exact reflected diffusion]
\label{lem:exact-reflected-entropy}
Let $P_t$ be the exact reflected diffusion semigroup for a fitted
potential $F$, and suppose $\pi_F$ satisfies an LSI with constant
$\alpha_\tau>0$.  For every initial law $\nu$ with finite relative entropy,
\begin{equation*}
\KL(\nu P_t\|\pi_F)
\le e^{-2\alpha_\tau\tau t}\KL(\nu\|\pi_F),\qquad t\ge0.
\end{equation*}
\end{lemma}

\begin{proof}
The Gibbs density has zero flux because
$\tau\nabla\pi_F=(\nabla F)\pi_F$, and is invariant for the exact
normally reflected diffusion.  Fix $n,\tau$ and a finite time $T$.
Retain a Young parameter $0<\varepsilon<1$ in the entropy calculation
of Lemma~\ref{lem:ula-entropy}, using the notation $H_t,I_t,\bar b_t$ from that lemma.
The score--drift cross term is at most
$\varepsilon\tau I_t+
(4\varepsilon\tau)^{-1}\int\|\bar b_t-b\|_2^2\rho_t$.
Thus the entropy dissipation is
$-2(1-\varepsilon)\alpha_\tau\tau H_t$, with the same displacement
remainder multiplied by a constant depending on $\varepsilon$.
For $h=T/j$, iteration therefore gives
\begin{align*}
\KL\{\nu(P_{h,F}^{\rm refl})^j\|\pi_F\}
\le{}&e^{-2(1-\varepsilon)\alpha_\tau\tau T}\KL(\nu\|\pi_F)
+C_\varepsilon\left\{\frac{d_aL_{b,n}^2h}{\alpha_\tau\tau}
 +\frac{L_{b,n}^2B_{1,n}^2h^2}{\alpha_\tau\tau^2}\right\}.
\end{align*}
The final recursion in that lemma preserves the unit coefficient on the
initial-entropy term.

Couple the continuous frozen path $X^h$ and the
exact path $X$ using the same initial point and Brownian motion.
The Skorokhod-map inequality and the Lipschitz drift imply
\begin{align*}
\bE\sup_{u\le t}\norm{X_u-X_u^h}_2
\lesssim L_{b,n}\int_0^t
\left\{\bE\sup_{v\le u}\norm{X_v-X_v^h}_2
+B_{1,n}h+\sqrt{d_a\tau h}\right\}\dd u,
\qquad t\le T.
\end{align*}
Here each within-step displacement of $X^h$ is bounded using the
restart property of the cube Skorokhod map and the Brownian maximal
inequality.  Gronwall's inequality gives convergence to zero as
$h\downarrow0$ for fixed $n,\tau,T$.  This argument is used only for
convergence on a fixed interval, not as a numerical error bound uniform
in $T$.  Lower semicontinuity of relative entropy first gives the bound
with exponent $-2(1-\varepsilon)\alpha_\tau\tau T$.  Letting
$\varepsilon\downarrow0$ proves the assertion.
\end{proof}

\subsection{Time-uniform numerical comparison and the actor bound}
Short-time smoothing and entropy mixing yield summable exact-semigroup
bounds, which propagate local numerical errors uniformly in time.

\begin{lemma}%[Reflected smoothing and summable Lipschitz bounds]
\label{lem:reflected-semigroup-smoothing}
Let $P_t$ be the exact reflected diffusion semigroup with drift
$b=\nabla F$, $\operatorname{Lip}(b)\le L_{b,n}$, and invariant law
$\pi_F$ satisfying the LSI with constant $\alpha_\tau$.
For $t>0$ and $x,y\in\cA$,
\begin{equation*}
\KL\{P_t(x,\cdot)\|P_t(y,\cdot)\}
\le
\frac{L_{b,n}|x-y|^2}
     {2\tau(1-e^{-2L_{b,n}t})}.
\end{equation*}
For every $1$-Lipschitz function $\phi$ and $0<\eta\le1$, the general
choice of $\mathcal C_{\tau,n}$ in Section~\ref{sec:sampling} satisfies
\begin{equation*}
\int_0^\infty\operatorname{Lip}(P_t\phi)\dd t
\lesssim\mathcal C_{\tau,n},\qquad
\sum_{j=0}^\infty\operatorname{Lip}(P_{j\eta}\phi)
\lesssim1+\frac{\mathcal C_{\tau,n}}{\eta}.
\end{equation*}
If $F$ is $\rho$-strongly concave, the latter two bounds also hold with
$\mathcal C_{\tau,n}=1+\rho^{-1}$.
\end{lemma}

\begin{proof}
Write $L=L_{b,n}$ and fix a target time $t>0$ and initial points $x,y$.
We adapt the change-of-measure coupling of \citet{wang2011harnack}
to coordinatewise reflection on the cube.
Let $X$ start from $x$ with its uncontrolled drift and Brownian motion.
Drive a second process $Y$, initially at $y$, by the same Brownian
motion and add, until they meet, the drift
$u_s(X_s-Y_s)/|X_s-Y_s|$, where
\begin{equation*}
u_s:=\frac{|x-y|e^{-Ls}}{\int_0^t e^{-2Lv}\dd v},
\qquad 0\le s\le t.
\end{equation*}
The case $x=y$ is immediate.  For $x\ne y$, construct the coupled
equation on the product cube stopped at distance $\epsilon$ from the
diagonal.  Its coefficients admit bounded Lipschitz extensions, so the
Lipschitz Skorokhod map gives existence and uniqueness by Picard
iteration.  The stopped solutions agree as $\epsilon$ decreases.
Their bounded drift integrals and common continuous Brownian driver,
together with continuity of the Skorokhod map, extend both paths to the
limiting meeting time.  Thereafter set $Y=X$ and remove the control.
Before meeting, $R_s=|X_s-Y_s|$ has no martingale part.  Convexity gives
$\langle X_s-Y_s,\dd K_s^X-\dd K_s^Y\rangle\le0$, hence
\begin{equation*}
\dd R_s\le LR_s\dd s-u_s\dd s,\qquad
e^{-Ls}R_s\le |x-y|-\int_0^se^{-Lv}u_v\dd v.
\end{equation*}
Thus the meeting time is at most $t$.  The control energy satisfies
\begin{equation*}
\int_0^t u_s^2\dd s
=\frac{|x-y|^2}{\int_0^te^{-2Lv}\dd v}
=\frac{2L|x-y|^2}{1-e^{-2Lt}}.
\end{equation*}
Put $v_s=u_s(X_s-Y_s)/(\sqrt{2\tau}|X_s-Y_s|)$ before
meeting and $v_s=0$ afterward.  Its energy has a deterministic bound, so
the stochastic exponential
\begin{equation*}
Z_t=\exp\left\{-\int_0^t\langle v_s,\dd B_s\rangle
                   -\frac12\int_0^t|v_s|^2\dd s\right\}
\end{equation*}
is a martingale by Novikov's condition.  Under the measure with density $Z_t$,
$B_s+\int_0^sv_u\dd u$ is a Brownian motion and $Y$ has its uncontrolled
reflected law from $y$.  Since $X_t=Y_t$ pathwise, the entropy
variational inequality gives, for every bounded $f\ge1$,
\begin{align*}
P_t\log f(y)
&=\bE\{Z_t\log f(X_t)\}\\
&\le\log P_tf(x)+\bE(Z_t\log Z_t)\\
&\le\log P_tf(x)+c_t|x-y|^2,\qquad
c_t:=\frac{L}{2\tau(1-e^{-2Lt})}.
\end{align*}
Taking the variational supremum and exchanging $x,y$ proves the
relative-entropy statement in either direction.  When $L=0$, the same
argument gives $c_t=1/(4\tau t)$ by continuity.

Pinsker's inequality now gives, for every bounded measurable $g$,
\begin{equation*}
\operatorname{Lip}(P_tg)
\le\operatorname{osc}(g)\sqrt{c_t/2},\qquad
c_t\lesssim(L+t^{-1})/\tau.
\end{equation*}
Let $D=\sqrt{d_a}$ and $H=c_1D^2$.  Invariance and convexity of relative
entropy in its second argument yield
\begin{equation*}
\KL\{P_1(x,\cdot)\|\pi_F\}
\le\int_{\cA}\KL\{P_1(x,\cdot)\|P_1(y,\cdot)\}\pi_F(\dd y)
\le H.
\end{equation*}
Lemma~\ref{lem:exact-reflected-entropy} and Pinsker's inequality therefore imply,
uniformly in $x,y$ and $u\ge1$,
\begin{equation*}
\TV\{P_u(x,\cdot),P_u(y,\cdot)\}
\le\min\{1,\sqrt{2H}e^{-\alpha_\tau\tau(u-1)}\}.
\end{equation*}
For a $1$-Lipschitz function $\phi$, $\operatorname{osc}(\phi)\le D$.
At times $0<t\le2$, smoothing gives
\begin{equation*}
\operatorname{Lip}(P_t\phi)
\lesssim D\tau^{-1/2}(\sqrt L+t^{-1/2}).
\end{equation*}
For $t\ge2$, apply the time-one smoothing inequality to
$P_{t-1}\phi$ and then use the preceding total-variation bound:
\begin{equation*}
\operatorname{Lip}(P_t\phi)
\lesssim D\sqrt{\frac{1+L}{\tau}}
 \min\{1,\sqrt{2H}e^{-\alpha_\tau\tau(t-2)}\}.
\end{equation*}
The short-time singularity is integrable.  Moreover, for $A,a>0$,
\begin{equation*}
\int_0^\infty\min\{1,Ae^{-au}\}\dd u
\le\frac{1+\log_+A}{a},
\qquad \log_+A:=\max\{0,\log A\}.
\end{equation*}
Together with $H\lesssim d_a(1+L)/\tau$, this proves the integrated
Lipschitz bound.

For the discrete sum, retain the term $j=0$ as
$\operatorname{Lip}(\phi)\le1$.  The positive short-time terms satisfy
\begin{equation*}
\sum_{1\le j\le2/\eta}(j\eta)^{-1/2}
\le 2\eta^{-1/2}\sqrt{2/\eta}\lesssim\eta^{-1}.
\end{equation*}
The term containing the constant $\sqrt L$ has at most $2/\eta$ summands.
For $j\eta>2$, the sum of the nonincreasing exponential upper bounds
is at most the corresponding integral plus the first term.  This proves
$\sum_{j\ge0}\operatorname{Lip}(P_{j\eta}\phi)
\lesssim1+\mathcal C_{\tau,n}/\eta$ without an additional
$\eta^{-1/2}$ factor.

Finally, under $\rho$-strong concavity, synchronous noise and convex
reflection give
$\dd|X_t-Y_t|^2\le-2\rho|X_t-Y_t|^2\dd t$.
Thus $\operatorname{Lip}(P_t\phi)\le e^{-\rho t}$.
Its integral is $\rho^{-1}$, and its discrete sum is
$(1-e^{-\rho\eta})^{-1}\le1+(\rho\eta)^{-1}$.
\end{proof}

\begin{lemma}%[Time-uniform perturbation by the projection microgrid]
\label{lem:uniform-grid-coupling}
Let $P_t$ be the exact reflected semigroup for a fitted potential with
the bounds on the drift and its Lipschitz constant in Lemma~\ref{lem:architecture-drift} and the common
sampling profile.  Let $Q_{\eta,\delta}$ be one executable frozen-drift
projection macrostep, where $0<\eta\le1$, $0<\delta\le\eta/2$, and
$\eta/\delta\in\mathbb N$.  For every probability law $\nu$ and $M\ge1$,
\begin{align*}
W_1\left(\nu Q_{\eta,\delta}^M,\nu P_{M\eta}\right)
\lesssim\mathcal C_{\tau,n}
\left\{L_{b,n}B_{1,n}\eta+L_{b,n}\sqrt{d_a\tau\eta}
+(1+L_{b,n})\sqrt{d_a\tau\delta\log(e/\delta)}\right\}.
\end{align*}
The general and strongly concave choices of $\mathcal C_{\tau,n}$
are both valid under their respective conditions.
\end{lemma}

\begin{proof}
Write $L=L_{b,n}$, $B_1=B_{1,n}$, $b=\nabla F$, and
$Q=Q_{\eta,\delta}$.  Take the cube Skorokhod constant $L_\Gamma\ge1$.
We compare whole blocks of macrosteps before propagating their errors.
Set
\[
h_0=\{2L_\Gamma(1+L)\}^{-1},\qquad
q=\max\{1,\lfloor h_0/\eta\rfloor\},\qquad h=q\eta.
\]
Then $0<h\le1$, $h\ge\eta$, and
$
h^{-1}\le2h_0^{-1}=4L_\Gamma(1+L).
$
Indeed, if $\eta\le h_0$, then $h_0/2\le h\le h_0$; otherwise
$q=1$ and $h=\eta>h_0$.

For a block starting at $x$, let $B^\delta$ be the piecewise linear
interpolation of a Brownian path at its microgrid knots.  Put
$\kappa(u)=\eta\lfloor u/\eta\rfloor$ and couple
\begin{align*}
X_t&=\Gamma_{\cA}
 \left[x+\int_0^\cdot b(X_u)\dd u+\sqrt{2\tau}B\right](t),\\
Z_t&=\Gamma_{\cA}
 \left[x+\int_0^\cdot b(Z_{\kappa(u)})\dd u
                    +\sqrt{2\tau}B^\delta\right](t),\qquad 0\le t\le h.
\end{align*}
The macrogrid is contained in the microgrid.  Thus the drift integral
in the driving path of $Z$ is already linear on every microinterval.
Coordinatewise reflection of this linear driver equals the recursive
projection update at its knots, so $Z_h$ has law $Q^q(x,\cdot)$.

The restart property of the Skorokhod map, the uniform bound on the drift, and the
Brownian maximal inequality give, uniformly for $0\le u\le h$,
\[
\bE\|X_u-X_{\kappa(u)}\|_2
\lesssim B_1\eta+\sqrt{d_a\tau\eta}=:a_\eta.
\]
For a positive integer multiple $v$ of $\eta$ with $v\le h$, the
Brownian interpolation error satisfies
\[
\bE\sup_{u\le v}\|B_u-B_u^\delta\|_2
\lesssim\sqrt{d_a\delta\log(ev/\delta)}.
\]
To see this, each scalar bridge on a microinterval has a Gaussian
supremum tail with variance scale $\delta$.  A union bound over the
$v/\delta$ intervals and tail integration bound its squared maximum by
$C\delta\log(ev/\delta)$; summing over coordinates and applying
Cauchy--Schwarz proves the displayed estimate.

First, suppose $\eta\le h_0$, so $h\le h_0$.  For
$E_v=\bE\sup_{u\le v}\|X_u-Z_u\|_2$, the path Lipschitz inequality and
$\|b(X_u)-b(Z_{\kappa(u)})\|_2
\le L\{\|X_u-X_{\kappa(u)}\|_2+
          \|X_{\kappa(u)}-Z_{\kappa(u)}\|_2\}$ give
\begin{align*}
E_v
&\le L_\Gamma Lv E_v+C Lv a_\eta
       +C\sqrt{d_a\tau\delta\log(ev/\delta)}.
\end{align*}
Since $L_\Gamma Lv\le L_\Gamma Lh_0\le1/2$, the first term is absorbed
into the left-hand side.  If instead $\eta>h_0$, then $q=1$ and $v=h=\eta$.
In this case $Z_{\kappa(u)}=x$ for almost every $u\in[0,h]$; the same
path comparison uses only $\bE\|X_u-x\|_2\lesssim a_\eta$ and has no
$E_v$ term.  Consequently, in both cases,
\[
\sup_xW_1\{Q^q(x,\cdot),P_h(x,\cdot)\}
\le\varepsilon_h,
\qquad
\varepsilon_h=C\{Lh a_\eta+
                 \sqrt{d_a\tau\delta\log(eh/\delta)}\}.
\]
When $0<r<q$, the first case applies to $v=r\eta$ and also gives
$\sup_xW_1\{Q^r(x,\cdot),P_{r\eta}(x,\cdot)\}\le\varepsilon_h$.
For $r=0$ this error is zero.

Write $M=Nq+r$ with $0\le r<q$, and let $R=Q^q$.  The block operators
satisfy the exact identity
\[
Q^M-P_{M\eta}
=R^N(Q^r-P_{r\eta})+
 \sum_{i=0}^{N-1}R^i(R-P_h)P_{(N-1-i)h+r\eta}.
\]
For a $1$-Lipschitz function $\phi$, put $\psi=P_{r\eta}\phi$.
Synchronous coupling and convex reflection give
$\operatorname{Lip}(P_t\phi)\le e^{Lt}$.
If $r>0$, then $q>1$, so $r\eta\le h_0$ and
$\operatorname{Lip}(\psi)\le e^{Lh_0}\le e^{1/2}$; if $r=0$, its
Lipschitz seminorm is at most one.  Apply the preceding local bounds
and Lemma~\ref{lem:reflected-semigroup-smoothing} with spacing $h$ to
the block identity.  Since all iterates of $R$ preserve probability,
\begin{equation*}
|(\nu Q^M-\nu P_{M\eta})\phi|
\lesssim\varepsilon_h
 \left\{1+\sum_{j=0}^{N-1}\operatorname{Lip}(P_{jh}\psi)\right\}
\lesssim\varepsilon_h
 \left(1+\frac{\mathcal C_{\tau,n}}{h}\right).
\end{equation*}
This also covers $N=0$ by the shorter-block bound.  Finally,
$h\le1$, $\mathcal C_{\tau,n}\ge1$, $h^{-1}\lesssim1+L$, and
$\log(eh/\delta)\le\log(e/\delta)$.  Taking the supremum over $\phi$
proves the result.  The Brownian interpolation error is accumulated
once per block of duration comparable to $(1+L)^{-1}$ when $\eta$ is
small, rather than once per macrostep.  
 \end{proof}

\begin{proof}[Proof of Theorem~\ref{thm:annealed-langevin}]
Condition on the past and the fitted critic, including the Bellman
sample and the independent points used for its Sobolev penalty.
Lemma~\ref{lem:uniform-grid-coupling} gives a numerical error bound uniformly
in the state and the conditioning variables, using Lemma~\ref{lem:architecture-drift}.
For the exact diffusion, Lemmas~\ref{lem:exact-reflected-entropy}
and~\ref{lem:init-entropy}, followed by the profile transport inequality,
give
\begin{equation*}
W_1^2(\lambda_{\cA}P_{M\eta},\pi_F)
\le\frac{4\chi_\tau B_F}{\tau}
             e^{-2\alpha_\tau\tau M\eta}.
\end{equation*}
For the critic-induced Gibbs-target error, put $F=\widehat Q_{k+1}$ and
$Q=Q^{\widehat\pi_k}$.  Apply both bounds in
Lemma~\ref{lem:gibbs-risk-transfer} at each fixed state.
Integrate over $\lambda_\cS$ and use Cauchy--Schwarz, then take
$\bE_k$ over the current data and penalty points and use Jensen's
inequality.  The gradient route uses \eqref{eq:gradient-l2-rate}.
For the value route, coverage gives
\[
\|F-Q\|_{L^2(\lambda_d)}^2
\le(1-\gamma)^{-1}\|F-Q\|_{L^2(\mu_k)}^2,
\]
and \eqref{eq:value-l2-rate} applies.  Both comparisons hold for the same
fitted critic, so their minimum is also an upper bound.
The triangle inequality links the executable actor law, the exact
finite-time diffusion law, the fitted Gibbs target, and the population Gibbs
target.  Combining the two square-root numerical terms by
Cauchy--Schwarz and using $1+L_{b,n}\le2L_{b,n}$ gives the stated
numerical remainder.
\end{proof}

\section{Proofs for End-to-End Actor--Critic Error Propagation}
\label{app:return-propagation}
The transition-density bound transfers integrated actor error to the
measures in the policy-improvement recursion, yielding the return bounds
and temperature calibrations in Section~\ref{sec:end-to-end}.

\begin{proof}[Proof of Theorem~\ref{thm:annealed-end-to-end}]
The state optimality operator maps $C(\cS)$, the space of continuous
functions on the state space, into itself by Assumption~\ref{ass:primitive-mdp},
compactness of the action cube, and dominated convergence.  It is
a $\gamma$-contraction, and its fixed point $V^\star$ is continuous.
The function $\int u\,\mathcal R(\dd u\mid s,a)+\gamma\int V^\star(s')p(s'\mid s,a)\dd s'$ is
continuous; the measurable maximum theorem supplies a measurable maximizing
action.  The resulting stationary policy $\pi^\star$ satisfies
$V^{\pi^\star}=V^\star$.

Fix $k$ and write $\pi=\widehat\pi_k$, $\pi'=\widehat\pi_{k+1}$.
The action-Lipschitz constant of $Q^\pi$ with respect to the Euclidean norm is at most
$L_a=\sqrt{d_a}L$.  Let $e_k=\cT_V^\star V^\pi-\cT_V^{\pi'}V^\pi\ge0$.  For
$q_k=\cG_\tau(Q^\pi)$, Lemma~\ref{lem:soft-gap} and the coupling
definition of $W_1$ imply
\begin{equation*}
e_k(s)\le b_\tau+
\left|\int Q^\pi(s,a)(q_k-\pi')(\dd a\mid s)\right|
\le b_\tau+L_aW_1\left(q_k(\cdot\mid s),\pi'(\cdot\mid s)\right).
\end{equation*}
To obtain the last inequality, integrate
$|Q^\pi(s,a)-Q^\pi(s,a')|\le L_a\|a-a'\|_2$ against an arbitrary
coupling and take the infimum.  Define
$(P_{\cS}^{\nu}v)(s)=\int\int v(s')P(\dd s'\mid s,a)\nu(\dd a\mid s)$.
Subtracting the Bellman equations for the two policies yields
\begin{equation*}
V^{\pi'}-V^\pi=(I-\gamma P_{\cS}^{\pi'})^{-1}
(\cT_V^{\pi'}V^\pi-V^\pi).
\end{equation*}
Since $\cT_V^\star V^\pi\ge V^\pi$ and the inverse is a positive
operator, $V^\pi-V^{\pi'}\le(I-\gamma P_{\cS}^{\pi'})^{-1}e_k$.
Adding and subtracting the optimality update now gives
\begin{align*}
V^\star-V^{\pi'}
&\le\gamma P_{\cS}^{\pi^\star}(V^\star-V^\pi)
+e_k+\gamma P_{\cS}^{\pi'}(V^\pi-V^{\pi'})\\
&\le\gamma P_{\cS}^{\pi^\star}(V^\star-V^\pi)
+(I-\gamma P_{\cS}^{\pi'})^{-1}e_k.
\end{align*}
Iterating this inequality and expanding each positive resolvent gives
\begin{align*}
V^\star-V^{\widehat\pi_K}
\le{}&\gamma^K(P_{\cS}^{\pi^\star})^K
                   (V^\star-V^{\widehat\pi_0}) +\sum_{k=0}^{K-1}\sum_{j=0}^{\infty}
\gamma^{K-1-k+j}(P_{\cS}^{\pi^\star})^{K-1-k}
                   (P_{\cS}^{\widehat\pi_{k+1}})^j e_k.
\end{align*}
Nonnegativity permits interchanging integration and summation of the
$e_k$ terms.  Each measure
$\rho_0(P_{\cS}^{\pi^\star})^{K-1-k}
(P_{\cS}^{\widehat\pi_{k+1}})^j$
is either $\rho_0$ or a state law after at least one transition.
For any input state law $\nu$ and any measurable policy $\pi$,
\[
\frac{\dd(\nu P_{\cS}^{\pi})}{\dd\lambda_{\cS}}(s')
=\int_{\cS}\int_{\cA}p(s'\mid s,a)\pi(\dd a\mid s)\nu(\dd s)
\le\overline p.
\]
Each measure of integration therefore has density at most
$\max\{\overline\rho,\overline p\}$, pathwise for learned policies.

Apply $e_k(s)\le b_\tau+
L_aW_1\{q_k(\cdot\mid s),\widehat\pi_{k+1}(\cdot\mid s)\}$ before
summing.  The constant $b_\tau$ integrates exactly under every probability
measure.  The two geometric sums therefore give
\begin{align*}
J(\pi^\star)-J(\widehat\pi_K)
\le{}&2B_Q\gamma^K+
\frac{1-\gamma^K}{(1-\gamma)^2}b_\tau +\frac{L_a\max\{\overline\rho,\overline p\}}{1-\gamma}
\sum_{k=0}^{K-1}\gamma^{K-1-k}
\mathsf w_1\{\widehat\pi_{k+1},
                    \cG_\tau(Q^{\widehat\pi_k})\}.
\end{align*}
Insert $\widetilde\pi_{k+1}$ into each Wasserstein comparison and apply
the triangle inequality.  Taking expectations gives
\eqref{eq:outer-error-overview}.
For the sharper return bound, set $F=\widehat Q_{k+1}$ and
$Q=Q^{\widehat\pi_k}$.  Compare the executed policy first with
$\widetilde\pi_{k+1}$, and apply the two Gibbs perturbation routes in
Lemma~\ref{lem:gibbs-risk-transfer}.  Integration over the state variable, conditional
Cauchy--Schwarz, and coverage give
\begin{align*}
\bE_k\int e_k\,\dd\lambda_\cS
\lesssim{}&b_\tau+
\bE_k\mathsf w_1(\widehat\pi_{k+1},\widetilde\pi_{k+1})\\
&+\min\Biggl\{
\left(\frac{\chi_\tau\min\{D_\tau,\tau^{-d_a}\}}
                 {\alpha_\tau\tau^2}
 \bE_k\|\nabla_a F-\nabla_a Q\|_{L^2(\lambda_d)}^2\right)^{1/2},\\
&\hspace{18mm}
 e^{3B_F/(2\tau)}\bE_k\|F-Q\|_{L^2(\mu_k)}^2
 \Biggr\}.
\end{align*}
The fitted-target sampling error contains only the numerical and
mixing terms in the proof of Theorem~\ref{thm:annealed-langevin}.
Insert the two critic risks from Theorem~\ref{thm:critic-l2} and apply
the preceding positive-resolvent recursion, its pathwise density bound,
and the tower property.  Summing the geometric series proves
\eqref{eq:annealed-error-decomposition}, retaining the factor
$(1-\gamma)^{-2}$ from outer propagation.
\end{proof}

\begin{proof}[Proof of Corollary~\ref{cor:multimodal-end-to-end}]
On the branch $\tau_n=c_\tau/\log n$, the sampling profile gives
\begin{equation*}
\alpha_{\tau_n}\asymp n^{-2B_F/c_\tau},\quad
D_{\tau_n}\le n^{2B_F/c_\tau},\quad
\log\mathcal C_{\tau_n,n}\lesssim(\log n)^2.
\end{equation*}
The last estimate follows from the explicit coupling constant and
$\log(1+B_{1,n})+\log(1+L_{b,n})\lesssim(\log n)^2$.
The two macrostep contributions obey
\begin{equation*}
\mathcal C_{\tau_n,n}L_{b,n}\sqrt{\tau_n\eta}
\lesssim(\log n)^{-5/2},\qquad
\mathcal C_{\tau_n,n}L_{b,n}B_{1,n}\eta
\lesssim(\log n)^{-4}.
\end{equation*}
With $\delta=\eta/2$, the architecture bounds give
$\log(e/\delta)\lesssim(\log n)^2$ and hence
\[
\mathcal C_{\tau_n,n}(1+L_{b,n})
\sqrt{\tau_n\delta\log(e/\delta)}
\lesssim(\log n)^{-3/2}.
\]
The fixed multiplier in $M\eta$
makes the exact-diffusion mixing term smaller than these errors, and the
multiplier in $K$ does the same for $\gamma^K$.  Finally,
\begin{align*}
b_{\tau_n}&\lesssim\frac{1+\log\log n}{\log n},\\
e^{3B_F/(2\tau_n)}n^{-2\beta/(2\beta+d)}(\log n)^3
&=n^{-2\beta/(2\beta+d)+3B_F/(2c_\tau)}(\log n)^3.
\end{align*}
The condition $c_\tau>3B_F(2\beta+d)/(4\beta)$ makes the power of $n$
strictly negative.  The value route in the actor theorem also vanishes:
its statistical term is bounded by
$n^{-\beta/(2\beta+d)+3B_F/(4c_\tau)}(\log n)^2$.
On the remaining finite branch $\tau_n=1$,
the return is bounded by $2B_Q$, and the displayed positive rate function
has a positive minimum; enlarging a fixed constant covers this branch.
For the operation counts, $M\eta\asymp
n^{2B_F/c_\tau}(\log n)^2$ and the schedules give
$\log(1/\eta)+\log(1/\delta)\lesssim(\log n)^2$.
Multiplying by the $O\{Kn/(1-\gamma)\}$ expected number of actor calls and by
$O(\mathcal M_n)$ per gradient evaluation preserves the
$\exp\{O((\log n)^2)\}$ bound.
\end{proof}

\begin{proof}[Proof of Corollary~\ref{cor:strong-concave-end-to-end}]
Use the estimator and calibration of Theorem~\ref{thm:critic-l2}.
The value and action-gradient risk bounds
\eqref{eq:value-l2-rate} and \eqref{eq:gradient-l2-rate} apply directly.
On the common probability-one event in the hypothesis, the fitted
and population action potentials satisfy the strong-concavity profile
of Lemma~\ref{lem:profile-verification}, uniformly in state and iteration.

\emph{Gradient-risk route.}
Condition on the fitted critic and fix $s$;
write $Q(a)=Q^{\widehat\pi_k}(s,a)$, $F(a)=\widehat Q_{k+1}(s,a)$,
and let $\pi_F$ and $\pi_Q$ be their Gibbs laws at temperature $\tau$.
Choose $a^\star\in\argmax_{a\in\cA}Q(a)$.
The $2\rho$-strong concavity of $Q$ gives
\begin{equation*}
Q(a^\star)-Q(a)
\le\nabla Q(a)^\top(a^\star-a)-\rho\|a^\star-a\|_2^2.
\end{equation*}
Let $p_F$ be the density of $\pi_F$, $n_{\rm out}$ the outward unit
normal on each face of the cube, and $\dd\sigma$ surface measure.
Since $\nabla p_F=\tau^{-1}p_F\nabla F$, integration by parts yields
\begin{align*}
\int_\cA\nabla F(a)^\top(a^\star-a)\dd\pi_F(a)
&=d_a\tau+
\tau\int_{\partial\cA}p_F(a)(a^\star-a)^\top
                 n_{\rm out}(a)\dd\sigma(a)\\
&\le d_a\tau.
\end{align*}
The last inequality holds on each face because
$(a^\star-a)^\top n_{\rm out}(a)\le0$; it does not require an interior
maximizer.  Young's inequality gives
\begin{equation*}
(\nabla Q-\nabla F)^\top(a^\star-a)
\le\frac{\|\nabla Q-\nabla F\|_2^2}{4\rho}
     +\rho\|a^\star-a\|_2^2.
\end{equation*}
Combining these inequalities with Lemma~\ref{lem:profile-verification} gives
\begin{align*}
\int_\cA\{Q(a^\star)-Q(a)\}\dd\pi_F(a)
&\le d_a\tau+\frac1{4\rho}
                \int_\cA\|\nabla Q-\nabla F\|_2^2\dd\pi_F\\
&\lesssim\tau+\tau^{-d_a}
                \|\nabla Q-\nabla F\|_{L^2(\lambda_{\cA})}^2.
\end{align*}
Integrate over $\lambda_{\cS}$, take $\bE_k$, and apply
\eqref{eq:gradient-l2-rate}.

\emph{Value-risk route.}
For the same fixed state, put $h=F-Q$ and
$F_t=Q+th$ for $0\le t\le1$.  The identity
\eqref{eq:gibbs-path-kl} gives
\begin{equation*}
\KL(\pi_F\|\pi_Q)
=\tau^{-2}\int_0^1t\operatorname{Var}_{\pi_{F_t}}(h)\dd t
\lesssim\tau^{-d_a-2}
                 \|F-Q\|_{L^2(\lambda_{\cA})}^2.
\end{equation*}
Each $F_t$ is concave and satisfies
$\|F_t\|_\infty\le B_F$.  The homothety argument in
Lemma~\ref{lem:profile-verification} therefore bounds its Gibbs density
by a fixed multiple of $\tau^{-d_a}$, independently of the uniform
bound on the fitted gradient; $F_t$ need not belong to the DNN class.
To bound the greedy gap, let
$U(a)=Q(a^\star)-Q(a)$ and
$Z_U(v)=\int_\cA e^{-U(a)/v}\dd a$ for $v>0$.
Concavity of $Q$ makes $U$ convex, with $U(a^\star)=0$.
The affine map $a\mapsto a^\star+(a-a^\star)/2$ maps the cube into
itself and has Jacobian $2^{-d_a}$.  Thus
\begin{equation*}
Z_U(\tau)
\ge2^{-d_a}\int_\cA
   e^{-U(a^\star+(a-a^\star)/2)/\tau}\dd a
\ge2^{-d_a}Z_U(2\tau).
\end{equation*}
Therefore \eqref{eq:gibbs-greedy-entropy} bounds
$(2\tau)^{-1}\int U\dd\pi_F-\KL(\pi_F\|\pi_Q)$ by $d_a\log2$.
Consequently,
\begin{equation*}
\int_\cA\{Q(a^\star)-Q(a)\}\dd\pi_F(a)
\lesssim\tau+\tau^{-d_a-1}
                         \|F-Q\|_{L^2(\lambda_{\cA})}^2.
\end{equation*}
After integration over $\lambda_{\cS}$, coverage gives
\begin{equation*}
\|\widehat Q_{k+1}-Q^{\widehat\pi_k}\|_{L^2(\lambda_d)}^2
\le\frac1{1-\gamma}
\|\widehat Q_{k+1}-Q^{\widehat\pi_k}\|_{L^2(\mu_k)}^2.
\end{equation*}
Take $\bE_k$ and apply \eqref{eq:value-l2-rate}.

\emph{Executable actor and outer iteration.}
In state-dependent notation, for
$e_k(s)=\max_aQ^{\widehat\pi_k}(s,a)
-\int Q^{\widehat\pi_k}(s,a)\widehat\pi_{k+1}(\dd a\mid s)$,
the action-Lipschitz bound $L_a=\sqrt{d_a}L$ gives
\begin{align*}
e_k(s)\le{}&
\int_\cA\{\max_u Q^{\widehat\pi_k}(s,u)
                 -Q^{\widehat\pi_k}(s,a)\}
                  \widetilde\pi_{k+1}(\dd a\mid s) +L_aW_1\{\widehat\pi_{k+1}(\cdot\mid s),
                  \widetilde\pi_{k+1}(\cdot\mid s)\}.
\end{align*}
Both preceding risk bounds apply to the first term.  For the second, apply
the numerical and mixing bounds from the proof of Theorem~\ref{thm:annealed-langevin}.
The strong-concavity
profile gives $\alpha_\tau=\rho/\tau$, $\chi_\tau=\tau/\rho$, and
$\mathcal C_{\tau,n}=1+\rho^{-1}$.  Hence
\begin{align*}
\bE_k\mathsf w_1(\widehat\pi_{k+1},\widetilde\pi_{k+1})
\lesssim{}&L_{b,n}\sqrt{\tau\eta}+L_{b,n}B_{1,n}\eta+
       (1+L_{b,n})\sqrt{\tau\delta\log(e/\delta)}
       +e^{-\rho M\eta}.
\end{align*}
The entropy comparison involves only smooth Gibbs laws, not the projection law.

Apply the positive-resolvent recursion from the proof of
Theorem~\ref{thm:annealed-end-to-end} to this bound on $e_k$.
The state-density bound holds pathwise, so state integration and the
tower property give
\begin{align*}
\bE\{J(\pi^\star)-J(\widehat\pi_K)\}
\lesssim{}&\frac{\gamma^K}{1-\gamma}
+\frac{1}{(1-\gamma)^2}\Bigl[\tau+L_{b,n}\sqrt{\tau\eta}
+L_{b,n}B_{1,n}\eta\\
&\quad+(1+L_{b,n})\sqrt{\tau\delta\log(e/\delta)}
 +e^{-\rho M\eta}\\
&+\min\Bigl\{
 \tau^{-d_a-1}n^{-2\beta/(2\beta+d)}(\log n)^3,\\
&\hspace{17mm}
 \tau^{-d_a}n^{-2(\beta-1)/(2\beta+d)}
                     (\log n)^{3(\beta-1)/\beta}\Bigr\}\Bigr].
\end{align*}
The minimum is valid because both conditional risk bounds hold for the
same fitted critic.  Balancing $\tau$ with each statistical term gives
the two rates in \eqref{eq:strong-concave-return-rate}.  Choosing $\tau_n$
of the smaller order makes at least one statistical term of order
$\tau_n$.
With
$\eta\asymp\tau_n\{(1+L_{b,n})(1+B_{1,n})\}^{-2}$,
both macrostep terms are at most a fixed multiple of $\tau_n$.
The derivative bounds and the mesh choice imply
$\log(e/\delta)\lesssim(\log n)^2$, and hence
\begin{equation*}
(1+L_{b,n})\sqrt{\tau_n\delta\log(e/\delta)}
\lesssim\tau_n.
\end{equation*}
The conditions $e^{-\rho M\eta}\le\tau_n$ and $\gamma^K\le\tau_n$
bound mixing and outer iteration by $\tau_n$.
Integer rounding preserves these inequalities.
This proves \eqref{eq:strong-concave-return-rate}.
Finally, each action uses $M\asymp(\log n)/\eta$ gradients and
$M\eta/\delta\asymp(\log n)/\delta$ projections.  Multiplying
by the $O\{Kn/(1-\gamma)\}$ expected number of actor calls gives the stated sampling counts.
\end{proof}

\section{Auxiliary Results}
\label{app:auxiliary}

This appendix records the external analytic inputs and derives the
required $W_1$-transport consequence of LSI on the cube.

\begin{lemma}[Bounded perturbation and tensorization: Proposition 5.1.6 and Proposition 5.2.7 of \citet{bakry2014analysis}]
\label{lem:aux-lsi-information}
Suppose a probability measure $\mu$ satisfies
$I(p\|\mu)\ge2\alpha\KL(p\|\mu)$ and
$\dd q=e^V\dd\mu/\int e^V\dd\mu$ for a bounded function $V$.  Then $q$
satisfies an LSI with constant at least $\alpha e^{-\osc(V)}$.  Log-Sobolev
inequalities tensorize over finite products.
\end{lemma}

After tensorizing the interval inequality, apply the
Holley--Stroock principle \citep{holley1987logarithmic} with
$V=F/\tau$, whose oscillation is at most $2B_F/\tau$.

\begin{lemma}[Theorem A1 of \citet{schlichting2019poincare}]
\label{lem:aux-strong-lsi-transport}
Let $\Omega\subset\bR^q$ be convex, and let
$\dd\mu=e^{-U}\dd x/\int_\Omega e^{-U}\dd x$, where $U$ is twice
differentiable and $\nabla^2U\succeq\alpha I_q$ on $\Omega$ for some
$\alpha>0$.  Then $\mu$ satisfies
\begin{equation*}
I(\nu\|\mu)\ge2\alpha\KL(\nu\|\mu).
\end{equation*}
\end{lemma}

On the action cube, this LSI implies the required transport bound by
the following Herbst argument.  For a smooth $1$-Lipschitz function
$f$, put $\psi(t)=\log\int e^{tf}\dd\mu$ and
$\dd\mu_t=e^{tf-\psi(t)}\dd\mu$.  Applying the LSI to $\mu_t$ gives
\[
t\psi'(t)-\psi(t)
=\KL(\mu_t\|\mu)
\le\frac{t^2}{2\alpha}\int\|\nabla f\|_2^2\dd\mu_t
\le\frac{t^2}{2\alpha}.
\]
Integrating $(\psi(t)/t)'\le(2\alpha)^{-1}$ from zero yields
$\psi(t)\le t\int f\dd\mu+t^2/(2\alpha)$.
For any $\nu$ with finite relative entropy, nonnegativity of
$\KL(\nu\|\mu_t)$ therefore gives, for $t>0$,
\[
\int f\dd\nu-\int f\dd\mu
\le\frac{\KL(\nu\|\mu)}{t}+\frac{t}{2\alpha}.
\]
Optimizing over $t$ bounds the left-hand side by
$\{2\KL(\nu\|\mu)/\alpha\}^{1/2}$.
Every $1$-Lipschitz function on the cube admits a $1$-Lipschitz
extension to $\bR^{d_a}$; convolution with a smooth probability kernel
gives smooth $1$-Lipschitz approximations converging uniformly on the
cube.  Passing to the limit and using Kantorovich duality yields
\[
W_1^2(\nu,\mu)\le\frac{2}{\alpha}\KL(\nu\|\mu).
\]
The inequality is immediate when the relative entropy is infinite.
For $\mu=\cG_\tau(F)$ and a $\rho$-strongly concave $F$, take
$U=-F/\tau$ and $\alpha=\rho/\tau$.

\begin{lemma}[Pinsker's inequality: Lemma 2.5 of \citet{tsybakov2009nonparametric}]
\label{lem:aux-information}
For any probability measures $\mu$ and $\nu$,
$
\TV(\mu,\nu)\le\sqrt{\KL(\mu\|\nu)/2}.
$
\end{lemma}

\begin{lemma}[Theorem 1.4 and Corollary 1.6 of \citet{kruk2007explicit}]
\label{lem:aux-reflection}
The two-sided Skorokhod problem on $[0,1]$ has a unique solution for
every continuous driving path.  Its nonanticipating map $\Gamma_{[0,1]}$
satisfies, with a constant independent of $T$,
\[
\sup_{t\le T}|\Gamma_{[0,1]}(x)(t)-\Gamma_{[0,1]}(y)(t)|
\lesssim\sup_{t\le T}|x(t)-y(t)|.
\]
\end{lemma}

Normal reflection on the cube acts coordinatewise, so the same
inequality holds for $\Gamma_{\cA}$ in the Euclidean uniform path norm,
with Lipschitz constant $C_{d_a}$.  For a globally Lipschitz drift
$b$, apply Picard iteration to
\[
X=\Gamma_{\cA}\!\left\{X_0+\int_0^\cdot b(X_u)\,\dd u
                         +\sqrt{2\tau}B_\cdot\right\}.
\]
This Picard map is a contraction on intervals satisfying
$C_{d_a}\operatorname{Lip}(b)T<1$.  Successive restarts give a
unique adapted strong solution on every finite interval.

\section{Supplementary experiments and evaluation details}
\label{app:experiments}

\subsection{Continuous-control evaluation details}
\label{app:d4rl-details}

\paragraph{Data and comparison methods.}
Within each outer trajectory split, we sample at most 20,000 training
and 5,000 test transitions.  Inner validation holds out 20\% of the
trajectories represented in training.  Methods share the partitions
and training-based coordinate and target scaling; coordinates outside
the training range are clipped to $[0,1]$.
Inner RMSE selects among 12 sigmoid DNN candidates combining depths
$\{1,2,3\}$, widths $\{128,256\}$, and learning rates
$\{0.001,0.003\}$.  Each polynomial baseline selects ridge strength from
$\{0,10^{-7},10^{-5},10^{-3},10^{-1},1,10,100\}$ in the unnormalized
squared-error objective, leaving the intercept unpenalized.
Quadratic features include coordinates, their squares, and
adjacent-coordinate products.  Search budgets differ across model
classes, and neural optimization has no certified global optimum.

\paragraph{Fixed-critic evaluation.}
The REM actor starts uniformly on the training action cube and uses
$M=20$, $\eta=0.01$, four projection microsteps with $\delta=0.0025$,
and $\tau=0.10$ in standardized target units.  Actions are mapped back
using training minima and ranges and clipped to simulator bounds;
this may cover only part of the simulator action box.  Reset states
and actor random streams are paired across methods, although induced
trajectories may differ.  Evaluation returns do not select or retune
models, and the fixed sampling budget provides no mixing certificate.
The five-episode return average defined in the main text uses observed
rewards without a fitted-value tail.

\paragraph{Standardization and interpretation.}
In the score $Z_{jh}$ defined in the main text, $\sigma_j>0$ is the root
mean square deviation of split $j$'s training targets, using their
count as the divisor.  Table~\ref{tab:d4rl-policy-z} reports
\[
\overline Z_h=\frac1{100}\sum_{j=1}^{100} Z_{jh},\qquad
s_h=\left\{\frac1{99}\sum_{j=1}^{100}
(Z_{jh}-\overline Z_h)^2\right\}^{1/2}
\]
as mean $\pm$ sample standard deviation, retaining all 100 splits
separately for each dataset.
Logged targets include sampled transition positions, whereas evaluation
returns start at resets.  A negative $Z_{jh}$ therefore indicates a
return below the training-target mean, not below a random policy.
Because $\sigma_j$ varies across splits, the mean standardized-score
difference between two methods weights their discounted-return
differences by $1/\sigma_j$.  Scores are not pooled across environments
or dataset variants.  Matching discounted random--expert references
are unavailable for these MuJoCo-v5 evaluations, so the scores are not
official D4RL benchmark scores.

Logged D4RL v2 data and MuJoCo-v5 evaluations differ in simulator
version, and logged returns need not equal the extracted policy's
action values.  These differences, state--action distribution shift,
and finite sampling limit inference from prediction to control;
their separate effects are not identified by this experiment.

\subsection{Observed building-energy decision utility}
\label{app:energy-utility}
We use the Energy Efficiency data of \citet{tsanas2012energy} from the
University of California, Irvine (UCI) Machine Learning Repository.\footnote{\url{https://archive.ics.uci.edu/dataset/242/energy+efficiency}}
The 768 Ecotect simulations cover 12 building shells and 64 feasible
designs per shell.  Relative compactness, surface area, wall area, roof
area, and overall height define the shell context; orientation, glazing
area, and glazing-area distribution define the action.
The task is to select designs that reduce heating load for an unseen shell.

We use 100 outer splits shared across methods, each assigning eight
complete shells (512 observations) to training and four shells
(256 observations) to testing.  The linear and sigmoid
critics use the same training-based scaling
and standardized heating-load target.  The linear baseline uses ridge
coefficient $10^{-3}$ in the unnormalized least-squares objective,
with an unpenalized intercept.
The sigmoid critic uses squared prediction loss with the first-order
action-derivative penalty.
Two of the eight training shells form
the inner validation set, which selects
$\lambda_n\in\{10^{-4},10^{-3},10^{-2}\}$ by validation RMSE.
Each neural fit uses $\ell_n=512$ independent uniform integration points;
the selected critic is refitted on all eight training shells.
Test coordinates outside the training range
are clipped to $[0,1]$.  Gibbs weights are computed from the negative
standardized predicted heating load at temperature $\tau=0.10$ over
the 64 observed feasible actions, using direct normalization rather than
Langevin sampling.  We define the reward as $-Y_1(c,a)$, where
$Y_1(c,a)$ is the observed heating load, so higher reward corresponds to
lower load.  For the four held-out contexts $\mathcal C_{\rm te}$, define
\begin{align*}
\widehat L(\pi)
&:=
\frac1{|\mathcal C_{\rm te}|}
\sum_{c\in\mathcal C_{\rm te}}\sum_{a\in\cA_c}
\pi(a\mid c)Y_1(c,a),
\\
\widehat{\operatorname{Reg}}(\pi)
&:=
\widehat L(\pi)
-
\frac1{|\mathcal C_{\rm te}|}
\sum_{c\in\mathcal C_{\rm te}}\min_{a\in\cA_c}Y_1(c,a),
\\
\widehat M_{25}(\pi)
&:=
\frac1{|\mathcal C_{\rm te}|}
\sum_{c\in\mathcal C_{\rm te}}\sum_{a\in\cA_c}
\pi(a\mid c)
\ind\{Y_1(c,a)\le q_{.25}(c)\},
\end{align*}
where $\cA_c$ is the 64-design set for shell $c$, and $q_{.25}(c)$ is the
within-shell lower quartile of heating load.  Reward equals
$-\widehat L(\pi)$; lower load, lower regret, and greater efficient-quartile
mass indicate better decisions.  Oracle selects a minimum-load design
using the observed outcomes for each test shell and serves as a
reference.

\begin{figure}[H]
\centering
\includegraphics[width=0.78\textwidth]{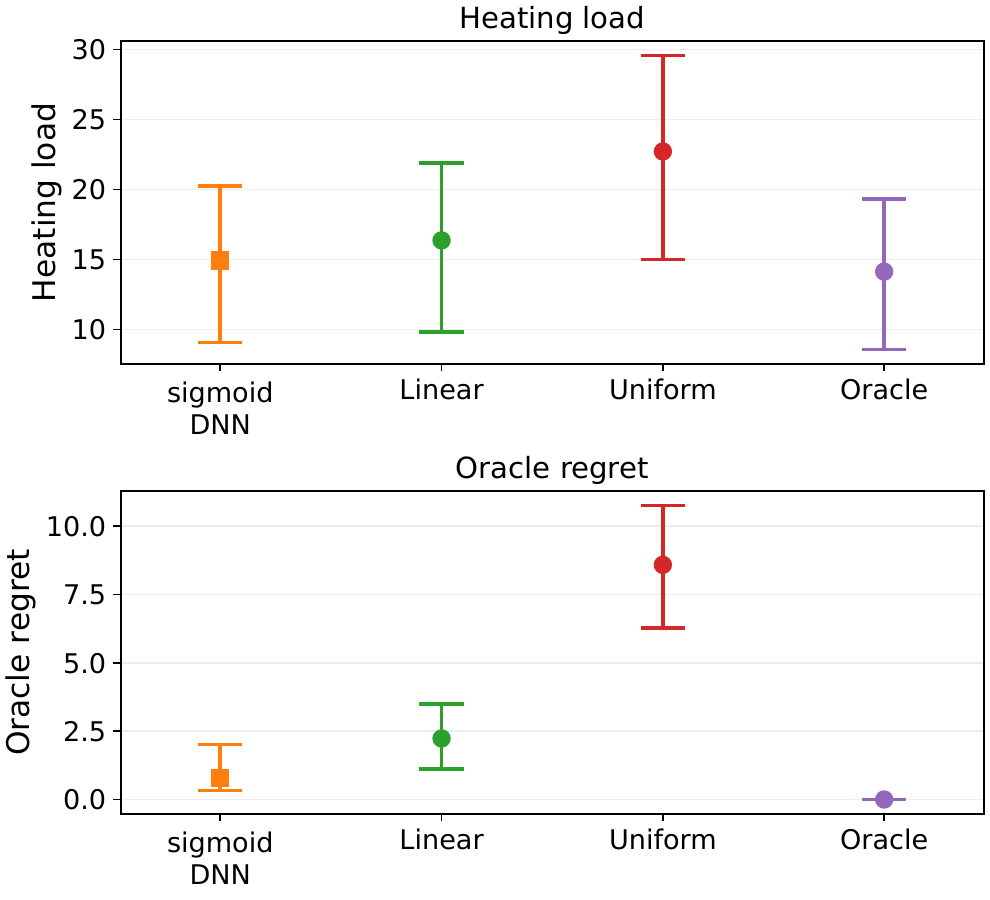}
\caption{Directly observed decision metrics on held-out building shells; error bars are empirical 2.5\%--97.5\% ranges over 100 context-level splits.}
\label{fig:energy-policy}
\end{figure}

Table~\ref{tab:energy-policy} reports mean decision metrics with
empirical 2.5\%--97.5\% split ranges in brackets; Figure~\ref{fig:energy-policy}
displays heating load and regret.  These ranges describe variation
across partitions of the fixed dataset; they are not confidence
intervals for performance across a population of buildings.
Table entries use the two-decimal convention of
Section~\ref{sec:interactive-experiments}, with Oracle values rounded to
the nearest hundredth and split-range endpoints rounded outward.  Boldface identifies the best mean
among the implementable policies, excluding Oracle.  The sigmoid DNN
policy has lower mean heating load and regret and greater
efficient-quartile mass than Linear and Uniform.  Its paired
heating-load difference from the Linear policy averages $-1.4423$,
with an empirical 2.5\%--97.5\% split range of $[-2.6301,-0.1179]$.

\begin{table}[H]
\centering
\caption{Observed held-out-shell decision performance over 100 context splits.}
\label{tab:energy-policy}
\vspace{0.2em}
\footnotesize
\setlength{\tabcolsep}{6pt}
\setlength{\extrarowheight}{5pt}
\renewcommand{\arraystretch}{1.3}
\begin{tabular}{lrrrr}
\toprule
Method & \shortstack{Heating\\load} & Reward & \shortstack{Oracle\\regret} & \shortstack{Efficient-quartile\\probability} \\
\midrule
sigmoid DNN & \shortstack{\(\mathbf{14.93}\)\\\([9.07,\,20.28]\)} & \shortstack{\(\mathbf{-14.93}\)\\\([-20.28,\,-9.07]\)} & \shortstack{\(\mathbf{0.78}\)\\\([0.32,\,2.02]\)} & \shortstack{\(\mathbf{0.95}\)\\\([0.84,\,1.00]\)} \\
Linear & \shortstack{\(16.38\)\\\([9.82,\,21.91]\)} & \shortstack{\(-16.37\)\\\([-21.91,\,-9.82]\)} & \shortstack{\(2.24\)\\\([1.11,\,3.49]\)} & \shortstack{\(0.88\)\\\([0.82,\,0.93]\)} \\
Uniform & \shortstack{\(22.72\)\\\([15.01,\,29.58]\)} & \shortstack{\(-22.72\)\\\([-29.58,\,-15.01]\)} & \shortstack{\(8.58\)\\\([6.27,\,10.75]\)} & \shortstack{\(0.25\)\\\([0.25,\,0.25]\)} \\
Oracle & \shortstack{\(14.15\)\\\([8.58,\,19.32]\)} & \shortstack{\(-14.15\)\\\([-19.32,\,-8.58]\)} & \shortstack{\(0.00\)\\\([0.00,\,0.00]\)} & \shortstack{\(1.00\)\\\([1.00,\,1.00]\)} \\
\bottomrule
\end{tabular}
\end{table}

\bibliographystyle{spbasic}
\bibliography{ref}

\end{document}